\documentclass{article} %
\usepackage{iclr2025_conference,times}

\usepackage{amsmath,amsfonts,bm}

\def\eqref#1{equation~\ref{#1}}

\def\1{\bm{1}}

\DeclareMathAlphabet{\mathsfit}{\encodingdefault}{\sfdefault}{m}{sl}
\SetMathAlphabet{\mathsfit}{bold}{\encodingdefault}{\sfdefault}{bx}{n}

\usepackage{hyperref}
\usepackage{url}

\title{QPM: Discrete Optimization for Globally\\ Interpretable Image Classification}

\usepackage[utf8]{inputenc} %
\usepackage[T1]{fontenc}    %
\usepackage{url}            %
\usepackage{booktabs}       %
\usepackage{amsfonts}       %
\usepackage{nicefrac}       %
\usepackage{microtype}      %
\usepackage{amssymb}%
\usepackage{pifont}%
\usepackage{tikz}
\usetikzlibrary{shapes.geometric, arrows, fit,positioning, fit,calc}
\usepackage{algorithm}
\usepackage{placeins}
\usepackage{algorithmic}
\usepackage[algo2e]{algorithm2e} 
\usepackage{times}
\usepackage{epsfig}
\usepackage{graphicx}
\usepackage{amsmath}
\usepackage{numprint}
\usepackage{amssymb}
\usepackage{rotating}
\usepackage{makecell}
\usepackage{svg}
\usepackage{lineno}
\usepackage{multirow}
\usepackage{smartdiagram}
\usepackage{csquotes}
\usepackage{caption}
\usepackage{arydshln}
\usepackage{subcaption}
\usepackage{rotating}

\usepackage[acronym,nogroupskip]{glossaries}
\glsdisablehyper
\usepackage{cleveref}

\newcommand{\helperindex}[1]{\ensuremath{\mathbf{h}^{#1}}}
\newcommand{\vspacehack}{\vspace{-.9mm}}
\newcommand{\cindex}{\ensuremath{ c}} %
\newcommand{\findex}{\ensuremath{d}} %
\newcommand{\cmark}{\textcolor{green}{\ding{51}}}%
\newcommand{\xmark}{\textcolor{red}{\ding{55}}}%

\newcommand{\stanfordheader}{\gls{stanfordheader}}
\newcommand{\cubheader}{\gls{cubheader}}
\newcommand{\travelingheader}{\gls{travelingheader}}
\newcommand{\imgnetheader}{\gls{imgnetheader}}

\newglossaryentry{customLoss}
{
name={\ensuremath{\mathcal{L}_{\mathrm{div}}}},
description={Custom Loss für unterschiedliche Features}
}
\newglossaryentry{cLW}
{
name={\ensuremath{\beta}},
description={Gewichtung für customLoss}
}
\newglossaryentry{elaWeight}
{
name={\ensuremath{\lambda}},
description={Gewichtung für elasticNet}
}
\newglossaryentry{elaW}
{
name={\ensuremath{\alpha}},
long = {\ensuremath{\alpha\in[0,1]}},
description={Gewichtung zwischen l1 und l2 für elasticNet}
}
\newglossaryentry{trainDataset}
{
name={\ensuremath{\boldsymbol{D}_t}},
first = {\ensuremath{\boldsymbol{D}_t\in \mathbb{R}^{\gls{nTrainImages}\times 3\times w\times h}}},
description={LokalisierungMaps mit Missingness}
}
\newglossaryentry{nFeatures}
{
name={\ensuremath{n_{f}}},
description={Anzahl der verwendeten Features}
}
\newglossaryentry{nReducedFeatures}
{
name={\ensuremath{\gls{nFeatures}^*}},
description={Anzahl der verwendeten Features im Sparse Decision Layer}
}
\newglossaryentry{outputVector}
{
name={\ensuremath{\boldsymbol{y}}},
long = {\ensuremath{\boldsymbol{y}\in \mathbb{R}^{\gls{nClasses}}}},
description={Finaler Ausgang des Netzes}
}
\newglossaryentry{features}
{
name={\ensuremath{\boldsymbol{f}}},
description={Aus Bild berechnete Features}
}
\newglossaryentry{LocalizationMaps}
{
name={\ensuremath{\boldsymbol{L}}},
long = {\ensuremath{\boldsymbol{L}_p}\in \mathbb{R}^{\gls{nReducedFeatures}\times \frac{w}{p}\times \frac{h}{p}}},
description={LokalisierungMaps mit Missingness}
}
\newglossaryentry{featuresMapwidth}
{
name={\ensuremath{ w_M}},
description={Weite der FeatureMap}
}
\newglossaryentry{featuresMapheigth}
{
name={\ensuremath{ h_M}},
description={Weite der FeatureMap}
}
\newglossaryentry{featureMaps}
{
first ={\ensuremath{\boldsymbol{M} \in \mathbb{R}^{\gls{nFeatures} \times \gls{featuresMapwidth}\times \gls{featuresMapheigth}}}},
name={\ensuremath{m}},
plural={\ensuremath{\boldsymbol{M}}},
description={Aus Bild berechnete Features}
}
\newglossaryentry{featureMapsSmall}
{
first ={\ensuremath{\boldsymbol{M} \in \mathbb{R}^{\gls{nReducedFeatures} \times \gls{featuresMapwidth}\times \gls{featuresMapheigth}}}},
name={\ensuremath{m}},
plural={\ensuremath{\boldsymbol{M}}},
description={Aus Bild berechnete Features}
}
\newglossaryentry{featureMapsTrain}
{
first ={\ensuremath{\boldsymbol{M}_\mathrm{train} \in \mathbb{R}^{\gls{nTrainImages}\times\gls{nFeatures} \times \gls{featuresMapwidth}\times \gls{featuresMapheigth}}}},
name={\ensuremath{m}},
plural={\ensuremath{\boldsymbol{M}}_\mathrm{train}},
description={Aus Bild berechnete Features}
}
\newglossaryentry{trainFeatures}
{
first ={\ensuremath{\boldsymbol{F} \in \mathbb{R}^{\gls{nTrainImages} \times \gls{nFeatures}}}}, %
name={\ensuremath{\boldsymbol{F}^{\mathrm{train}}}},
plural = {\ensuremath{F^{\mathrm{train}}}},
description={Aus Bild berechnete Features}
}
\newglossaryentry{denseNet}
{
first = {DenseNet121~\citep{huang2017densely}},
name ={DenseNet121},
description={Final layer in the neural network}
}
\newglossaryentry{resNet}
{
first = {Resnet50~\citep{he2016deep}},
name ={Resnet50},
description={Final layer in the neural network}
}
\newglossaryentry{incv}
{
first = {Inception-v3~\citep{szegedy2016rethinking}},
name ={Inception-v3},
description={Final layer in the neural network}
}
\newglossaryentry{birdsheader}
{
first = {NABirds~\citep{7298658}},
name = {NABirds},
description={Final layer in the neural network}
}
\newglossaryentry{fgvcheader}
{
first = {FGVC-Aircraft~\citep{FGVCAircraft}},
name={FGVC-Aircraft},
description={Final layer in the neural network}
}
\newglossaryentry{stanfordheader}
{
first = {Stanford Cars~\citep{StanfordCars}},
name={Stanford Cars},
description={Final layer in the neural network}
}
\newglossaryentry{cubheader}
{
first = {CUB-2011~\citep{wah2011caltech}},
long = {CUB-2011~\citep{wah2011caltech}},
name={CUB-2011},
description={Final layer in the neural network}
}
\newglossaryentry{travelingheader}
{
first = {TravelingBirds~\citep{koh2020concept}},
long = {TravelingBirds~\citep{koh2020concept}},
name={TravelingBirds},
description={Final layer in the neural network}
}
\newglossaryentry{decisionLayer}
{
name={decision layer},
description={Final layer in the neural network}
}
\newglossaryentry{fittingLossTarget}
{
name={\ensuremath{\mathcal{L}_{\mathrm{target}}}},
description={Main goal of fitting}
}
\newglossaryentry{layerName}
{
name ={\textit{SLDD-Model}},
first ={\textit{SLDD-Model}~\citep{norrenbrocktake}},
description={The proposed benchmark}
}
\newglossaryentry{NewlayerName}
{
name ={QPM},%
description={The proposed benchmark}
}
\newglossaryentry{denseLayer}
{
name={{dense high-dimensional \gls{decisionLayer}}},
description={The layer that results from training}
}
\newglossaryentry{correlationMatrix}
{
name={\ensuremath{\boldsymbol{Q}}},
first = {\ensuremath{\boldsymbol{Q}\in \mathbb{R}^{\gls{nFeatures}\times\gls{nFeatures}}}},
long = {\ensuremath{q}},
description={Correlation Matrix}
}
\newglossaryentry{featureVector}
{
name={\ensuremath{f}},
long={\ensuremath{\boldsymbol{f}}},
first ={\ensuremath{\boldsymbol{f} \in \mathbb{R}^{\gls{nFeatures}}}},
description={The features of the dense alyer}
}
\newglossaryentry{RedfeatureVector}
{
name={\ensuremath{\boldsymbol{f^*}}},
first ={\ensuremath{\boldsymbol{f^*} \in \mathbb{R}^{\gls{nReducedFeatures}}}},
description={The selected feature Vector}
}
\newglossaryentry{OnlyInteractionVector}
{
name={\ensuremath{\boldsymbol{P}}},
long={\ensuremath{\boldsymbol{P} \in \mathbb{R}^{ \gls{nInteractions}}}},
description={The Interaction Vector}
}
\newglossaryentry{InteractionVector}
{
name={\ensuremath{\boldsymbol{f^*_{\phi}}}},
first ={\ensuremath{\boldsymbol{f^*_{\phi}} \in \mathbb{R}^{\gls{nReducedFeatures} + \gls{nInteractions}}}},
description={The Extended Interaction Vector}
}
\newglossaryentry{nInteractions}
{
name={\ensuremath{n_I}},
description={number of interaction term}
}
\newglossaryentry{dnn}
{
name={\ensuremath{f_\theta(x})},
description={Deep neural network}
}

\newglossaryentry{bias}
{
name={\ensuremath{\boldsymbol{b}}},
long ={\ensuremath{\boldsymbol{b} \in \mathbb{R}^{\gls{nClasses}}}},
description={The bias in the decison layer}
}
\newglossaryentry{classifyFunc}
{
name={\ensuremath{C}},
description={The classifier on the featuers}
}
\newglossaryentry{WeightMatrix}
{
name={\ensuremath{\boldsymbol{W}}},
long = {\ensuremath{\boldsymbol{W}\in \mathbb{R}^{\gls{nClasses}\times \gls{nReducedFeatures} }}},
plural = {\ensuremath{w}},
description={The Weight matrix in the decision layer}
}
\newglossaryentry{qWeightMatrix}
{
name={\ensuremath{\boldsymbol{W}^{Q}}},
first ={\ensuremath{\boldsymbol{W}^{Q}\in \{-\alpha,0,\alpha\}^{\gls{nClasses}\times \gls{nReducedFeatures} }}},
long = {\ensuremath{\boldsymbol{W}^{Q}\in \{-\alpha,0,\alpha\}^{\gls{nClasses}\times \gls{nReducedFeatures} }}},
plural = {\ensuremath{w}},
description={The Weight matrix in the decision layer}
}

\newglossaryentry{nClasses}
{
name={\ensuremath{n_c}},
description={Number of Classes}
}
\newglossaryentry{nTrainImages}
{
name={\ensuremath{n_T}},
description={Number of Train Images}
}
\newglossaryentry{nWeights}
{
name={\ensuremath{n_w}},
description={Number of Entries != 0 in \gls{WeightMatrix}}
}
\newglossaryentry{nperClass}
{
name={\ensuremath{n_{\mathrm{wc}}}},
description={Number of Entries != 0 in \gls{WeightMatrix} per Class}
}
\newglossaryentry{interpTrans}
{
name={\ensuremath{\phi}},
description={Interpretable Transformation}
}
\newglossaryentry{quantizedValue}
{
name={\ensuremath{\alpha}},
description={Interpretable Transformation}
}
\newglossaryentry{targetVector}
{
name={\ensuremath{\hat{\boldsymbol{y}}}},
description={Target Vector in Training}
}
\newglossaryentry{glmsaga}
{
first = {\mbox{\textit{glm-saga}~\citep{wong2021leveraging}}},
long = {\mbox{\textit{glm-saga}~\citep{wong2021leveraging}}},
name={\mbox{\textit{glm-saga}}},
description={Target Vector in Training}
}
\newglossaryentry{cbm}
{
name={CBM},
first ={\textit{Concept Bottleneck Model}~(CBM)~\citep{koh2020concept}},
long = {CBM~\citep{koh2020concept}},
description={Target Vector in Training}
}
\newglossaryentry{ProtoPNet}
{
name={\textit{ProtoPNet}},
first ={\textit{ProtoPNet}~\citep{chen2019looks}},
description={Target Vector in Training}
}
\newglossaryentry{ProtoPShare}
{
name={\textit{ProtoPShare}},
first ={\textit{ProtoPShare}~\citep{rymarczyk2021protopshare}},
description={Target Vector in Training}
}
\definecolor{myGreen}{RGB}{34, 139, 34}
\newglossaryentry{ProtoPool}
{
name={\textit{ProtoPool}},
first ={\textit{ProtoPool}~\citep{rymarczyk2022interpretable}},
long ={\textit{ProtoPool}~\citep{rymarczyk2022interpretable}},
description={Target Vector in Training}
}
\newcommand{\pipnettable}{PIP-Net} %
\newcommand{\checktext}[1]{#1}
\newcommand{\protopooltable}{ProtoPool} 
\newcommand{\qsenntable}{Q-SENN}
\newcommand{\slddtable}{SLDD-Model} %
\newglossaryentry{protopnet}
{
name={\textit{ProtoPNet}},
first ={\textit{ProtoPNet}~\citep{chen2019looks}},
description={Target Vector in Training}
}
\newglossaryentry{qsenn}
{
name={\textit{Q-SENN}},
first ={\textit{Q-SENN}~\citep{norrenbrock2024q}},
description={Target Vector in Training}
}
\newglossaryentry{PIP-Net}
{
name={\textit{PIP-Net}},
first ={\textit{PIP-Net}~\citep{nauta2023pipnet}},
description={Target Vector in Training}
}

\newglossaryentry{imgnetheader}
{
first = {ImageNet-1K~\citep{imagenet15russakovsky}},
long = {ImageNet-1K~\citep{imagenet15russakovsky}},
name={ImageNet-1K},
description={Final layer in the neural network}
}
\newglossaryentry{labelfreecbm}
{
name={\textit{Label-free CBM}},
first ={\textit{Label-free CBM}~\citep{oikarinen2023label}},
description={Target Vector in Training}
}
\newglossaryentry{ProtoTree}
{
name={\textit{Prototree}},
first ={\textit{ProtoTree}~\citep{nauta2021neural}},
description={Target Vector in Training}
}
\newglossaryentry{ImageSample}
{
name={\ensuremath{\boldsymbol{I}}},
first ={\ensuremath{\boldsymbol{I} \in \mathbb{R}^{3\times w\times h}}},
description={The classifier on the featuers}
}

\newcommand{\suppl}{Suppl.}

\newcommand{\incv}{\gls{incv}}
\newcommand{\resnet}{\gls{resNet}}

\newcommand{\attributeset}[1]{\ensuremath{\rho_{#1}}}

\newcommand{\oldloc}[1]{\textrm{diversity@#1}}
\newcommand{\loc}[1]{\textrm{SID@#1}}
\newcommand{\glm}{\gls{glmsaga}}
\newcommand{\glmtable}{glm-saga\textsubscript{5}}%

\newcommand{\eg}{\textit{e}.\textit{g}. }

\newcommand{\st}{so that}

\newcommand{\gtmatrix}{\ensuremath{A}^{\mathrm{gt}}}

\newcommand{\arrowDown}{\ensuremath{\boldsymbol{\downarrow}}}

\newcommand{\arrowUp}{\ensuremath{\boldsymbol{\uparrow}}} %

\tikzstyle{process} = [rectangle, minimum width=1cm, minimum height=1cm, text centered, text width=1.8cm, draw=black]
\tikzstyle{decision} = [diamond, minimum width=1cm, minimum height=1cm, text centered, draw=black]
\tikzstyle{arrow} = [thick,->,>=stealth]
\newcommand{\distEdge}{\ensuremath{d_e}}
\newcommand{\ClassSim}{\Psi}
\newcommand{\classSim}{\psi}
\newcommand{\simmat}{\ensuremath{\boldsymbol{A}}} 
\newcommand{\insimmat}[2]{\ensuremath{a_{#1,#2}}}
\newcommand{\infvec}[1]{\ensuremath{s_{#1}}}
\newcommand{\symbolFeaFea}{r}
\newcommand{\FSimeaturemat}{\ensuremath{\boldsymbol{R}}}
\newcommand{\BSimeaturemat}{\ensuremath{\boldsymbol{b}}} 
\newcommand{\gurobi}{\textit{Gurobi}~\citep{gurobi}}
\newcommand{\objective}{\ensuremath{Z}}
\newcommand{\fvecgurobi}{\ensuremath{\boldsymbol{s}}}%
\newcommand{\wgurobi}{\ensuremath{\boldsymbol{W}}}

\newcommand{\simieq}{\ensuremath{\sum_{\cindex=1}^{\gls{nClasses}}(\boldsymbol{a}_\cindex\circ \boldsymbol{w}_\cindex)^T \fvecgurobi}}
\newcommand{\correq}{\ensuremath{\fvecgurobi^T \FSimeaturemat \fvecgurobi}}
\newcommand{\biaseq}{\ensuremath{\BSimeaturemat{}^T \fvecgurobi}}

\newcommand{\generality}{\ifmmode
    \mathrm{Class\text{-}Independence}
  \else
    Class-Independence
  \fi}
\newcommand{\contrastiveness}{Contrastiveness}
\newcommand{\cubsim}{Structural Grounding}
\newcommand{\correlation}{Correlation}

\newcommand{\boldnessstatement}{Among more interpretable models, the best result is marked in bold, second best underlined.}

\newcommand{\zerovec}[1]{\ensuremath{\boldsymbol{0}^{#1}}}
\newcommand{\onevec}[1]{\ensuremath{\boldsymbol{1}^{#1}}}
\newcommand{\zeromat}[2]{\ensuremath{\boldsymbol{0}^{#1, #2}}}

\newcommand{\interpmetricstablestart}[1]{Comparison on Interpretability metrics with #1. Due to required annotations, \cubsim{} can only be computed for \travelingheader{} and \cubheader{}. 
 \boldnessstatement}
\newcommand{\accmetricstablestart}[1]{Comparison on compactness and accuracy with #1: \gls{NewlayerName} shows increased accuracy and compactness.}

\author{Thomas Norrenbrock, Timo Kaiser \& Bodo Rosenhahn  \\
 Institute for Information Processing (tnt)\\
  L3S - Leibniz Universität Hannover, Germany \\
\texttt{\{norrenbr,kaiser,rosenhahn\}@tnt.uni-hannover.de} \\
\And
Sovan Biswas \& Ramesh Manuvinakurike \\
Intel  Labs, USA  \\
\texttt{\{sovan.biswas,ramesh.manuvinakurike\}@intel.com} \\
}

\iclrfinalcopy %

\begin{document}

\maketitle
\begin{abstract}
Understanding the classifications of deep neural networks, \eg{}used in safety-critical situations, is becoming increasingly important.
While recent models can locally explain a single decision, to provide a faithful global explanation about an accurate model's general behavior is a more challenging open task.
Towards that goal, we introduce the 
Quadratic Programming Enhanced Model (\gls{NewlayerName}), which learns globally interpretable class representations.
\gls{NewlayerName} represents every class with a binary assignment of very few, typically $5$, features, that are also assigned to other classes, ensuring easily comparable contrastive class representations.
This compact binary assignment is found using discrete optimization  based on predefined similarity measures and interpretability constraints.
The resulting optimal assignment is used to fine-tune the diverse features, \st{} each of them becomes the shared general concept between the assigned classes.
Extensive evaluations show that \gls{NewlayerName} delivers unprecedented global interpretability across small and large-scale datasets while setting the state of the art for the accuracy of interpretable models. 

\end{abstract}

\begin{figure*}[h]
  \centering
  \includegraphics[width=\linewidth]{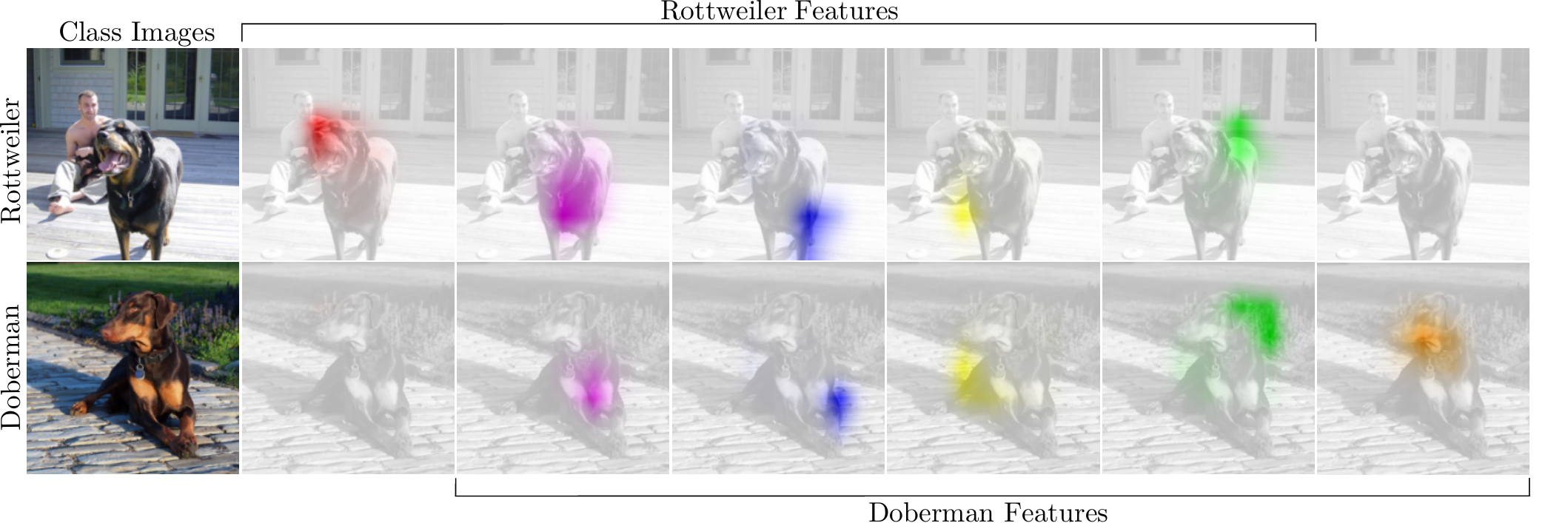}
   \vspace{-.5cm}\caption{Faithful global interpretability of our \gls{NewlayerName}: %
   Without any additional supervision, 
   \gls{NewlayerName} learns to represent 
   Rottweiler and Doberman using  $5$ diverse and general features.
   \gls{NewlayerName} faithfully explains that it differentiates them 
   exclusively 
   via their visibly distinct
   head.}
   \vspace{-.1cm}
  \label{fig:metrics_full}
\end{figure*}
\newcommand{\featurevert}{.15cm}
\definecolor{mydarkgreen}{rgb}{0,0.5,0}
\begin{figure}
\centering
\resizebox{.9\linewidth}{!}{
\begin{tikzpicture}[node distance=.5cm]
           \node(feature_labelU) at (0, 0)[text width=1.3cm ] {Features};  
         \node[inner sep=0pt,below=.1cm of feature_labelU] (left1U) {\includegraphics[width=1.5cm]{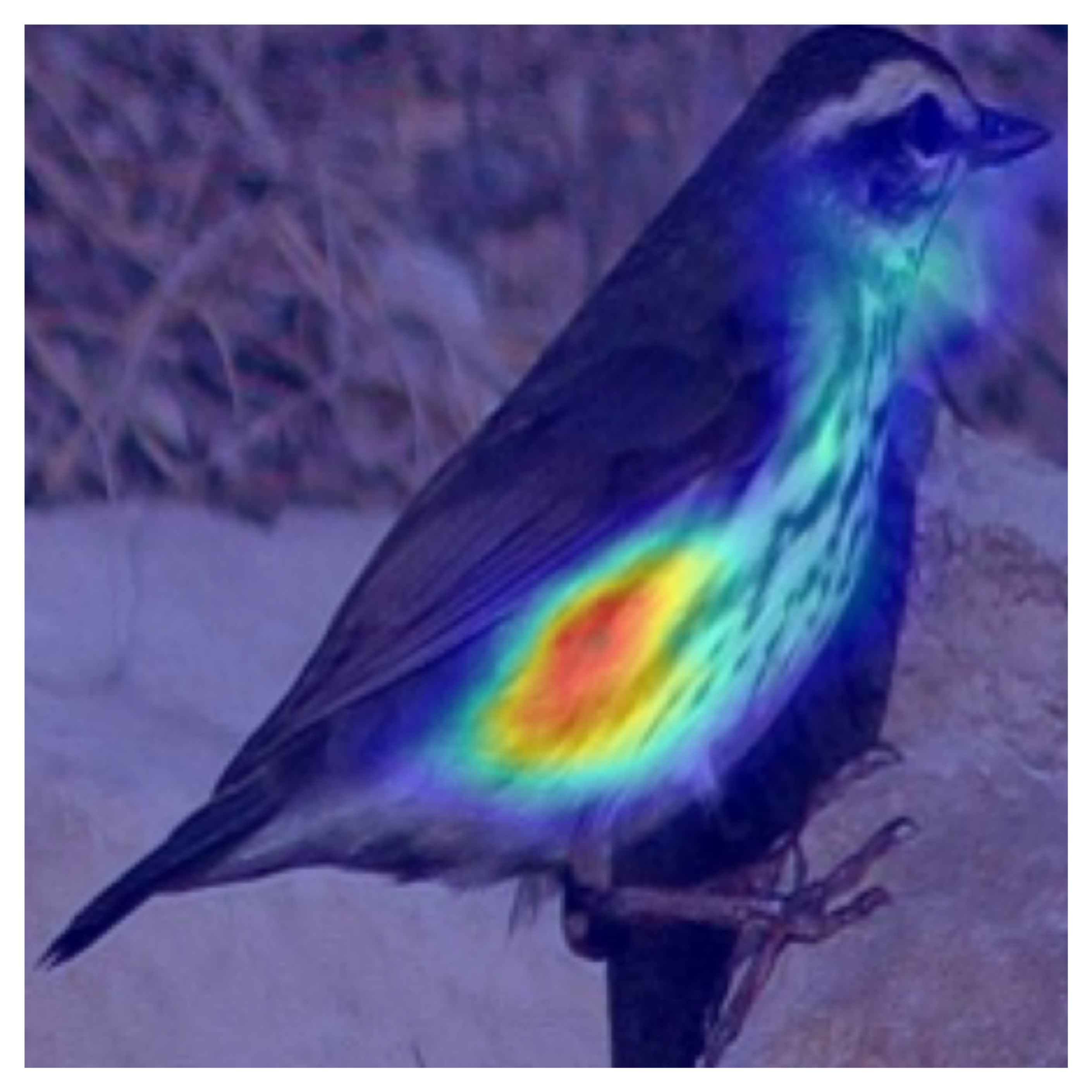}};
         \node[inner sep=0pt, below=\featurevert of left1U] (left4U) {\includegraphics[width=1.5cm]{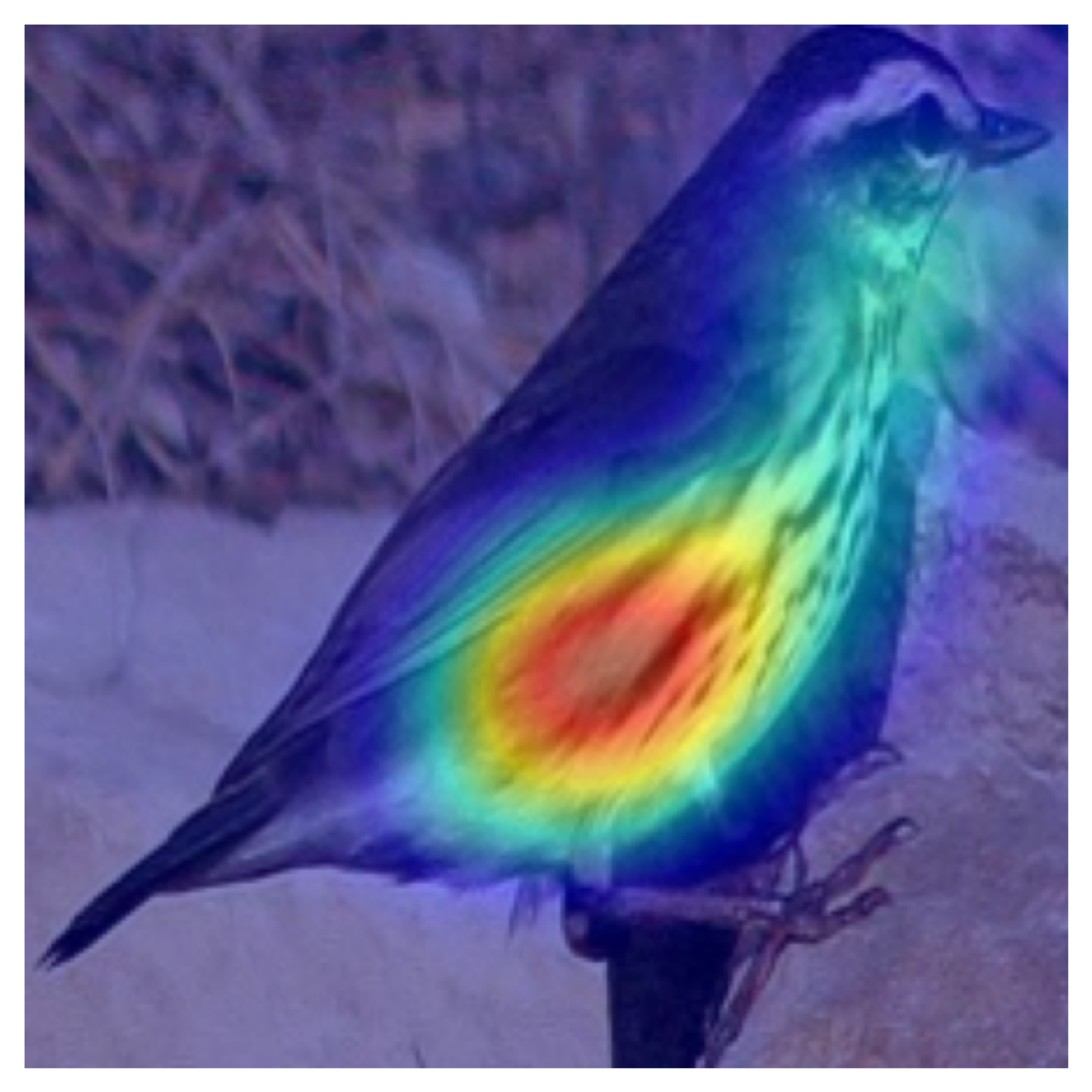}};
          \node[inner sep=0pt, below=\featurevert of left4U] (left3U) {\includegraphics[width=1.5cm]{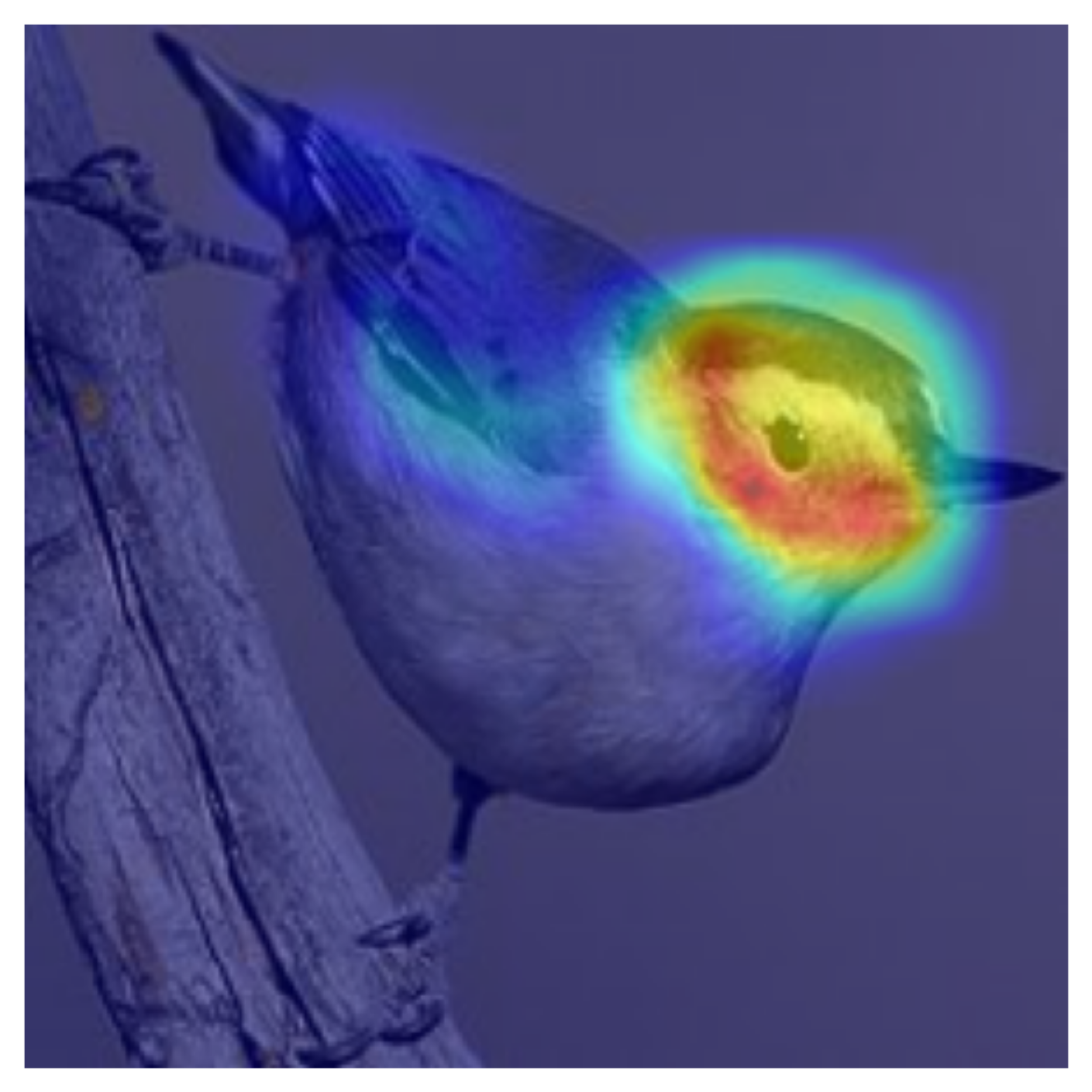}};
        \node[inner sep=0pt,below=\featurevert of left3U] (left2U) {\includegraphics[width=1.5cm]{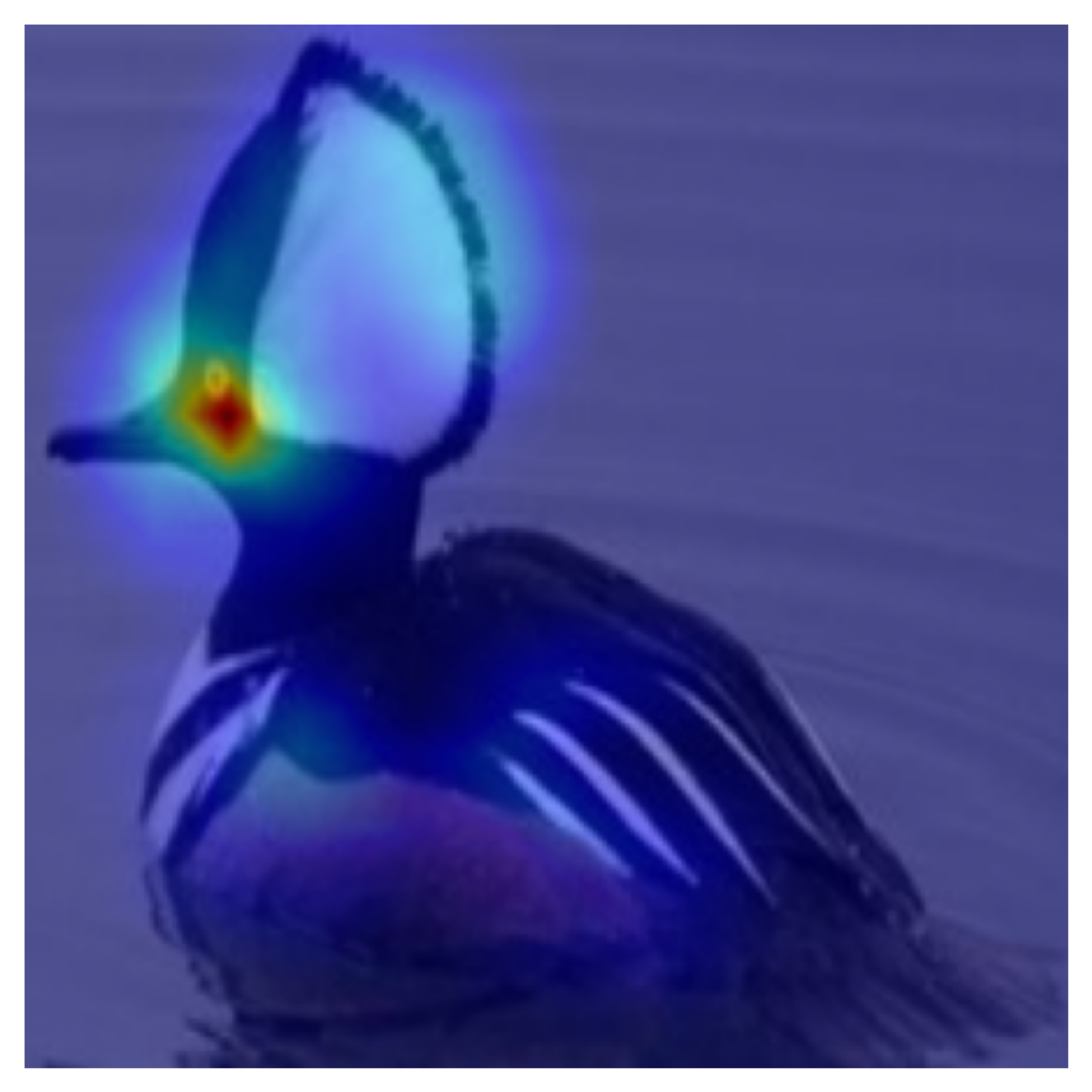}};

        \node[inner sep=0pt, below right=0cm and 1.3cm of left1U] (right1U) {\includegraphics[width=1.5cm]{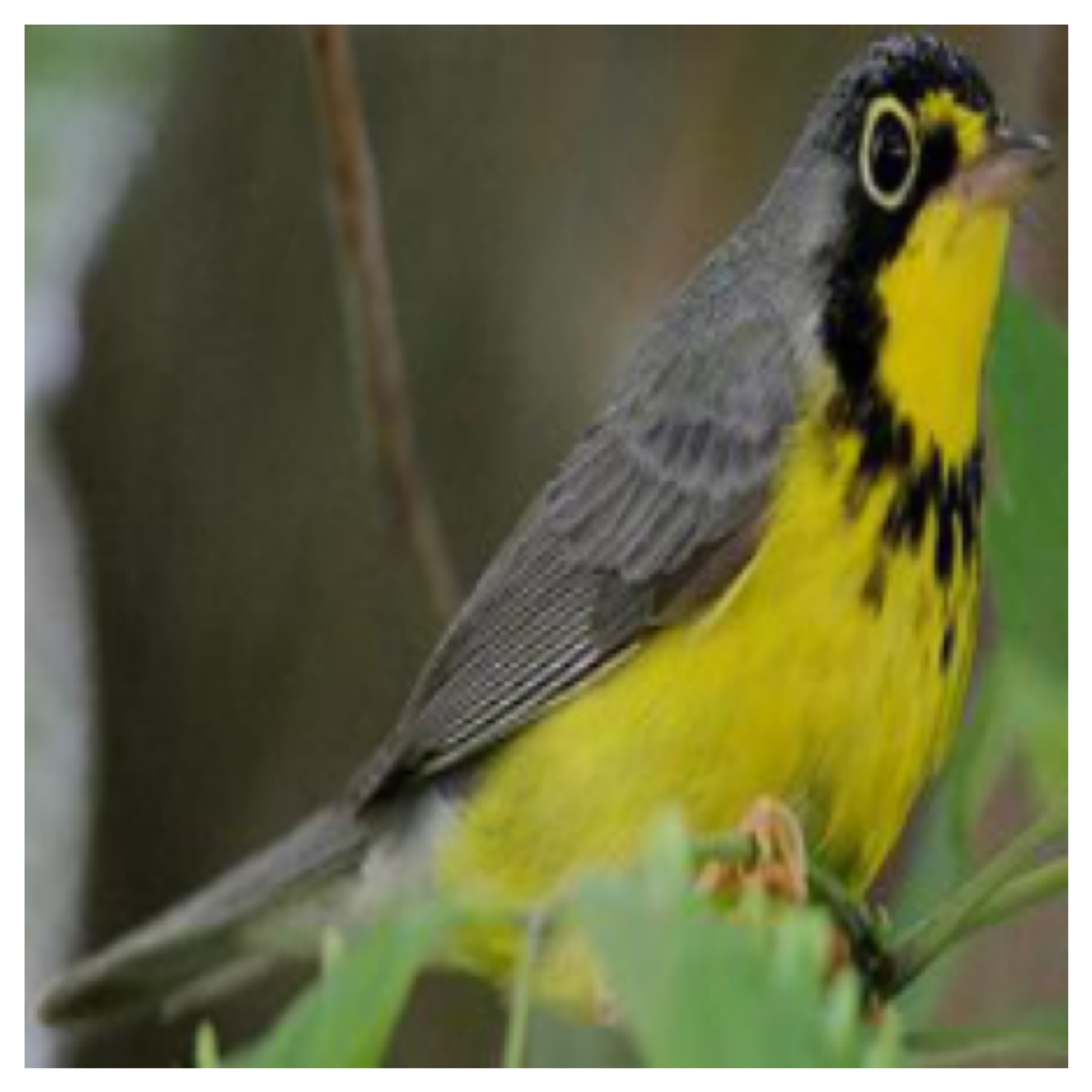}};
        \node[inner sep=0pt, below=of right1U] (right2U) {\includegraphics[width=1.5cm]{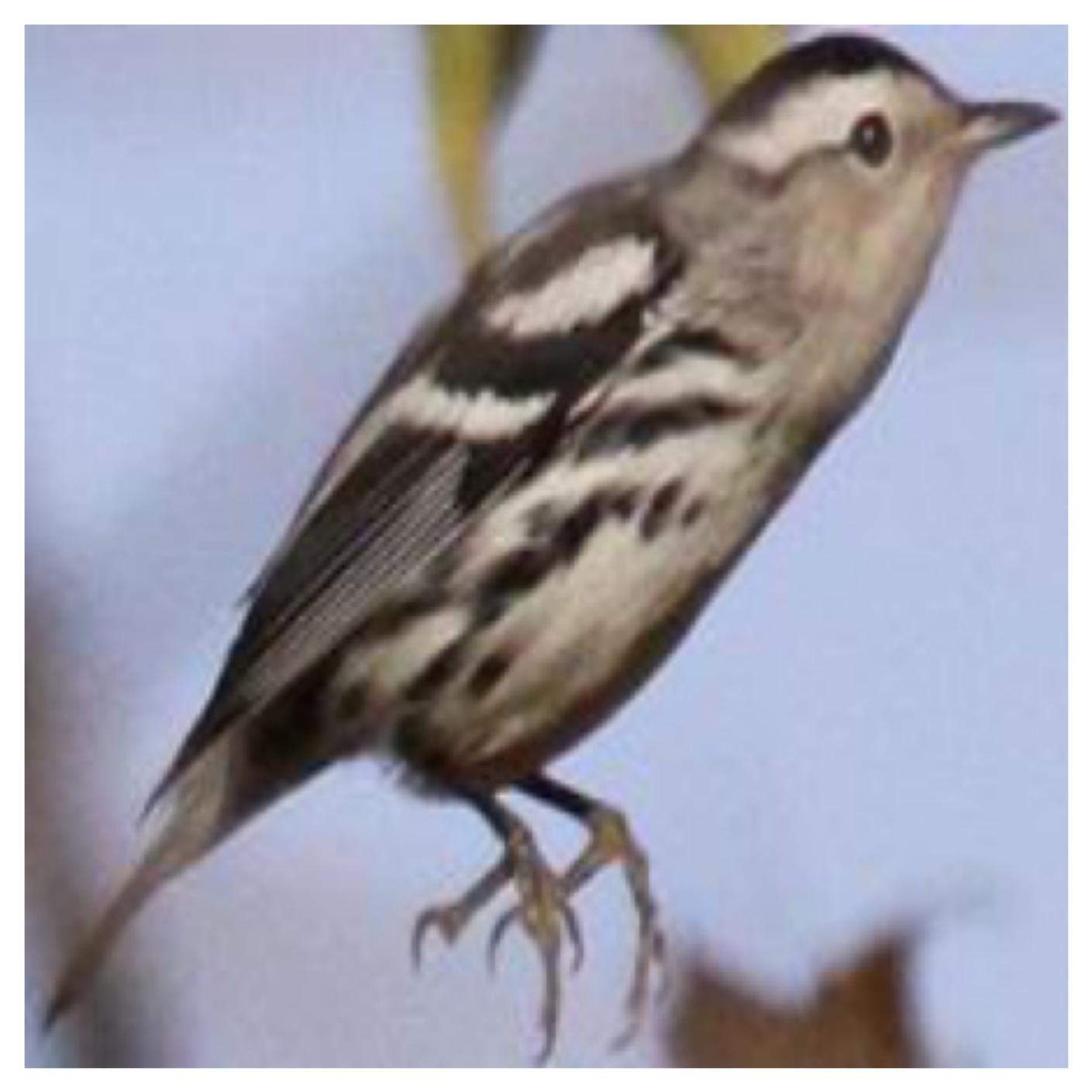}};
        \node (class_labelU) at (feature_labelU -| right1U){Classes};
        
        \draw[-, line width=1.4pt, mydarkgreen, line cap=round] (left1U.east) -- (right1U.west)  node[pos=.45, sloped, above] {\footnotesize1.4};
         \draw[-, line width=1.9pt, mydarkgreen, line cap=round] (left1U.east) -- (right2U.west) node[pos=0.3, sloped, below] {\footnotesize1.9};
        \draw[-, line width=1.9pt, mydarkgreen, line cap=round] (left2U.east) -- (right1U.west) node[near start, sloped, above] {\footnotesize1.5};
         \draw[-, line width=2pt, red, out=0, in=180, line cap=round] (left2U.east) -- (right2U.west) node[pos=.45, sloped, below] {\footnotesize-2};
        \draw[-, line width=.5pt, mydarkgreen, line cap=round] (left3U.east) -- (right1U.west) node[pos=0.35, sloped, above] {\footnotesize{0.2}};
        \draw[-, line width=1.7pt, mydarkgreen, line cap=round] (left3U.east) -- (right2U.west) node[pos=.33,  above] {\footnotesize1.7};
         \draw[-, line width=2.1pt, mydarkgreen, line cap=round] (left4U.east) -- (right1U.west) node[pos=0.1,  left] {\footnotesize2.1};
        \draw[-, line width=1.8pt, mydarkgreen, line cap=round] (left4U.east) -- (right2U.west) node[pos=0.25,sloped,  above] {\footnotesize1.8};
  \node(feature_label) at (9, 0)[,text width=1.3cm] {Features};  
      \node[inner sep=0pt,below=.1cm of feature_label] (left1) {\includegraphics[width=1.5cm]{QpImgs2/0_Color}};
      \node[inner sep=0pt, below=\featurevert of left1] (left4) {\includegraphics[width=1.5cm]{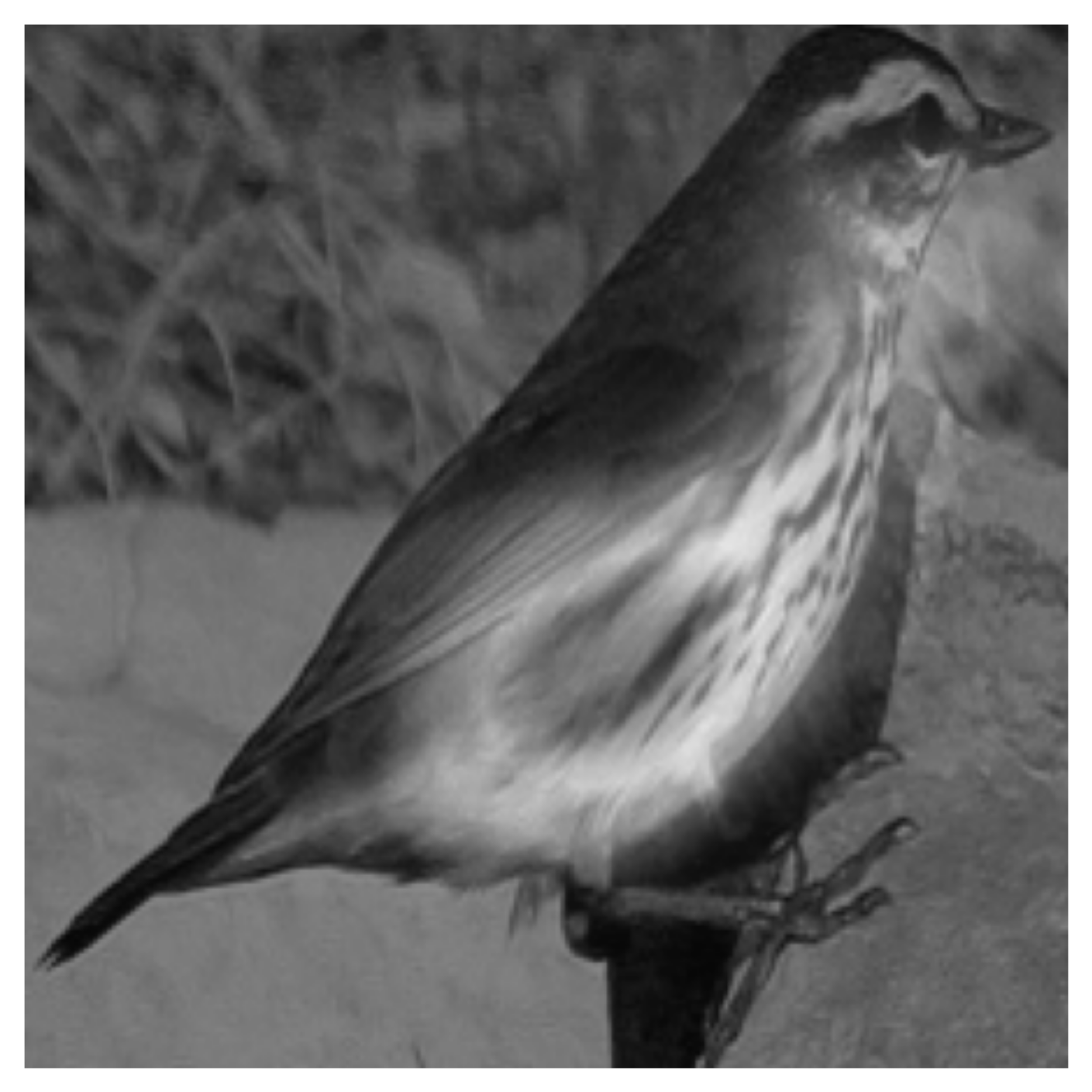}};
      \node[below=-0.16 cm of left4, font=\scriptsize] {(deselected)};

              \node[inner sep=0pt, below=\featurevert of left4] (left3) {\includegraphics[width=1.5cm]{QpImgs2/34_Color}};
        \node[inner sep=0pt,below=\featurevert of left3] (left2) {\includegraphics[width=1.5cm]{QpImgs2/35_Color}};
        \node[inner sep=0pt, below right=0cm and 1.3 cm of left1] (right1) {\includegraphics[width=1.5cm]{QpImgs2/Class161_4828}};
        \node[inner sep=0pt, below=of right1] (right2) {\includegraphics[width=1.5cm]{QpImgs2/Class158_4763}};
          \node (class_label) at (feature_label -| right1) {Classes};
        \draw[-, line width=2pt, mydarkgreen, line cap=round] (left1.east) -- (right1.west)   node[pos=.45, sloped, above] {\footnotesize1} ;
         \draw[-, line width=2pt, mydarkgreen, line cap=round] (left1.east) -- (right2.west)  node[pos=0.2, sloped, below] {\footnotesize1} ;
 \draw[-, line width=2pt, mydarkgreen, line cap=round] (left2.east) -- (right1.west)  node[near start, sloped, above] {\footnotesize1} ;
        \draw[-, line width=2pt, mydarkgreen, line cap=round] (left3.east) -- (right2.west)node[pos=.2,  above] {\footnotesize1} ;

          \node [fit=(feature_labelU)(right2U.east)(class_labelU)(left2U.south), draw, inner sep=1pt] (dense_box) {};

        \node [above=of dense_box.south,xshift=0.8cm, anchor=center] {\large\textbf{Dense}};
         \node [fit=(feature_label)(right2.east)(class_label)(left2.south), draw, inner sep=1pt, yshift=-0.0cm] (qpm_box) {};

         \node [above=of qpm_box.south, anchor=center, xshift=0.8cm] {\large\textbf{\gls{NewlayerName}}};
         
        \draw[->, line width=2pt] ([yshift=-.2cm] dense_box.east) -- node[midway, above] {$\text{Maximize:}  \frac{1}{2} \mathbf{x}^T \mathbf{Q} \mathbf{x} + \mathbf{c}^T \mathbf{x}$} node[midway, below=.15cm] {$\mathbf{x}$ = $\begin{bmatrix}
\mathrm{Feature\text{ }Selection} \\
\mathrm{Sparse\text{ }Assignment}\\
\end{bmatrix}$} ([yshift=-.2cm] qpm_box.west);

    \end{tikzpicture}
    }
    \caption{Exemplary Application of the QP to a dense model with just 4 features and 2 classes with the aim of selecting 3 features and assigning 2 per class. 
    The different weights are indicated by thickness of connection and color indicates sign.
    The result is a binary assignment of selected features to classes. 
    Typical values for \resnet{} on \cubheader{}  are selecting 50 features out of 2048 and assigning 5 to each of the 200 classes.}
         \label{fig:QP}
         \vspace{-.5cm}
\end{figure}

\section{Introduction}
\label{sec:intro}
Deep Learning has made remarkable advances in various fields, such as image classification, segmentation or generation~\citep{krizhevsky2012imagenet,kirillov2023segment,rombach2021highresolution,ramesh2022hierarchical}.
For high-stakes decisions, \eg{}applying image classification in the medical domain, legislation moves towards requiring a certain level of interpretability~\citep{veale2021demystifying}, whose measurement is a fairly open task on its own.
However, some desirable and measurable qualities of explanations have been identified~\citep{miller2019explanation}. 
Human-friendly explanations  should be contrastive~\citep{lipton1990contrastive}, diverse~\citep{alvarez2018towards}, general and compact~\citep{read1993explanatory}.
As humans can consider $7\pm2$ cognitive aspects at once~\citep{miller1956magical}, an explanation size of up to $5$ is desirable.
Additionally, an explanation should faithfully explain the model, which is where many post-hoc methods fail~\citep{kindermans2019reliability, adebayo2018sanity,daras2022discovering}.
Therefore, we focus on models that are interpretable by design with built-in faithful explanations.\\
Previous works, such as SENN~\citep{alvarez2018towards}, \gls{qsenn}, \gls{cbm}, \gls{labelfreecbm}, \gls{PIP-Net}, \gls{ProtoPool}, \gls{ProtoTree},  \gls{protopnet} or the \gls{layerName} rely on combining understandable features in an interpretable manner.
However, while most models can offer convincing \textit{local} explanations for a single decision, they struggle with the \textit{global} explanation of their behavior in general.
Some models with global interpretability do not show competitive accuracy~\citep{oikarinen2023label, koh2020concept} and it is debated~\citep{molnar2020interpretable}, if ensembles of very deep decision tress~\citep{nauta2021neural} or dense high-dimensional linear layers~\citep{koh2020concept, rymarczyk2022interpretable, alvarez2018towards} are truly intrinsically interpretable as they lack desired qualities like compactness.
For that reason \gls{PIP-Net} focuses on learning sparse class representations.
These representations lie in a high dimensional feature space, which causes \gls{PIP-Net}'s features to be connected to very few or only one class each.
This leads to the emergence of features that are already detecting the class and no general 
concept.
The sparse representations of \gls{PIP-Net} thus have no interpretable meaning, as classes are represented with themselves.
To alleviate that issue, the \gls{layerName} and \gls{qsenn} reduce both dimensions of compactness: 
They not only reduce the number of features per class \gls{nperClass}, which in isolation leads to class-specific features  but also the number of features in total \gls{nReducedFeatures} to be significantly below the number of classes \gls{nClasses}.
That causes each of the fewer features to be assigned to multiple classes, which prevents the emergence of class detectors.
However, these models still have shortcomings when it comes to global interpretability. 
Their class representations are real-valued, or ternary for \gls{qsenn}, include a bias, and are composed of a varying number of features.
Therefore, the global class explanations are hardly comparable or contrastive.\\
In this work, we introduce the Quadratic Programming Enhanced Model (\gls{NewlayerName}) that offers interpretable class representations and sets a new state of the art for the accuracy of
compactness-based 
interpretable models.
It represents every class with the binary assignment of a low user defined number of features \gls{nperClass}, which themselves are contrastive, general and diverse.
We typically choose $5$, in line with previous work~\citep{norrenbrock2024q, norrenbrocktake}, to accommodate for human limitations~\citep{miller1956magical}.
As shown in~\cref{fig:metrics_full}, \gls{NewlayerName} offers built-in faithful global explanations for classes and enables the intuitive comparison of different learned class representations.
These easy comparisons between compact binary class representations even enable reasoning about the differentiating feature between the classes, like the head in~\cref{fig:metrics_full}.
The improvements in faithful global interpretability of class representations are summarized in \cref{tab:ClassRepresent}.\\
The crucial step in training a \gls{NewlayerName} is
solving a binary QP, applied to a dense black-box model, which jointly finds an optimal solution to both the selection of a reduced subset of the model's features and the sparse assignment between the features and classes, as shown in \cref{fig:QP}.
It maximizes the similarity between features and their assigned classes, while minimizing the similarity of jointly selected features.
Further, the linear term can steer the selection towards desired biases, while the desired interpretability is incorporated via constraints.
This optimal solution is then fixed for the following fine-tuning during which the features adapt to their assigned classes.
As every class is assigned to the same number of features, each of the features detects shared general concepts between its assigned classes instead of also detecting the entire class. This leads to state-of-the-art accuracy.
Finally, the assignments are not maximizing inter-class distance, resulting in more similar representations for similar classes and a form of structural grounding.
Code: \url{https://github.com/ThomasNorr/QPM}

\begin{table} \centering 
\caption{ Properties of class representation for class $i$, $y_i = \mathbf{w_i}\mathbf{f} + b_i$, for \cubheader: 
Only \gls{NewlayerName} represents each of its classes with the binary assignment of a fixed number of general features (quantified in~\cref{tab:Interpproto-table}) and no class Bias.
Therefore, classes can also be represented as set of $5$ feature indices $S_i$, $y_i = \sum_{j\in S_i}f_j$. These contrastive class explanations enable faithful global interpretability. If applicable, all methods are configured to $\gls{nperClass} = 5$ and $\gls{nReducedFeatures}=50$.} 
\label{tab:ClassRepresent}
\resizebox{\linewidth}{!}{
\begin{tabular}{l|ccc|c} 
\hline Method & Size of $\mathbf{w_i}$ & Equal Class Sparsity & No Class Bias & Contrastive Representation \\ 
\hline Baseline \resnet{} & $\mathbf{w_i}\in\mathbb{R}^{2048}$ & \cmark & \xmark & \xmark  \\ 
\glmtable{} & $\mathbf{w_i}\in\mathbb{R}^{809}$ & \xmark & \xmark & \xmark  \\ 
\pipnettable{} & $\mathbf{w_i}\in\mathbb{R}^{731}$ & \xmark & \cmark & \xmark  \\
\protopooltable{} & $\mathbf{w_i}\in\mathbb{R}^{202}$ & \cmark & \cmark & \xmark  \\ \slddtable{} & $\mathbf{w_i}\in\mathbb{R}^{50}$ & \xmark & \xmark & \xmark  \\
\qsenntable{} & $\mathbf{w_i}\in\{-\alpha, 0, \alpha\}^{50}$ & \xmark & \xmark & \xmark  \\ 
\hline \gls{NewlayerName} (Ours) & $\mathbf{w_i}\in\{0,1\}^{50}$ & \cmark & \cmark &$S_i\in\{1, \dots, 50\}^5$ \quad \cmark \\ 
\hline 
\end{tabular} 
}

\end{table}

Our main \textbf{contributions} are as follows:
\begin{itemize}
\item We propose the Quadratic Programming Enhanced Model (\gls{NewlayerName}), which incorporates an optimal feature selection and their binary assignment of a few, \eg{} 5 features per class.
It is found by formulating the quadratic problem and solving it optimally.

\item We demonstrate improvements in accuracy, compactness and structural grounding of \gls{NewlayerName} on multiple benchmark datasets and architectures for image classification, including \gls{imgnetheader}. 
Due to optimally using the given capacity, \gls{NewlayerName} sets the new state of the art for 
compactness-based globally 
interpretable models.

\item We show that the learned features exhibit several desired quantifiable properties, such as contrastiveness, generality and diversity, and can be steered towards user-defined criteria.
\item Representing classes as a contrastable compact set of these general features makes \gls{NewlayerName} faithfully globally interpretable, while further closing the accuracy gap to black-box models.

\end{itemize}

\section{Related Work}
\label{sec:RelatedWork}

Research towards Interpretable machine learning includes the direct design of models providing interpretability by themselves \citep{alvarez2018towards, sawada2022concept, norrenbrocktake, nauta2023pipnet, nauta2021neural, rymarczyk2022interpretable,zarlenga2022concept, marconato2022glancenets, koh2020concept, rymarczyk2021protopshare,chen2019looks}  or to find post-hoc methods which aim to explain the decision process or single features of the model ~\citep{kim2018interpretability,bau2017network,AlphaZero,Fel_2023_CVPR,yuksekgonul2022posthoc,pmlr-v202-kalibhat23a,oikarinen2023clipdissect}.
As our method is designed to find a compact set of human-understandable features, our work can be assigned to the former type, which we focus on within this section.
However, the alignment of the learned features of our proposed \gls{NewlayerName} with human attributes can be guided by the post-hoc methods. 
When considering the interpretability of a model, a distinction is made between local interpretability, which refers to the explanation of a single decision, and global interpretability, which describes the holistic behavior of the model over the entirety of a dataset~\citep{molnar2020interpretable}.
For local interpretability, \textit{B-Cos Networks}~\citep{bohle2023holistically} already offer faithful explanations in the form of saliency maps.
Therefore, this work focuses on the more challenging global interpretability, which also improves local interpretability.
In the social sciences~\citep{miller2019explanation}, human-friendly explanations are contrastive~\citep{lipton1990contrastive}, concise and general~\citep{read1993explanatory}. 
Further, SENN~\citep{alvarez2018towards} describes diversity and grounding as desirable attributes for features of an interpretable model.
Grounding refers to the alignability with any human concept and is very difficult to quantify, as one would need a full dataset of potentially learned concepts. 
Problematically, deep neural networks typically exhibit superposition and polysemantic neurons~\citep{scherlis2022polysemanticity, elhage2022toy, templeton2024scaling}, which is why we focus on more clearly quantifiable aspects in this work.\\
Models such as \gls{ProtoTree}~\citep{nauta2021neural}, \gls{ProtoPNet}, \gls{ProtoPShare}, \gls{ProtoPool}, and \gls{PIP-Net} aim to learn prototypes from data by employing deep feature extractors.
These prototypes' similarities are subsequently integrated into interpretable models.
However, the extent of their interpretability remains debatable, as \citet{kim2021hive} and \citet{hoffmann2021looks} reveal a gap between human and computed similarities.
Similar to this work, \gls{PIP-Net} also aims for compactness via sparse weights in the final decision layer.
However, they apply a local optimization that aims for sparsity solely, resulting in a big set of used features with many of them being class-specific.
\citet{norrenbrocktake, norrenbrock2024q} additionally select a compact feature set for their \textit{SLDD-Model} and \gls{qsenn}, where a class is to be related to only a few features.
Their diversity is ensured through
the Feature Diversity Loss~\gls{customLoss}, which incurs a higher cost when highly activated and weighted features localize on the same region. 
For both feature selection and the computation of the sparse layer, glm-saga~\citep{wong2021leveraging} is used. 
It locally and iteratively optimizes the problem, leading to a suboptimal feature selection and continuous weights.
In contrast, our global optimization with user-defined steerable criteria jointly finds an optimal selection
of the required number of features and computes their binary assignments.
This leads to a more effective use of the allocated capacity and built-in
easily interpretable class representations for global interpretability.
Another line of research is based on the \textit{Concept Bottleneck Model} (\gls{cbm}) which initially predicts the labeled concepts within a given dataset and subsequently leverages a basic model to predict the target category based on these identified concepts.
This approach remains an area of active exploration and development~\citep{sawada2022concept,zarlenga2022concept, marconato2022glancenets, oikarinen2023label}, but is limited by the annotations, or in case of the \gls{labelfreecbm} by the vision-language model, resulting in subpar accuracy and compactness.
Finally, 
\citet{rosenhahn2023optimization} applies discrete optimization to obtain sparse neural networks~\citep{GlaKai2023a}.

\section{Method}
\vspacehack{}

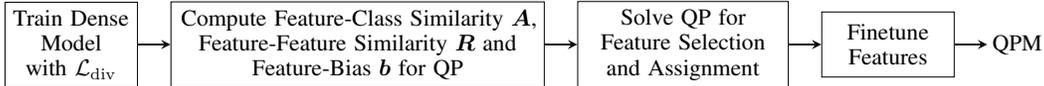
\begin{figure*}[t]
    \centering
    \resizebox{\linewidth}{!}{\begin{tikzpicture}[node distance=.01cm]
        \node (denseTraining) [process] {Train Dense Model with \gls{customLoss}};
          \node (ComputeMatrics) [process, right=0.5cm of denseTraining, text width=5.5cm] {Compute Feature-Class Similarity \simmat{},\\Feature-Feature Similarity \FSimeaturemat{} and\\Feature-Bias \BSimeaturemat{} for QP};
        \node (featureSelAssign) [process, right=0.5cm of ComputeMatrics, text width=3cm] {Solve QP for\\Feature Selection \\and Assignment};
        \node (finetune) [process, right=0.5cm of featureSelAssign] {Finetune Features};
        \draw [arrow] (denseTraining) -- (ComputeMatrics);
        \draw [arrow] (ComputeMatrics) -- (featureSelAssign);%

        \draw [arrow] (featureSelAssign) -- (finetune);
        \node(end) [right=0.5cm of finetune] {};
        \draw [arrow] (finetune) -- (end) node[midway, right, xshift=.2cm] {\gls{NewlayerName}};
    \end{tikzpicture}
    }
    \caption{Overview of our proposed pipeline to construct a \gls{NewlayerName}}
    \label{fig:OverviewAppraoch}
    \vspace{-.7cm}
\end{figure*}

Our proposed \gls{NewlayerName} is designed for the interpretable classification of an image %
as a class $c \in \{c_1,c_2, \dots, c_{\gls{nClasses}}\}$.
The \gls{NewlayerName} uses a deep feature extractor $\Phi$ to compute feature maps \Gls{featureMapsSmall} of width $\gls{featuresMapwidth}$ and height $\gls{featuresMapheigth}$ and averages them into a feature vector \gls{RedfeatureVector}.
The classification result \glsentrylong{outputVector} of the \gls{NewlayerName} is the matrix multiplication between the sparse binary matrix $\wgurobi^*~\in\{0,1\}^{\gls{nClasses}\times \gls{nReducedFeatures} }$ %
and the features
\gls{RedfeatureVector} formalized as $ \gls{outputVector} = \wgurobi^*\gls{RedfeatureVector}$. \\
The pipeline of our proposed method is shown in \cref{fig:OverviewAppraoch} and is motivated by \citep{norrenbrocktake, norrenbrock2024q}, following their presentation and notation.
It starts with training a conventional black-box model with initially \gls{nFeatures} features using the feature diversity loss~\gls{customLoss}~\citep{norrenbrocktake}, as a high diversity of features is desired for interpretable models.
A detailed explanation of~\gls{customLoss} is included in \cref{suppsec:ldiv}.
Using the black-box model as starting point, we aim to find a selection of \gls{nReducedFeatures} out of the initial \gls{nFeatures} features and their sparse binary assignment $\boldsymbol{W}^*$ to the classes to enable downstream interpretability.
The feature extractor $\Phi$ is then fine-tuned with this solution fixed, so that the features adapt to the sparse solution and become a shared concept of the assigned classes.
This is encouraged through selecting fewer features than there are classes, $\gls{nReducedFeatures} < \gls{nClasses}$, and representing every class with the same number \gls{nperClass}, typically $5$, of features. 
Using the same number of features for every class is beneficial for the interpretability in multiple ways. 
The class representations do not need a bias and can be contrasted as $S_{i}\in\{1, \dots, \gls{nReducedFeatures}\}^{\gls{nperClass}}$, while the composing features can focus on detecting general concepts.
Since we aim to optimize binary variables under constraints with a clear objective, we can formulate it as a discrete optimization problem to get the optimal solution. 
As indicated in~\cref{fig:QP}, we define the constants \simmat{}, \FSimeaturemat{} and \BSimeaturemat{} of the resulting QP so that
in the global optimum different (\FSimeaturemat{}) , localized (\BSimeaturemat{}) features are selected and assigned to classes for which they have high predictive power (\simmat{}).
These fixed simple binary class representations then lead to the emergence of interpretable features during fine-tuning.
How the 
quadratic problem with \simmat{}, \FSimeaturemat{} and \BSimeaturemat{} is formulated to ensure this goal is discussed in the following sections.
\subsection{Quadratic Problem}
\label{sec:QPMethod}
We consider the problem of selecting the~\gls{nReducedFeatures} out of~\gls{nFeatures} features and assigning them to the classes as a binary quadratic problem, that can be solved globally optimal.
Specifically, the feature selection $\fvecgurobi~\in\{0,1\}^{\gls{nFeatures}}$ and 
assignment between features and classes $\wgurobi~\in\{0,1\}^{\gls{nClasses}\times \gls{nFeatures} }$ are jointly
optimized, with $\wgurobi^*$ being \wgurobi{} for the selected features. 
Given a similarity matrix~$\simmat{}~\in \mathbb{R}^{\gls{nClasses}\times \gls{nFeatures} }$ the main objective is to maximize the similarity $\objective{}_A$ between the selected features and their assigned classes 
\begin{equation}
  \objective_A = \simieq 
  \label{eq:mainObj}
\end{equation}
with $\circ$ indicating the Hadamard product. 
Here,~\fvecgurobi{} indicates whether a feature is selected and~\wgurobi{} describes if a feature is assigned to the class.
\checktext{Note that we use \cindex{} to index classes and \findex{} for features.}
The sparsity and low-dimensionality are formulated as constraints for the optimization:
\begin{align}
  \sum_{\findex=1}^{\gls{nFeatures}}\infvec{\findex} &= \gls{nReducedFeatures}\label{eq:FeatureSel}\\
   \sum_{\findex=1}^{\gls{nFeatures}} w_{\cindex,\findex} \infvec{\findex} &= \gls{nperClass} \quad \forall \cindex \in \{1, \dots,\gls{nClasses}\}
   \label{eq:5perInit}
\end{align}
To allow the \gls{NewlayerName} the differentiation between all classes and enable effective fine-tuning, 
we
additionally add constraints that no two classes are assigned to the same set of features:
\begin{equation}
   (\boldsymbol{w}_\cindex \circ \boldsymbol{w}_{\cindex'})^T \fvecgurobi{} < \gls{nperClass} \quad \forall \cindex,\cindex' \in \{1, \dots,\gls{nClasses}\}
   \label{eq:Unique}
\end{equation}
Note that the constraints in \cref{eq:5perInit,eq:Unique} technically define a quadratically constrained quadratic program (QCQP).
To make the QCQP computationally tractable, the constraints are relaxed and added iteratively for classes that violate the constraints.
The 
efficient 
implementation is discussed in detail in \cref{sec:gurobitricks}. %
The general formulation of the problem allows us to add further nuance to the optimization and include more desiderata. Since a high representational capacity is desired for the selected features, the cross-feature similarity matrix$~\FSimeaturemat~\in\mathbb{R}^{\gls{nFeatures}\times \gls{nFeatures} }$ is incorporated to reduce the similarity between the selected features:
\begin{equation}
    \objective_R = -\correq\label{eq:corrObj}
\end{equation}
Additionally, the selection of specific features can be guided via a selection bias$~\BSimeaturemat~\in\mathbb{R}^{\gls{nFeatures} }$ 
\begin{equation}
    \objective_B = \biaseq\label{eq:locObj},
\end{equation}
where a higher value $\BSimeaturemat_i$ leads to a preferred selection of the feature $i$.
The combination of all these objectives leads to:
\begin{equation}
  \max_{\wgurobi,\fvecgurobi{}} \, \objective = \max_{\wgurobi{},\fvecgurobi{}} \,\objective_A + \objective_R + \objective_B\label{eq:fullObj}
\end{equation}

The formulation in standard form for quadratic problems $\frac{1}{2} \mathbf{x}^T \mathbf{Q} \mathbf{x} + \mathbf{c}^T \mathbf{x}$
with $\mathbf{Q}$ capturing the quadratic terms $\objective{}_A$ and $\objective{}_R$, and $\mathbf{c}$ incorporating the linear term $\objective{}_B$ is
included in ~\cref{suppsec:QP}.\\
\subsection{Class-Feature Similarity} %
The class-feature similarity matrix \simmat{} with entries
\insimmat{\cindex}{\findex} should reflect how beneficial the assignment of feature $\findex$ to class $\cindex$ \checktext{is}
for the classifier. 
As 
every feature gets assigned to multiple classes, which themselves become assigned to multiple features, 
the metric should
focus on a robust positive relation between the activation and likelihood of a sample being of the respective class.
This is captured by the Pearson correlation coefficient 
$\insimmat{\cindex}{\findex}$ 
between the feature distribution $\boldsymbol{f}_{:,\findex}$ and the label 
vector $\boldsymbol{l}^{\cindex}\in\{0,1\}^{\gls{nTrainImages}}$, 
in which for all \gls{nTrainImages} training images a $1$ indicates the label being $\cindex$. %

\subsection{Feature-Feature Similarity} %
Just maximizing \cref{eq:mainObj} can lead to very similar features being selected
which is neither beneficial for interpretability nor for accuracy as representational capacity is lost and multiple features develop towards the same concept during fine-tuning.
To prevent this, selecting similar features in \simmat{} should be penalized in the objective.
We choose
the cosine similarity between the class similarities of two features $\findex\neq \findex'$ in \simmat{}
for \FSimeaturemat{} with $ \symbolFeaFea_{\findex,\findex'} = \mathrm{ReLU} \left( \frac{\boldsymbol{a}_{:,\findex}^T\boldsymbol{a}_{:,\findex'}}{|\boldsymbol{a}_{:,\findex}||\boldsymbol{a}_{:,\findex'}|}\right)$,
using ReLU to focus on preventing redundant features and $\symbolFeaFea_{\findex,\findex'}=0$ for $\findex=\findex'$.
As we are only interested in preventing the selection of highly similar features, we can clip all entries in \FSimeaturemat{} below an $\epsilon$ to $0$ to enable a fast solving of the QP.
The details are discussed in \cref{sec:gurobitricks}. 

\subsection{Feature-Bias} %
\label{sec:fbias}
The Feature-Bias \BSimeaturemat{} describes the benefit of selecting each feature.
This can be used to steer the model towards specific desiderata. 
As diversity is generally preferred~\citep{norrenbrocktake, alvarez2018towards} for interpretable models, 
a bias towards more local features is used,
\begin{equation}
  b_\findex = \frac{1}{\gls{nTrainImages} \sum_j f_{j, \findex} } \sum_{j=1}^{\gls{nTrainImages}} max(\boldsymbol{S}^\findex_j) 
  f_{j,\findex}\quad.
\end{equation}
Here $\boldsymbol{S}^\findex_j$ is the softmax over the spatial dimensions of the $\findex$-th feature map for the image $j$.
Scaling the feature bias by their activation leads to the selection of features that are more localized when their activation is high. 
Alternatively, the bias can be used to steer the selection towards other criteria the practitioner might identify as relevant, which we demonstrate in the appendix.
We center \BSimeaturemat{} and scale the maximum absolute value to be $\lambda$, whose strength defines the priority put on the bias.

\section{Experiments}
\glsreset{resNet}
\glsreset{denseNet}
\glsreset{incv}
\glsreset{cubheader}
\glsreset{travelingheader}
\glsreset{stanfordheader}
\glsreset{fgvcheader}
\glsreset{imgnetheader}
\glsreset{PIP-Net}
Following prototype-based methods we applied our method to \cubheader{} and \stanfordheader{}. 
To showcase \gls{NewlayerName}'s broad applicability, we also include results on the large-scale dataset \imgnetheader, to which most interpretable methods are not applicable.
Notably, \cubheader{} contains annotations of human concepts which we use to measure \cubsim.
An overview of the used datasets is shown in \suppl{}  \cref{table:DatasetOverview}.
As our method is independent of the used backbone, 
we evaluated it across various architectures, but focus on \resnet{} in this paper.
Similar results on Resnet34,  \gls{incv} and Swin Transformer~\citep{liu2021swin}, as well as detailed results with standard deviations, are included in \suppl{} \cref{suppsec:Results}.
We do not apply our method to other interpretable models like \gls{PIP-Net}, as QPM is an alternative way of inducing compactness and the features of \gls{PIP-Net} are not general, thus ill-suited for a broad assignment.

\subsection{Implementation Details}
We generally followed \gls{PIP-Net} for the data preparation.
Specifically, the images are first cropped to the ground truth bounding box for \cubheader{} and \travelingheader{}.
For all datasets, the images are resized to $224\times224$. Following \gls{PIP-Net}, \textit{TrivialAugment}~\citep{Muller_2021_ICCV} is used and %
the strides of ResNets
are also set
to 1 to obtain more fine-grained feature maps.
The remaining parameters, including dense training for $150$ epochs on fine-grained datasets and directly using the pretrained model on \imgnetheader{} with subsequent $40$ epochs of fine-tuning, mirror the \gls{layerName}  and are described in \cref{suppsec:impl}.
Note that \gls{NewlayerName} is trained more efficiently than \gls{qsenn}, as it does not use multiple training iterations during fine-tuning.
We set $\gls{nperClass}=5$ and $\gls{nReducedFeatures}=50$ for \gls{NewlayerName}, unless stated otherwise.
We demonstrate the impact of changing the parameters in the ablation studies but choose these, as it is in line with prior literature~\citep{norrenbrock2024q, norrenbrocktake}, $\gls{nReducedFeatures}<\gls{nClasses}$, and it enables sufficiently compact explanations~\citep{miller1956magical}.
The shown results, \eg{} \cref{tab:aCCproto-table,tab:Interpproto-table}, are the mean across $5$ seeds, with the exception of $3$ for \imgnetheader{}, \gls{PIP-Net} and \gls{ProtoPool}.
For comparison, 
all models are exclusively pretrained on \imgnetheader{}.
This change did affect \gls{ProtoPool}, but even with iNaturalist~\citep{van2018inaturalist} pretraining, we could not reproduce the reported results by~\citet{rymarczyk2022interpretable}. 

\subsubsection{Quadratic Problem}
\label{sec:gurobitricks}
This section presents details on how the described quadratic problem \checktext{with \cref{eq:fullObj} as objective} is solved using \gurobi.
We incorporated deduplication 
and the assignment of an equal number of features to all classes of \cref{eq:5perInit,eq:Unique} using
an iterative approach with relaxed
constraints.
Specifically, 
the model 
is optimized
without these constraints, but instead 
$\boldsymbol{1}^T\wgurobi\fvecgurobi=\gls{nperClass}\gls{nClasses}$.
Then, after each iteration,
all violated constraints 
are added
to the model, but only limited to a running set of 
features $\Gamma\in\{0,1\}^{\gls{nFeatures}}$, which gets extended during the iteration.
Next to the features, we also maintain a set of classes $C_\mathrm{duplicates}$ that were equal at one iteration and classes $C_\mathrm{sparse}$ that ever had too few features assigned.
Instead of \cref{eq:5perInit,eq:Unique} the relaxed constraints
\begin{align}
  \boldsymbol{w}_{\cindex,\Gamma}^T
  \fvecgurobi_{\Gamma} & \geq \gls{nperClass} \quad \forall \, \cindex \in C_\mathrm{sparse} \label{eq:NotGLob}\\
     (\boldsymbol{w}_\cindex \circ \boldsymbol{w}_{\cindex'})_{\Gamma}^T\fvecgurobi_{\Gamma} &< \gls{nperClass} \quad \forall \, \cindex,\cindex' \in C_\mathrm{duplicates} \label{eq:4DedupQP}
\end{align}
are added, where $\wgurobi_{\cindex,\Gamma}$ describes indexing $\wgurobi_\cindex$ where $\Gamma=1$. 
Additionally, we set the start solution for the next optimization to a good, usually optimal, feasible solution for the currently selected set of features. 
As we need multiple iterations to enforce all constraints, we limit the time spent on one iteration to $3$ hours and set the gap to optimality to $10^{-4}$. 
In our experiments, the global optimum for the relaxed problem is usually found in less than $4$ hours 
for fine-grained datasets, and roughly $11$ hours for \imgnetheader{}
using a CPU like \textit{EPYC 72F3}.
While \cref{eq:NotGLob} changes the desired optimization problem, the resulting objective is very close (achievable gap of less than $1\%$) to the global optimum, which is infeasible to compute and does not lead to an improved model. 
The experiments to verify this claim are included in \suppl{} \cref{suppsec:optimal}.
Finally, alongside our experiments, previous work~\citep{hornakova2021making} shows that the exact global optimum is not always preferred for relevant metrics.
To make the relative weighting of the multiple objectives $\objective_A$, $\objective_R$ and $\objective_B$ easier, %
\simmat{} 
is scaled with \gls{nClasses} and \gls{nperClass}
to have a maximum of $1$ for $\gls{nClasses}=200$ and $\gls{nperClass}=5$. %
Since \gls{nReducedFeatures} features need to be chosen, 
all entries below $\epsilon$ in \FSimeaturemat{} are set to $0$, where $\epsilon$ is the highest value,
for which there still exists a selection with $\objective{}_R=0$. 
This is equivalent to finding the maximal $\epsilon$ for which the graph described by $\boldsymbol{G}$ with 
\begin{equation}
  g_{\findex,\findex'} =\begin{cases}
0 \quad\mathrm{if} \quad r_{\findex,\findex'} \geq \epsilon \\
1 \quad\mathrm{else},
\end{cases}
\end{equation}
has a maximum clique of size \gls{nReducedFeatures}.
We used approximations~\citep{pattabiraman2015fast,boppana1992approximating} and a sufficiently sized approximated maximum clique as the start value for \fvecgurobi.
Additionally, the remaining nonzero values in \FSimeaturemat{} are scaled to have a maximum of $1$. 
For scaling the bias \BSimeaturemat{}, we clipped outliers, centered the remaining values around $0$ and scaled the maximum absolute value to be $\lambda = \frac{1}{\sqrt{10}}$, which is empirically found. 

\subsection{Metrics}
\begin{figure*}[thp]
  \centering
\includegraphics[width=\linewidth]{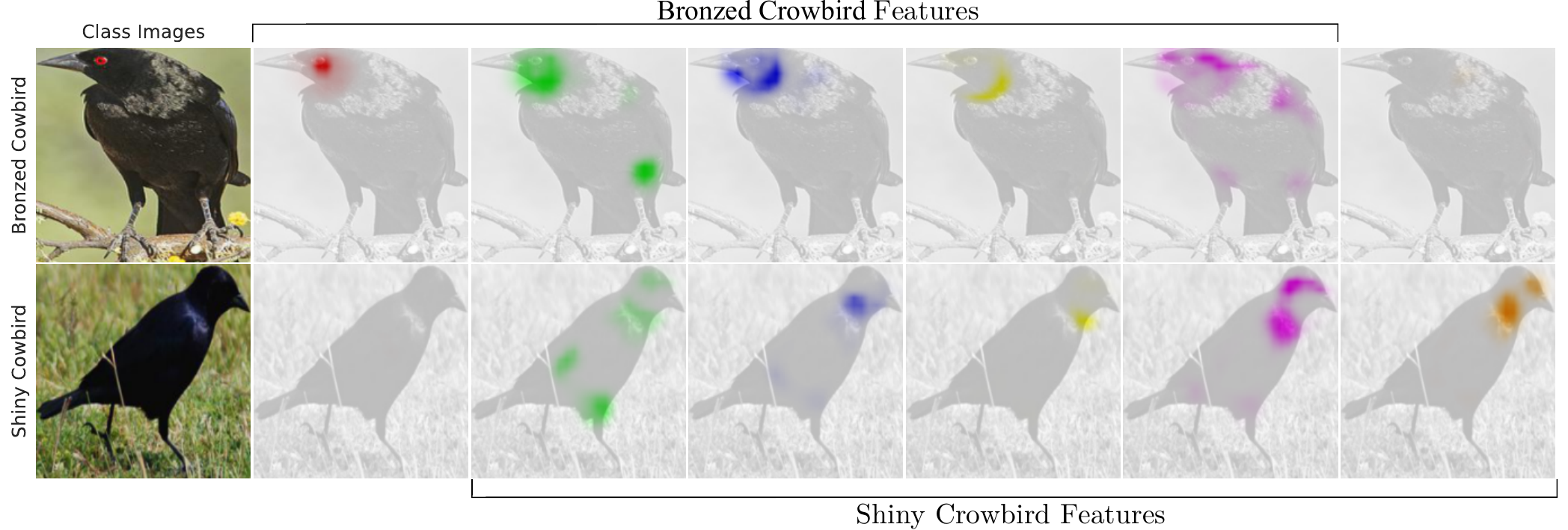}
   \vspace{-.5cm}\caption{Contrastive faithful class explanations for \gls{NewlayerName} trained on \cubheader{}: 
   Without any additional supervision, 
   \gls{NewlayerName} learns to differentiate 
   Shiny and Bronzed Cowbird ($\mathrm{\ClassSim}^{gt}=0.97$) using the red eye just like humans do, as the annotations in \cubheader{} or the screenshot in \cref{fig:ScreenshotDiffShiny} show.}
\label{fig:CubSim}
\vspace{-.5cm}
\end{figure*}
\newlength{\globaltextwidth}
\setlength{\globaltextwidth}{\textwidth}%
\begin{figure}
\centering
   \begin{subfigure}[t]{.495\linewidth}
     \centering
\includegraphics[width=.3\globaltextwidth]{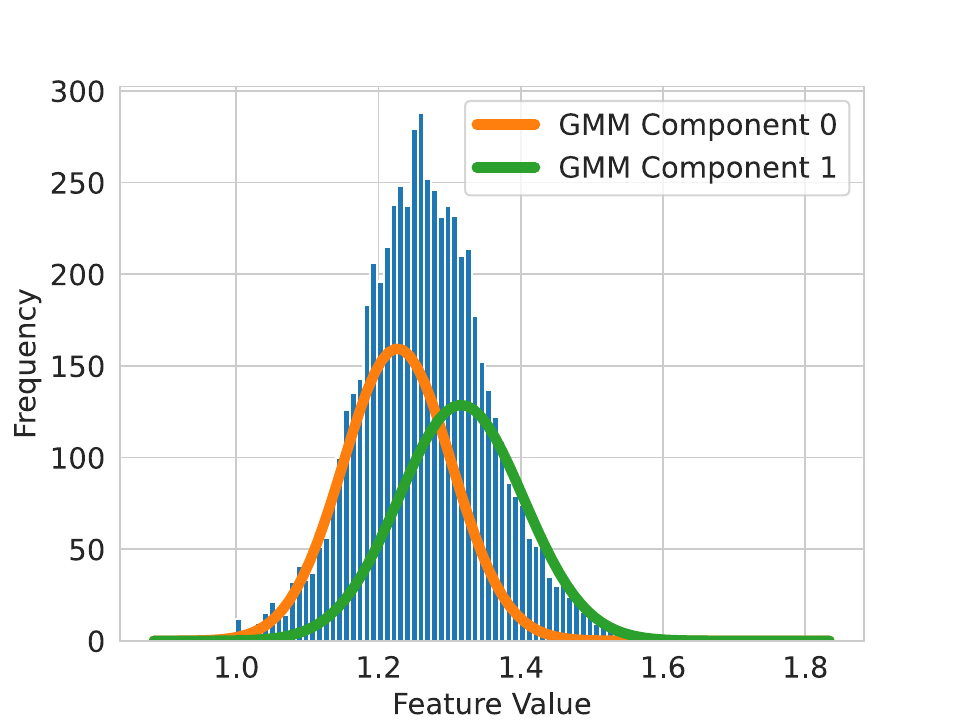}
         \captionsetup{justification=centering}
     \caption{
     Baseline 
     \resnet{} 
     with $\mathrm{\contrastiveness{}} = 41.8\%$}
   \end{subfigure}
   \hspace{5mm}
   \begin{subfigure}[t]{.45\linewidth}
     \centering
   \includegraphics[width=.3\globaltextwidth]{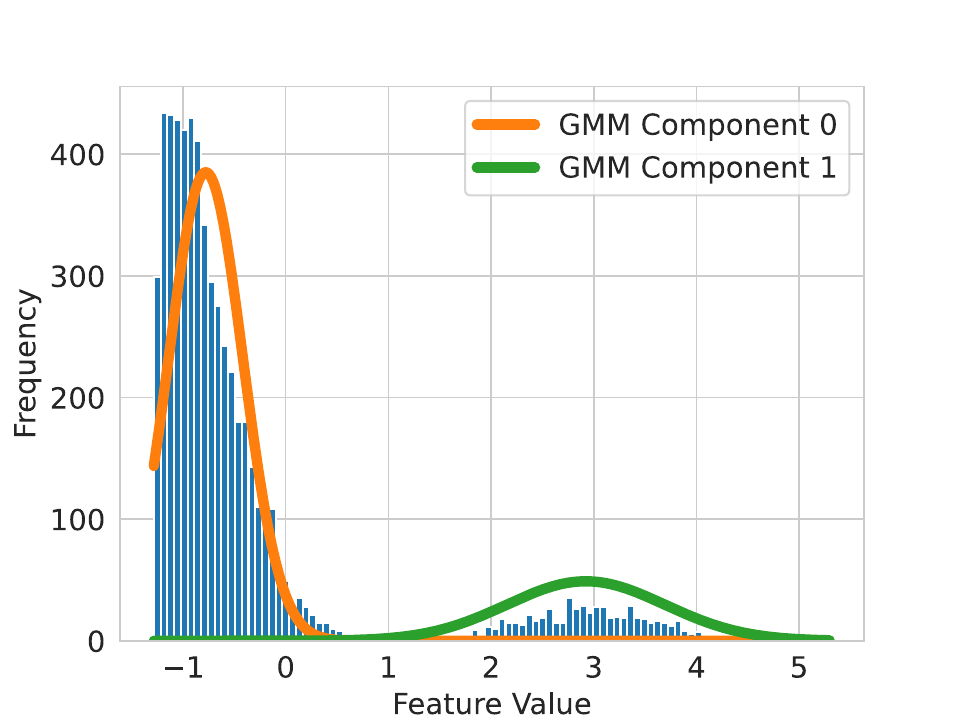}%
    \captionsetup{justification=centering}
     \caption{\gls{NewlayerName} with $\mathrm{\contrastiveness{}}= 99.9\%$}
   \end{subfigure}
   \hfill
    \caption{Extreme Examples for feature distributions and their \contrastiveness{} on \cubheader{}. }
    \label{fig:Contr}
    \vspace{-0.35cm}
\end{figure}

Following \gls{PIP-Net} as recent work, we evaluate the accuracy and compactness, measured as number of total features~\gls{nReducedFeatures} and number of features per class~\gls{nperClass}.
Additionally, our \gls{NewlayerName} learns interpretable class representations, summarized in~\cref{tab:ClassRepresent}, that are composed of features.
As discussed in \cref{sec:RelatedWork}, diversity, contrastiveness, generality and grounding are desired aspects of explanations.
While we believe that our sparse binary assignment is very well suited for a detailed analysis and alignment of the learned features, as it likely prohibits superposition, polysemantic neurons are still likely to occur and hard to measure for \gls{NewlayerName} and all end-to-end trained interpretable models.
Therefore, we omit measuring the grounding of features and instead focus on contrastive,  general and diverse as desirable \textit{and} quantifiable qualities of features as building blocks of our interpretable class representations, whose \cubsim{} we estimate using the attributes contained in \cubheader{}.
Specifically, every class $\cindex$ in \cubheader{} is annotated with a vector $\boldsymbol{a}_\cindex\in[0,1]^{312}$, where $a_{\cindex,j}$ indicates the fraction of images with label $\cindex$, in which a human perceives the attribute $j$ to be present. 
With these vectors,
we compute 
the ground truth structural class similarity
$\boldsymbol{\mathrm{\ClassSim}}^{gt}\in[0,1]^{\gls{nClasses}\times\gls{nClasses}}$ with ${\mathrm{\classSim}}^{gt}_{\cindex,\cindex'}$ being the cosine similarity between $\boldsymbol{a}_\cindex$ and $\boldsymbol{a}_{\cindex'}$.
Similarly,
$\boldsymbol{\mathrm{\ClassSim}}^{Model}\in[-1,1]^{\gls{nClasses}\times\gls{nClasses}}$
is
based on the class vectors in the interpretable classification layer.
We then report the similarity for the top $25$ most similar unique pairs of classes $C_\mathrm{Sim}$ in reality 
\begin{equation}
  \mathrm{\cubsim} = 
  \frac{\sum_{\cindex,\cindex' \in C_\mathrm{Sim} } \mathrm{\classSim}^{Model}_{\cindex,\cindex'} }{\sum_{\cindex,\cindex' \in C_\mathrm{Sim} } \mathrm{\classSim}^{gt}_{\cindex,\cindex'}}.
\end{equation}
Models with high \cubsim{} offer an interpretable human-like class-similarity, \eg{} using the apparently different head to differentiate between Rottweiler and Doberman in \cref{fig:metrics_full} or differentiating shiny and bronzed cowbird by its only separating attribute, shown in \cref{fig:CubSim}.\\
To measure the contrastiveness of features, 
a Gaussian mixture model with two components 
is fit 
to every feature distribution $\boldsymbol{f}_{:,\findex}$, resulting in the 
normal distributions $\mathcal{N}_1^\findex$ and $\mathcal{N}_2^\findex$, visualized in \cref{fig:Contr}.
We then compute the \textit{\contrastiveness{}} as average of all features using the overlap~\citep{doi:10.1080/03610928908830127} between the two 
distributions: \begin{equation}
  \mathrm{\contrastiveness{}} =\sum_{\findex=1}^{\gls{nReducedFeatures}} 1 - \mathrm{Overlap}(\mathcal{N}_1^\findex, \mathcal{N}_2^\findex),\end{equation}
as bi-modal contrastive features can be represented by two non-overlapping distributions.
The binary quality of the features is also indicated in \cref{fig:metrics_full,fig:CubSim}, as the features are normed per column.\\

Additionally, the features should capture a general concept, instead of a class-specific one. This can be measured via the \textit{\generality} $\tau$:
\begin{align}
\tau = 1-\frac{1}{\gls{nReducedFeatures}} \sum_{\findex=1}^{\gls{nReducedFeatures}} \max_\cindex
&\frac{\sum_{j=1}^{\gls{nTrainImages}} l^\cindex_j (f_{j,\findex} - \min \boldsymbol{f}_{:,\findex})}{\sum_{j=1}^{\gls{nTrainImages}} (f_{j,\findex}- \min \boldsymbol{f}_{:,\findex})}
\end{align}
It
measures which fraction of the zero-based feature activation across the entire dataset is
not
focussed on the most related class.
A model with high \generality{} has features that recognize a shared concept for multiple classes, like the $4$ central features in \cref{fig:metrics_full,fig:CubSim}. 
Notably, as opposed to Dependence~\citep{norrenbrock2024q}, \generality{} can capture the assignment of multiple class detectors to the same class.
\\
For measuring the spatial diversity of the features, \oldloc{5}~\citep{norrenbrocktake} has been proposed.
The \oldloc{5} however suffers from the non-linear behavior of the softmax, resulting in
scale-dependency (\cref{stab:oldloc5}).
Therefore, we propose the \textit{Scale-Invariant-Diversity@5} (\loc{5})
\begin{align}
\hat{M}^{\findex}_{i,j} = \frac{M^{\findex}_{i,j}}{\frac{1}{\gls{featuresMapwidth}\gls{featuresMapheigth}} \sum |\textbf{M}^{\findex}|} \quad \hat{S}^{\findex}_{i,j} = \frac{e^{\hat{M}^{\findex}_{i,j}}}{\sum_{m,n} e^{\hat{M}^{\findex}_{m,n}}} \\
   \loc{5} = \frac{\sum_{i=1}^{\gls{featuresMapheigth}}\sum_{j=1}^{\gls{featuresMapwidth}}\max(\hat{S}^{1}_{i,j},\hat{S}^{2}_{i,j}, \dots, \hat{S}^{5}_{i,j})}{5} , 
\end{align}
where $\hat{\textbf{S}}^\findex$ refers to the result of softmax applied to the $\findex$-th highest weighted feature map $\textbf{M}^{\findex}$, scaled by its absolute mean.
A high \loc{5} is visible in \cref{fig:metrics_full,fig:CubSim}, as the $5$ features used for each class, localize on very different regions in the image.

\subsection{Results}
\begin{table*}[t!]

  \caption{Comparison on compactness and accuracy with \resnet{}: \gls{NewlayerName} shows increased accuracy and compactness.
 The compactness-accuracy trade-off is 
 shown 
 in \cref{fig:interptradeoff}.
 \boldnessstatement
 }
 \label{tab:aCCproto-table}
 \vspace{-.3cm}
 \resizebox{\linewidth}{!}{
 \centering
 \begin{tabular}{l|ccc|ccc|ccc}
  \hline
  Method & \multicolumn{3}{c|}{Accuracy \arrowUp} &
  \multicolumn{3}{c|}{Total Features \arrowDown} & \multicolumn{3}{c}{Features / Class \arrowDown} \\
  & CUB& CARS & INET & CUB& CARS & INET & CUB& CARS & INET 
  \\
  \hline
  Baseline \resnet{}& 86.6 & 92.1 &76.1 & 2048 & 2048 &2048& 2048& 2048& 2048\\
  \hline
  \glmtable{}& 78.0 & 86.8  & 58.0 & 809 & 807& 1627 & \textbf{5} &\textbf{5}&\textbf{5} \\
  \pipnettable{} & 82.0 & 86.5  &-& 731 &669&-& 12 &11 &  - \\ %
  \protopooltable{} & 79.4 &87.5& -& 202 & 195 & - &202 & 195 &  - \\ %
  \slddtable{} & 84.5 & 91.1  & 72.7 & \textbf{50} & \textbf{50} & \textbf{50}& \textbf{5}& \textbf{5} &\textbf{5} \\
  \qsenntable{} & 84.7 & 91.5  & \textbf{74.3} & \textbf{50} & \textbf{50} & \textbf{50}& \textbf{5}& \textbf{5} &\textbf{5} \\
  \hline
  \gls{NewlayerName} (Ours) & \textbf{85.1} & \textbf{91.8}& \underline{74.2} & \textbf{50}  &\textbf{50}& \textbf{50}& \textbf{5}&\textbf{5}&\textbf{5} \\ %
  \hline
 \end{tabular}
}
 \vspace{-.3cm}
\end{table*}
\begin{table*}[t!]
 
 \caption{Comparison on Interpretability metrics with \resnet{}. Due to required annotations, \cubsim{} (abbreviated SG) can only be computed for \cubheader{}. }
\label{tab:Interpproto-table}
\vspace{-.3cm}
\resizebox{\linewidth}{!}{
 \centering 
 \begin{tabular}{l|ccc|ccc|ccc|c}
  \hline
  Method & \multicolumn{3}{c|}{\loc{5} \arrowUp} &
  \multicolumn{3}{c|}{\generality{}\arrowUp} & \multicolumn{3}{c|}{\contrastiveness{} \arrowUp} & SG \arrowUp\\
  & CUB& CARS  & INET & CUB& CARS  & INET & CUB& CARS & INET & CUB
  \\
  \hline
  Baseline \resnet{}& 57.7 & 54.4 &37.1 & 98.0 & 97.8& 99.4 & 74.4 & 75.1&71.6  & 34.0\\
  \hline
  \glmtable{} & 55.4 &51.8 &35.8 &\textbf{97.8} &\textbf{97.6}  &\textbf{99.4} & 74.0 & 74.5 &71.7 & 2.5  \\
  \pipnettable{} & \textbf{99.1} & \textbf{99.0} &- & 75.6 &62.9&- & \textbf{99.5} &\textbf{99.5} &- & 6.7  \\ %
  \protopooltable{} & 24.5 &30.7& -& {96.9}& {96.0} &- &76.7& 78.9 & -  & 13.9 \\
  \slddtable{}
   &88.2 & 88.6 & \underline{64.7} & {96.2} &{95.6}&{98.6} & 87.2 &89.7 &\textbf{93.4} & 29.2 \\
    \qsenntable{}
   &\underline{93.3} &\underline{94.4} &\textbf{82.0} & {95.5} &{94.8}&{98.7 } &  93.0 &94.2 &\underline{92.6} & 23.4 \\
   \hline
   \gls{NewlayerName} (Ours) & {90.1} & {89.6} &{64.1} & \underline{97.0}& \underline{96.5}&\underline{99.1}
   & \underline{96.0} & \underline{97.7} & {89.3} & \textbf{47.9}\\ 
  \hline
 \end{tabular}
 }
\vspace{-.4cm}
\end{table*}

This section discusses the experimental results. 
The usual metrics for compactness-based globally interpretable models are shown in \cref{tab:aCCproto-table}. 
For the fine-grained datasets, \gls{NewlayerName} is among the most compact models while showing the highest accuracy, thus setting the state of the art for interpretable models.
On \imgnetheader{}, where prototype-based methods are not even applicable, \gls{NewlayerName} is only marginally beaten by
\gls{qsenn}, which uses compute-intensive iterations and negative reasoning for some classes, which significantly hinders interpretability. 
A runtime analysis is shown in \cref{ssec:Runtime}.\\
The results for the interpretability metrics are shown in \cref{tab:Interpproto-table}.
Note that \glmtable{} and PIP-Net are hardly comparable, as \glmtable{} uses the uninterpretable features of a black-box model and PIP-Net learns very localized class-detectors, with some features activating to 99\% on just a single class.
In contrast, \gls{NewlayerName} achieves excellent values across all metrics and datasets in this multicriterial  task of self-explaining neural networks, summarized in \cref{fig:RadarMax}.
Its
interpretable class representations, composed of diverse, general and contrastive features, mirror reality, as measured by \cubsim{}. 
Note that \gls{NewlayerName} learns grounded representations as shown in \cref{fig:metrics_full,fig:CubSim} without any additional supervision and is able to communicate the only differentiating factor it uses. 
\gls{NewlayerName}'s local behavior then follows its faithful global explanations, which leads to trustworthy classifications and predictable errors when the differentiating factor is not present, as in \cref{fig:LocalFailureP}.
The appendix contains 
more visualizations, including a discussion of failure cases in \cref{ssec:Failure}, 
a discussion on polysemantic features (\cref{ssec:poly}), an extension of \cubsim{} to \imgnetheader{} (\cref{ssec:imgsim}) and a discussion of limitations and future work (\cref{ssec:limits}).\\

\subsection{Ablation Studies}
\begin{figure}[t]
    \centering
    \begin{minipage}[t]{.62\linewidth}
        \includegraphics[width=\linewidth]{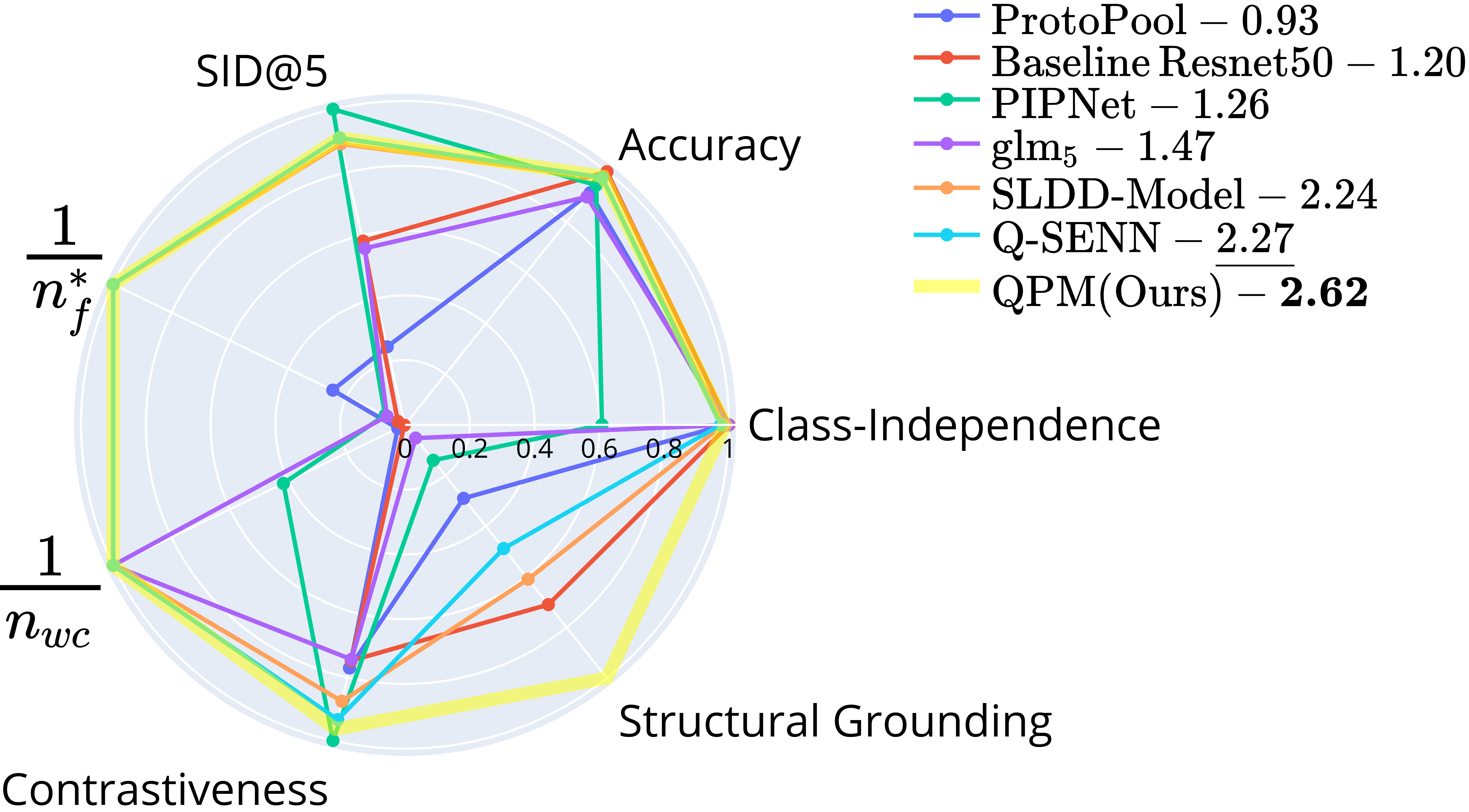}
        \caption{Radar plot across all considered metrics for \cubheader{}. Metrics which are preferred to be lower (\gls{nReducedFeatures}, \gls{nperClass}) are encoded as $\frac{1}{x}$ and every value is given as a fraction of the maximum. %
        Values in legend are
        the area of each radar plot.}
        \label{fig:RadarMax}
    \end{minipage}
    \hfill
    \begin{minipage}[t]{.35\linewidth}
        \centering
                \vspace{-50mm} %
\includegraphics[width=\linewidth]{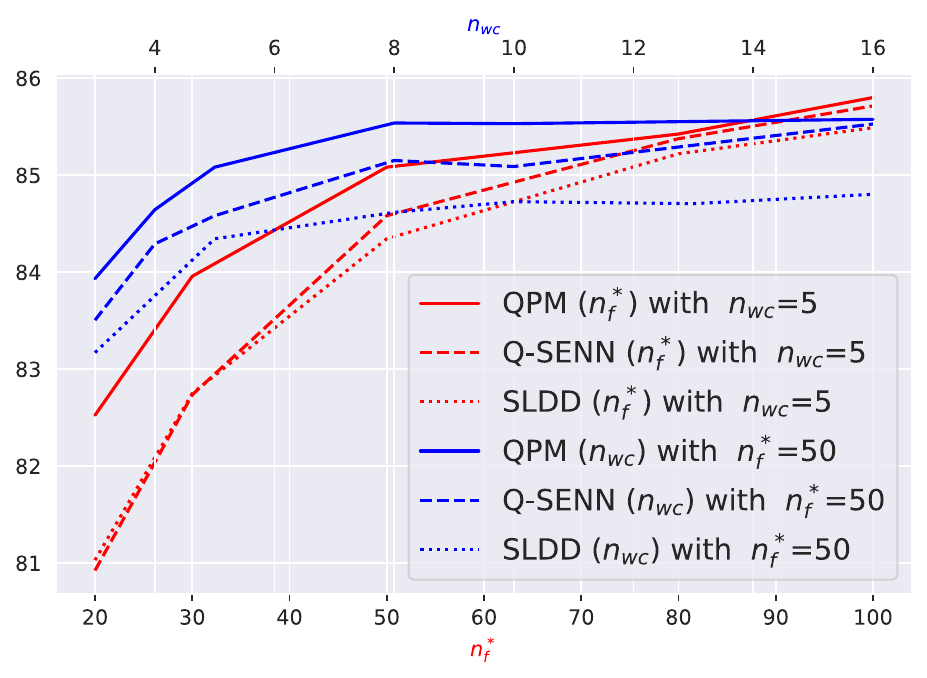}
        \caption{Compactness-accuracy trade-off compared with \gls{qsenn} (dashed) and \gls{layerName} (dotted) with Resnet50 on CUB-2011. 
        With increasing compactness, \gls{NewlayerName}'s optimal usage of the set \gls{nReducedFeatures} and \gls{nperClass} becomes more beneficial.
        }
        \label{fig:interptradeoff}
        
    \end{minipage}
    \vspace{-.3cm}
    \begin{minipage}{0.62\textwidth}
        \centering
        \begin{table}[H]
            \vspace{-.5cm}\caption{
            \checktext{Impact of including (\cmark) additional objectives $\objective_B$~(\cref{eq:locObj}) for locality and $\objective_R$~(\cref{eq:corrObj}}) to reduce correlation alongside $\objective_A$ in \cref{eq:fullObj} on \cubheader{} with \resnet{}.
            Correlation is measured as the average maximally similar feature according to cosine similarity, formulated in \cref{ssec:corremet}.
            }
            \label{tab:ablations}
            \centering
            \begin{tabular}{lcccc}
            \toprule
            $\objective_B$ & $\objective_R$
            & Accuracy \arrowUp & \loc{5} \arrowUp & Correlation \arrowDown\\
            
            \midrule
            \xmark & \xmark & 84.6 & 89.0 & 33.9 \\ 
            \cmark & \xmark & 84.4 & \textbf{90.3} & 33.5 \\ 
            \xmark & \cmark & \underline{85.0} & 88.5 & \textbf{22.7} \\
            \midrule
            \cmark & \cmark & \textbf{85.1} & \underline{89.6} & \underline{24.6}\\
            \bottomrule
            \end{tabular}
        \end{table}
    \end{minipage}%
    \quad
    \begin{minipage}{0.35\textwidth}
        \centering
        \vspace{-.4cm}\includegraphics[width=.6\textwidth]{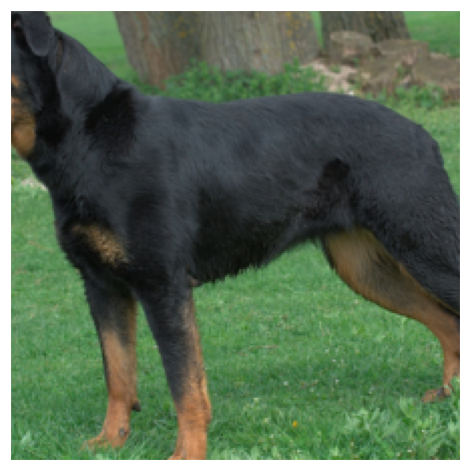}
        \caption{Misclassified example of a QPM, explained in \cref{fig:metrics_full}: Predictably given the explanation, the model classifies a Doberman as a Rottweiler due to the absent head.}
\label{fig:LocalFailureP}
    \end{minipage}
    \vspace{-.3cm}
\end{figure}

This section validates the impact of the individual objectives in the quadratic problem in 
\cref{tab:ablations} and presents the
 compactness trade-off in \cref{fig:interptradeoff}.
 We focus on \cubheader{} but observed similar results for other datasets.
  The compactness-accuracy tradeoff for \gls{NewlayerName} compared with \gls{qsenn} and the \gls{layerName} is visualized in \cref{fig:interptradeoff}. 
 The global optimization clearly leads to a more effective use of the defined capacity, with the highest uplift in the very high compactness regime, \eg{} 1.5 percent points at $\gls{nReducedFeatures}=20$, where a good selection and assignment naturally has more impact.\\
 \checktext{The impact of the feature-feature similarity matrix \FSimeaturemat{} and feature selection bias \BSimeaturemat{} is shown in \cref{tab:ablations}.} 
 Incorporating a bias \BSimeaturemat{} for local feature maps further increases the \loc{5}. 
 On the other hand,
 reducing feature similarity through \FSimeaturemat{} effectively reduces the %
 correlation
 between the resulting features, which improves accuracy, as the model uses its capacity more effectively. 
  \checktext{In summary, the inclusion of the secondary objectives $\objective_R$ and $\objective_B$ is beneficial for the resulting model, improving the desired aspects not just after solving the QP but also in the resulting model after fine-tuning.}\\
The appendix contains further ablation studies to support our claims, demonstrating the ability to steer (\cref{ssec:steer}), validating the choice of correlation as metric for \simmat{} (\cref{ssec:auroc}) and showing the benefits of enforcing exactly \gls{nperClass} features per class (\cref{ssec:balancedAbl}).

\section{Conclusion}
In this paper, we introduced the Quadratic Programming Enhanced Model (\gls{NewlayerName}). 
It uses discrete optimization to find an optimal feature selection and assignment of just $5$ to each class.
With this easy-to-understand assignment, the resulting \gls{NewlayerName} is more interpretable than previous methods, as it has contrastive faithfully interpretable class-representations, shows \cubsim{}, is steerable, and its features have excellent \loc{5}, \generality{}and \contrastiveness{}.
Additionally, it further closes the accuracy gap to the drastically less robust uninterpretable baseline.
\Cref{fig:RadarMax} shows that only \gls{NewlayerName} excels in all metrics, thus setting a new state of the art for compactness-based interpretable models, 
while delivering unprecedented global interpretability even to \imgnetheader{}.

\subsubsection*{Acknowledgments}
This work was supported by the Federal Ministry of Education and Research (BMBF), Germany, under the AI service center KISSKI (grant no. 01IS22093C), the Deutsche Forschungsgemeinschaft (DFG) under Germany’s Excellence Strategy within the Cluster of Excellence PhoenixD (EXC2122), 
the European Union  under grant agreement no. 101136006 – XTREME.
The work has been done in collaboration and partially funded by the Intel Corporation.
This work was partially supported by the German Federal Ministry of the Environment, Nature Conservation, Nuclear Safety and Consumer Protection (GreenAutoML4FAS project no. 67KI32007A).

\bibliography{iclr2025_conference}
\bibliographystyle{iclr2025_conference}

\appendix

\clearpage
\setcounter{page}{1}
\title{Supplementary Material for:\\ QPM: Discrete Optimization for Interpretable Image Classification}
\maketitle
\renewcommand\thesection{\Alph{section}}
\renewcommand\thesubsection{\thesection.\Alph{subsection}}
 \setcounter{section}{0}
\section{Appendix}
This appendix contains additional details of the implementation details, more results with standard deviations, further experiments on steerability, a discussion of failure cases and the formulation of the Feature Diversity Loss~\gls{customLoss}.
Further, the quadratic problem is presented in its standard form and the optimality of the found solutions is discussed.
\section{TravelingBirds}
\begin{table}
\caption{
Statistical overview of 
datasets. \travelingheader{} is used exclusively in the appendix.
}%
\label{table:DatasetOverview}
\vspace{-.1cm}
\begin{center}

\centering
\setlength\tabcolsep{6pt}
\begin{tabular}{cccccc}
\hline
Dataset & \cubheader& \stanfordheader{}& TravelingBirds&   \imgnetheader\\ 
\hline
\# Classes $\gls{nClasses}$ & 200 &196 & 200 &  1000\\
\# Training & \numprint{5994} &\numprint{8144} &  \numprint{5994} & \numprint{1281167}\\
\# Testing & \numprint{5774} &\numprint{8041} &  \numprint{5774} & \numprint{50000}\\
\hline
\end{tabular}
\vspace{-0.35cm}
\end{center}
\end{table}

We use \travelingheader{} as an additional dataset to validate our method and for our steering experiments in \cref{ssec:steer}.
It is based on \cubheader{} and designed to allow the measurement of the robustness of spurious correlations.
Specifically, the background of every class in the training set is replaced with an image of a constant class of Places365~\citep{zhou2017places}.
For the test set, backgrounds of random classes are used, thus measuring if the model learned to rely on the spuriously correlated background.
\section{Implementation Details}
\label{suppsec:impl}
We first describe the implementation on the fine-grained datasets \cubheader{}, \travelingheader{} and \stanfordheader{}. 
All deviating details for \imgnetheader{} are included in \cref{ssec:imgnet}.
The implementation details are similar to the \glsname{layerName}~\citep{norrenbrocktake}, but use the default data for prototype-based methods.
Specifically, we use the exact same dense model for both models in our experiments and only alter the following parts of the pipeline with the same hyperparameters for fine-tuning.
Therefore, the improved metrics can be attributed to the superior selection and assignment.
\paragraph{Architectures}
We implement our method using PyTorch~\citep{Pytorch} and its \gls{imgnetheader} pretrained models as %
feature extractors. 
For \resnet{} and Resnet34 we follow \gls{PIP-Net} and use a smaller stride size of 1 for the two last blocks.

\paragraph{Data}
For training with \cubheader{} and \travelingheader{}, the images are first cropped to the ground truth segmentation, following prototype-based methods~\cite{nauta2023pipnet, rymarczyk2022interpretable}.
After cropping, they are resized to $224\times224$ ($299\times299$ for \incv). 
For \stanfordheader{} and our steerability experiments in \cref{stab:noCrop}, a random crop after resizing one side to the target image size is used instead.
Then normalization, random horizontal flip, jitter and \textit{TrivialAugment}~\cite{Muller_2021_ICCV} is applied.
At test time, no augmentation is used and only cropping, random crop replaced by center crop, resizing and normalization is maintained. 
\paragraph{Dense Training}
We fine-tune the pretrained models on the fine-grained datasets using stochastic gradient descent with a batch size of $16$ for $150$ epochs.
The learning rate starts at $5\cdot10^{-3}$ for the pretrained layers and $0.01$ for the final linear layer and gets multiplied by $0.4$ every $30$ epochs.
We set momentum to $0.9$, $\ell_2$-regularization to $5\cdot10^{-4}$ and apply dropout with rate $0.2$ to the features.
The weighting \gls{cLW}, included in \cref{seq:customLossW}, of the Feature Diversity Loss~\cite{norrenbrocktake} is set to $0.196$ for the %
Resnets, $0.049$ for \incv{} and $0.0245$, the highest value we tried for which all dense models converged, for Swin Transformers.
Note that the values are scaled with the number of patches in the feature maps, leading to numerical values that do not align conveniently with powers of $10$.
\paragraph{Fine-tuning}
After solving the quadratic problem, the model is trained with the final layer fixed to the sparse assignment of selected features $\boldsymbol{W}^*$ for $40$ epochs. 
The learning rate starts at $100$ times the final learning rate of the dense training and decreases by $60\,\%$ every $10$ epochs. 
During fine-tuning, momentum is increased to $0.95$ and dropout on the features reduced to $10\%$.
For Swin Transformers, the batch size is set to 8 and Layer 
normalization~\cite{ba2016layer} 
is turned off after the dense training has finished, ensuring more unrelated features.
All other parameters equal the dense setting.

As the feature maps are the result of 
ReLU~\cite{nair2010rectified}, 
one might expect its values to be strictly $\geq0$.
However, just like for the \gls{layerName}, the features of \gls{NewlayerName} are normalized with a fixed mean and standard deviation before fine-tuning begins, resulting in the sub-zero $\min (\boldsymbol{f}_{:,i})$.
\paragraph{Reproducibility}
For reproducibility, all our experiments with $5$ seeds use the integers $16$ to $20$, ending at $18$ for the $3$ \imgnetheader{} runs, as seed for all random processes.
\paragraph{Scaling the Objective}
To keep a similar relative weighting across changing \gls{nperClass} and \gls{nClasses}, we also scale the main objective for the quadratic problem $\objective_{A}$ with them
\begin{equation}
    \objective_{A}^* =  \frac{1000 \cdot \objective_{A}}{\gls{nperClass} \cdot \gls{nClasses}},
\end{equation}
ensuring no additional scaling for $\gls{nClasses}=200$ and $\gls{nperClass}=5$.
\paragraph{Choice of Pretrained Weights}
We use the pretrained  \resnet{} weights \textit{V1} of PyTorch for our experiments, as the default \textit{V2} has very class-specific features already, with a \generality{} of $92.6\%$. 
For \textit{V2}, a sparse model computed by \glm{} with just $1.1$ features per class can already achieve $66\%$ accuracy on \imgnetheader{}, demonstrating the class-specificness of its features.
For Resnet34 and \gls{incv}, we use the only available set of weights from PyTorch.
For Swin Transformers, we used the original provided weights of PyTorch, as they are suitable for the used image resolution.
\subsection{ImageNet-1K}
\label{ssec:imgnet}
Due to computational constraints, we follow the \gls{layerName}, skip the dense training on \imgnetheader{} and directly use the pretrained model as dense model.
To facilitate the comparability of metrics between the dense model and our experiments, we use the default strides.
For augmentation, we use Lighting noise and omit \textit{TrivialAugment}.
Finally, the learning rate of the fine-tuning starts at  $\frac{1}{100}$ of the value used for the fine-grained datasets to account for the increased size of the dataset.

\subsection{Correlation Metric}
\label{ssec:corremet}
For measuring the effect of reducing correlation between selected features in \cref{tab:ablations}, the \textit{\correlation{}} is used:
\begin{equation}
  \mathrm{\correlation{}} = \frac{1}{\gls{nReducedFeatures}} \sum_{\findex=1}^{\gls{nReducedFeatures}}\max_{\findex\neq \findex'}
  \frac{\boldsymbol{f}_{:,\findex}^T\boldsymbol{f}_{:,\findex'}}{|\boldsymbol{f}_{:,\findex}||\boldsymbol{f}_{:,\findex'}|}\label{eq:CorrMet}
 \end{equation}
\subsection{Quadratic Problem}
\label{ssec:Quadratic}
This section presents further details on the quadratic problem and
the start solution $\boldsymbol{W}^{\mathrm{Start}}$ for the next iteration of solving the quadratic problem with updated constraints.
The start solution is a good, usually optimal, feasible solution for the currently selected set of features $\Lambda$.
To simplify the initial iterations,  only \cref{eq:NotGLob} is considered.
The deduplication of \cref{eq:4DedupQP} is only included after a solution is found that satisfies \cref{eq:NotGLob}.
The start solution is constructed from $\boldsymbol{W}^{\gls{nperClass}}$ which contains \gls{nperClass} assignments for each class to the most similar features in $\simmat{}_{:, \Lambda}$.
If the equal distribution of assignments per class is still exclusively optimized for, $\boldsymbol{W}^{\mathrm{Start}}=\boldsymbol{W}^{\gls{nperClass}}$ is already the start solution.
Else, we take care of all classes with equal assignment $C_{\mathrm{equal}}$ in $\boldsymbol{W}^{\gls{nperClass}}$.
Specifically, we remove all duplicate pairs $(\cindex, \cindex')\in C_{\mathrm{equal}}$:
\begin{align}
w^{\mathrm{Deduplication}}_{\cindex,\findex} &= \begin{cases}
\begin{aligned}
1  \quad\mathrm{if} \quad&(\cindex, \cindex')\in C_{\mathrm{equal}} \And\\ &(\cindex, \findex) = \mathrm{Maxi}(\cindex, \cindex') ,\\
\end{aligned}\\
\begin{aligned}
-1   \quad\mathrm{if}\quad &(\cindex, \cindex')\in C_{\mathrm{equal}} \And\\ &(\cindex,\findex) = \mathrm{Mini}(\cindex, \cindex') ,\\
\end{aligned}\\
0 \quad\mathrm{else}\\
\end{cases}\\
    \boldsymbol{W}^{\mathrm{Start}} &= \boldsymbol{W}^{\gls{nperClass}} +  \boldsymbol{W}^{\mathrm{Deduplication}}
\end{align}

Here, 
$\mathrm{Mini}(\cindex, \cindex')$ returns the indices to remove, that is of the current assignment with lowest similarity:
\begin{align}
 \helperindex{\cindex}= (\textbf{a}_{\cindex}&\circ \boldsymbol{w}^{\gls{nperClass}}_{\cindex}) \circ \fvecgurobi{} + \max(\textbf{a}_{\cindex}\circ \boldsymbol{w}^{\gls{nperClass}}_{\cindex}) \cdot ( \boldsymbol{1}-\fvecgurobi{})\\
 \mathrm{Mini}(\cindex, \cindex') &= \begin{cases}
   \cindex, \mathrm{argmin} (
  \helperindex{\cindex})  \quad\mathrm{if}\quad  \min(
     \helperindex{\cindex}) \leq \min(
     \helperindex{\cindex'})\\
    \cindex', \mathrm{argmin} (  \helperindex{\cindex'})
 \quad\mathrm{else}\\
    \end{cases}
\end{align}
Here, \fvecgurobi{} is the selection vector and ensures that all changes only apply to the selected features.
Similarly, $\mathrm{Maxi}(\cindex, \cindex')$ returns the indices of the assignment to add, which has the highest similarity of the the not currently assigned features:
\begin{align}
\helperindex{\cindex}= (\textbf{a}_{\cindex}&\circ \mathrm{\textbf{cand}}(\cindex)) \circ \fvecgurobi{}\\
 \mathrm{Maxi}(\cindex,\cindex') &= \begin{cases}
   \cindex, \mathrm{argmax} (
  \helperindex{\cindex})  \quad\mathrm{if}\quad  \max(
   \helperindex{\cindex} \geq \max(
   \helperindex{\cindex'})\\
    \cindex', \mathrm{argmax} ( \helperindex{\cindex'})
 \quad\mathrm{else}\\
    \end{cases}
\end{align}
The candidate function
\begin{align}
    \mathrm{\textbf{cand}}(\cindex) &=  (\boldsymbol{1}-\boldsymbol{w}^{\gls{nperClass}}_{\cindex})\cdot \mathrm{\textbf{wbnd}}(\cindex,\boldsymbol{w}^{\gls{nperClass}})
\end{align}
checks that the assignment is not made yet and the would-be-no-duplicate function $\mathrm{wbnd}(\cindex,\boldsymbol{W}^{\gls{nperClass}})_\findex\in\{0,1\}$
further ensures that the addition of the assignment of class $\cindex$ to feature $\findex$ would introduce no duplicate, returning $0$ in that case.
While this technically does not guarantee an optimal solution, first only finding the solution with \gls{nperClass} assignments per class and then deduplicating ensures that the number of duplicates is quite low already, which usually leads to finding the optimal feasible start solution.
\section{Steerability}
\label{ssec:steer}
  \begin{figure}[t]
     \centering\includegraphics[width=.6\linewidth]{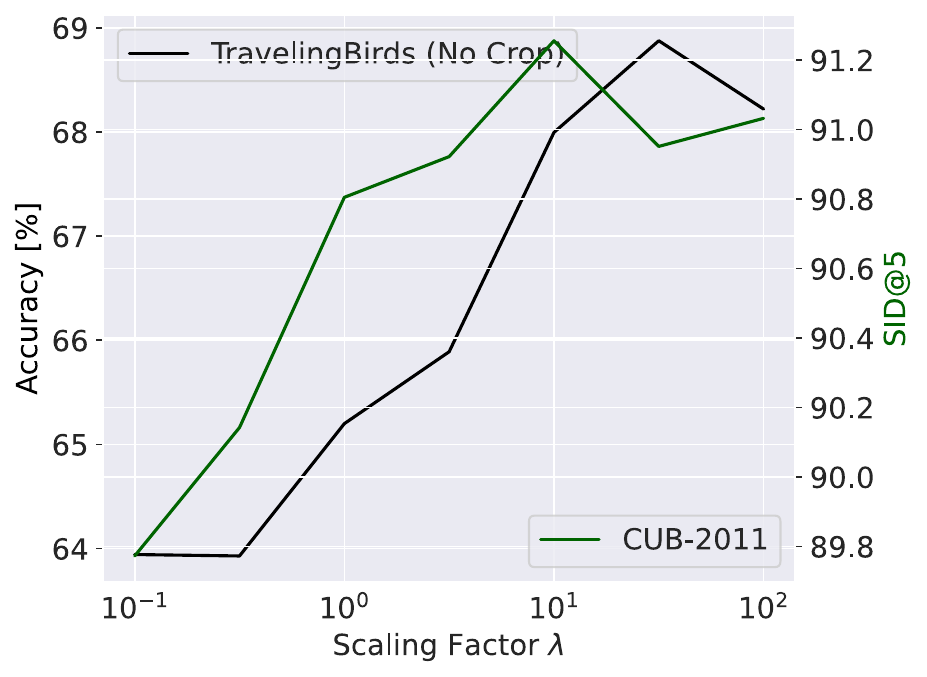}
     \caption{
     Steerability of the proposed \gls{NewlayerName}:
     When increasing the weighting of the bias $\lambda$, the desired metrics, accuracy or \loc{5} improve.}
     \label{fig:steering}
     \end{figure}

 This section is concerned with the ability of the practitioner to steer the model towards desired biases using the
 feature bias~\BSimeaturemat{}. 
 For example, if a human recognizes the erroneous focus on the background of a trained~\gls{NewlayerName}, enabled through global interpretability,
 the feature bias~$\BSimeaturemat{}^{\mathrm{Center}}$~(\cref{eq:centerBias}) can be used to steer the model towards more centered features.
\begin{equation}
  b^{\mathrm{Center}}_\findex = - \frac{1}{\gls{nTrainImages}\sum_j f_{j, \findex} } \sum_{j=1}^{\gls{nTrainImages}}  
  \frac{1}{1+\distEdge(\boldsymbol{M}^j_\findex)}
   f_{j,\findex}\label{eq:centerBias}
\end{equation}
where $\distEdge$ computes the distance between the maximum of 
\checktext{the $j$-th sample's map }
$\boldsymbol{M}^j_\findex$ at $(x,y)$ and the closest 
edge:
\begin{align}
  \distEdge(\textbf{M}_\findex^j) = \min(|x-\gls{featuresMapwidth}|,x-1, |y-\gls{featuresMapheigth}|, y-1)
\end{align}
 The resulting 
 improved accuracy on \travelingheader{} with $\lambda=10^\frac{3}{2}$, shown in \cref{stab:noCrop}, demonstrates this steerability.
Setting $\lambda$ allows a precise weighting of the emphasis put on the bias.
This direct control for both the center and diversity bias is visualized in \cref{fig:steering} and allows the incorporation
of any feature-level bias~\BSimeaturemat.
\section{Failure Cases}
\label{ssec:Failure}
\begin{figure}[h!t]
    \centering
    \begin{subfigure}{\textwidth}
        \centering     
\includegraphics[width=.24\textwidth]{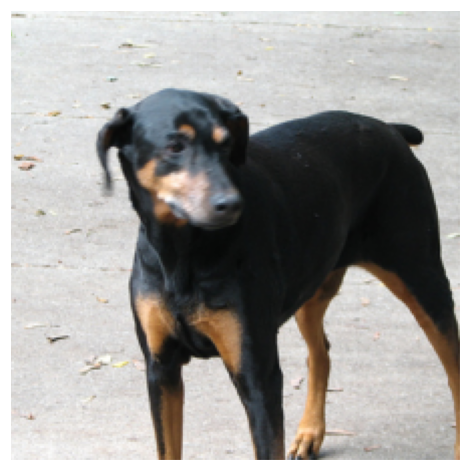}
        \hfill           \includegraphics[width=.24\textwidth]{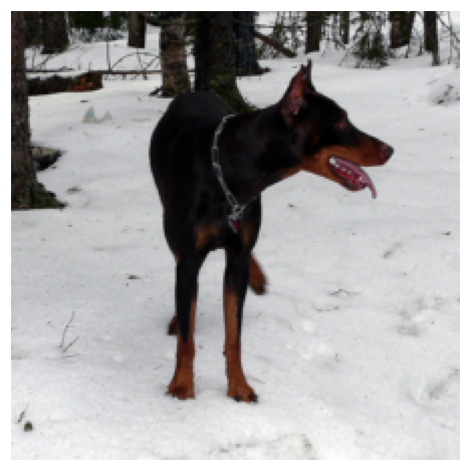}
        \hfill       \includegraphics[width=.24\textwidth]{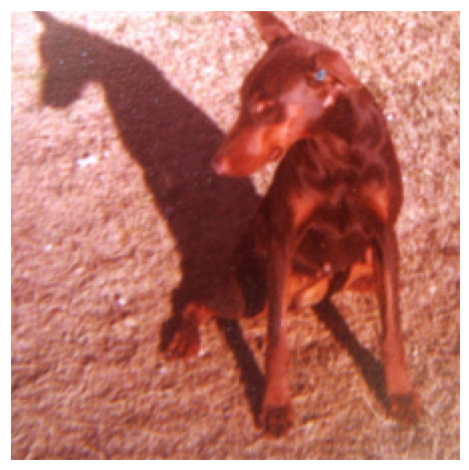}
        \hfill     \includegraphics[width=.24\textwidth]{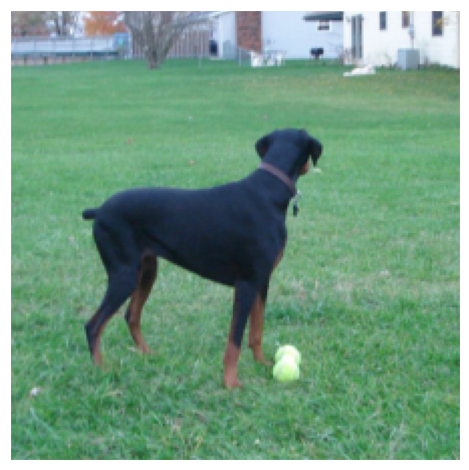}
        \caption{Correctly Classified Doberman Examples}
        \label{subfig:CDob}
        \end{subfigure}
        
        \vspace{0.3cm} %
        \begin{subfigure}{\textwidth}            \includegraphics[width=.24\textwidth]{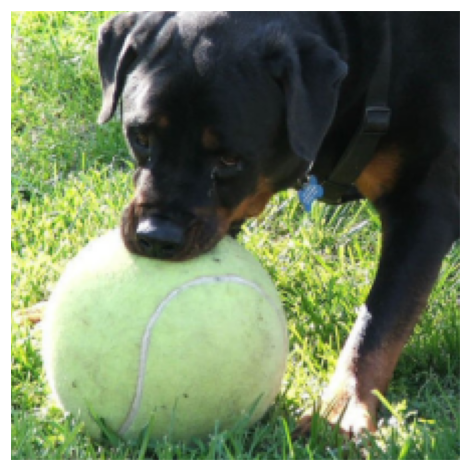}      
        \hfill     \includegraphics[width=.24\textwidth]{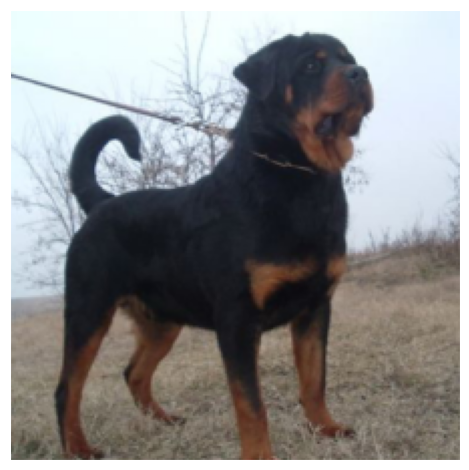}       
        \hfill     
\includegraphics[width=.24\textwidth]{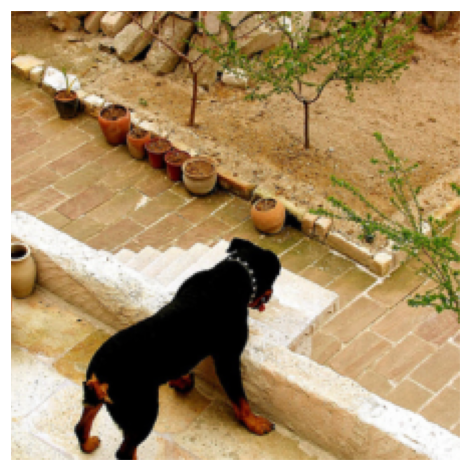}    
        \hfill     \includegraphics[width=.24\textwidth]{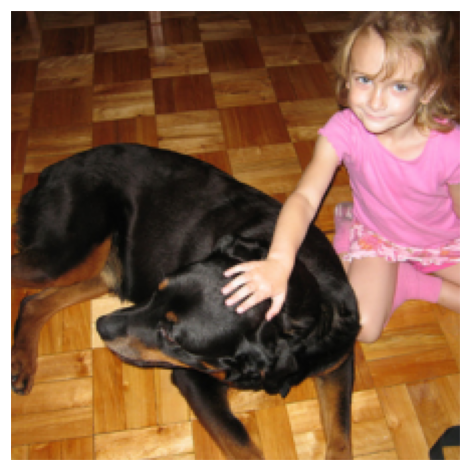}
        \caption{Correctly Classified Rottweiler Examples}
        \label{subfig:CRot}
        \end{subfigure}        
        
         \vspace{0.3cm}
        \begin{subfigure}[b]{0.24\textwidth}            \includegraphics[width=\textwidth]{plots/LocalExamplesCover/FalseDoberman/FeaturesRottweiler6.png}
            \caption{Wrongly classified as Rottweiler}
            \label{subfig:DRPCutoff}
        \end{subfigure}
        \hfill
        \begin{subfigure}[b]{0.24\textwidth}           \includegraphics[width=\textwidth]{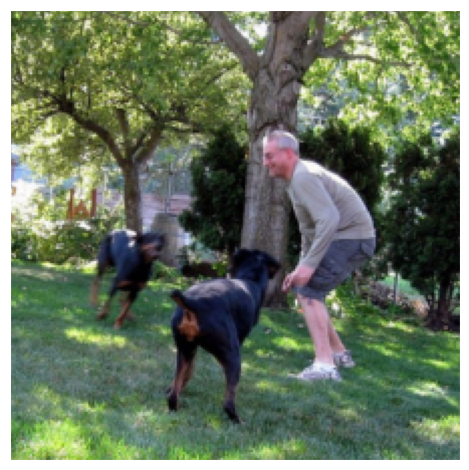}
            \caption{Wrongly classified as Rottweiler}
            \label{subfig:DRPPose}
        \end{subfigure}
        \hfill
        \begin{subfigure}[b]{0.24\textwidth}
        
            \includegraphics[width=\textwidth]{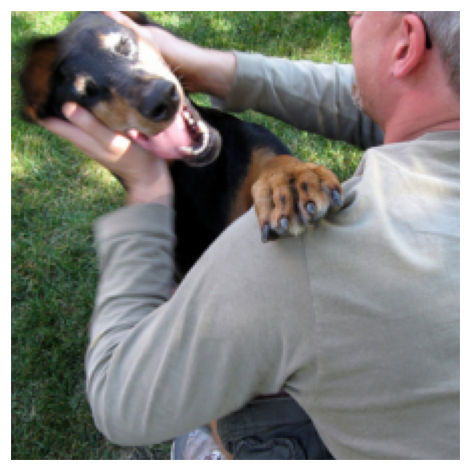}
            \caption{Wrongly classified as Black \& tan coonhound}
            \label{subfig:DBPCutoff}
        \end{subfigure}
        \hfill
        \begin{subfigure}[b]{0.24\textwidth}
            \includegraphics[width=\textwidth]{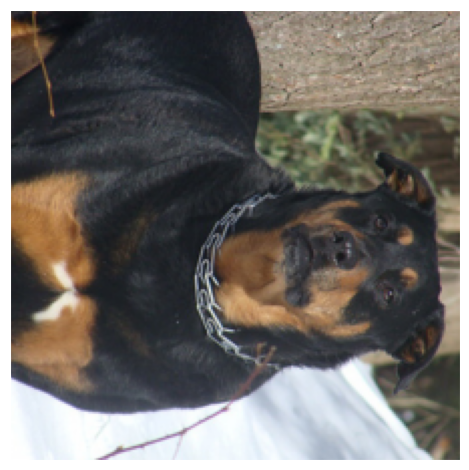}
            \caption{Classified as \\Rottweiler}\label{subfig:DRPLabel}
        \end{subfigure}
        \vspace{0.3cm} %
        \begin{subfigure}[b]{0.24\textwidth}
            \includegraphics[width=\textwidth]{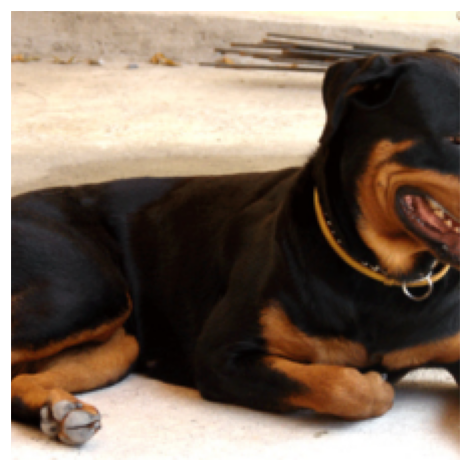}
            \caption{Wrongly classified as Doberman}
            \label{subfig:RDPCutoff}
        \end{subfigure}
        \hfill
        \begin{subfigure}[b]{0.24\textwidth}
            \includegraphics[width=\textwidth]{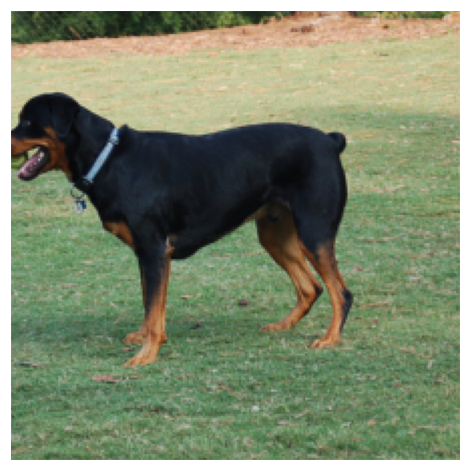}
            \caption{Wrongly classified as Doberman}
            \label{subfig:RDPPose}
        \end{subfigure}
        \hfill
        \begin{subfigure}[b]{0.24\textwidth}
            \includegraphics[width=\textwidth]{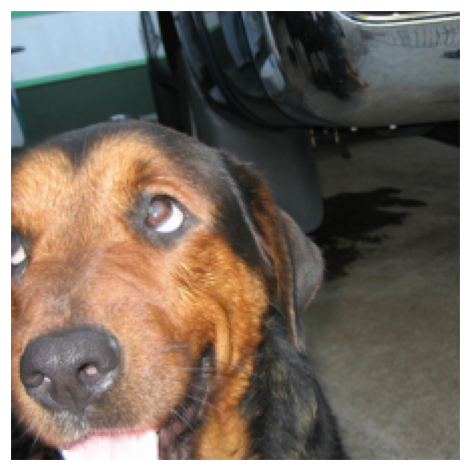}
            \caption{Wrongly classified as Black \& tan coon hound}
            \label{subfig:RBPCutoff}
        \end{subfigure}
        \hfill
        \begin{subfigure}[b]{0.24\textwidth}
            \includegraphics[width=\textwidth]{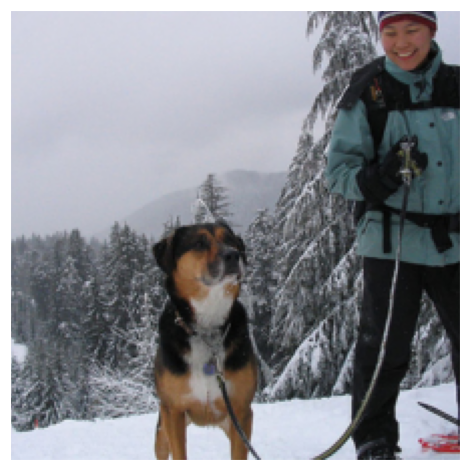}
            \caption{Classified as Greater Swiss Mountain dog}
            \label{subfig:RGPLabel}
        \end{subfigure}
    \caption{Examples for correctly and wrongly (according to ground truth labels) classified examples of the \gls{NewlayerName} with global explanations shown in
    ~\cref{fig:metrics_full,fig:blacktanS,fig:rottweilerS,fig:dobermanS}.
    \Cref{subfig:CRot,subfig:DRPCutoff,subfig:DRPPose,subfig:DBPCutoff,subfig:DRPLabel} (rows $1$ and $3$) show Doberman labeled images. \Cref{subfig:CDob,subfig:RDPCutoff,subfig:RDPPose,subfig:RBPCutoff,subfig:RGPLabel}  (rows $2$ and $4$) display Rottweiler labeled images. The resulting classifications match the expected behavior based on the global explanations.
    As the explained \gls{NewlayerName} uses the head to differentiate between Doberman and Rottweiler (\cref{fig:metrics_full}), they can be confused when it is occluded (\cref{subfig:DBPCutoff,subfig:DRPCutoff}) or in a difficult pose (\cref{subfig:DRPPose,subfig:RDPPose}). As the black and tan coon hound is assigned the same head features (\cref{fig:blacktanS}), they get confused, if only the head is visible (\cref{subfig:DBPCutoff,subfig:RBPCutoff}).
    Finally, the probed \gls{NewlayerName} correctly classifies according to its explanations (\cref{subfig:CRot,subfig:CDob}), also on wrongly labeled samples (\cref{subfig:DRPLabel,subfig:RGPLabel}).
    }
    \label{fig:failuresOver}
\end{figure}

\begin{figure}[h!t]
    \centering
    \begin{subfigure}{\textwidth}
        \centering     
\includegraphics[width=.24\textwidth]{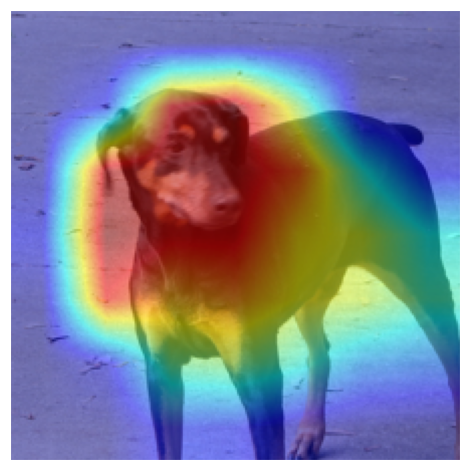}
        \hfill           \includegraphics[width=.24\textwidth]{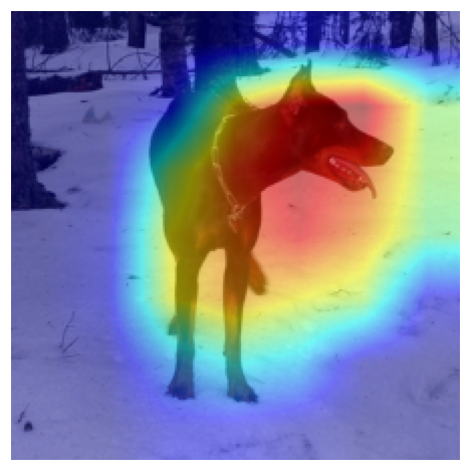}
        \hfill       \includegraphics[width=.24\textwidth]{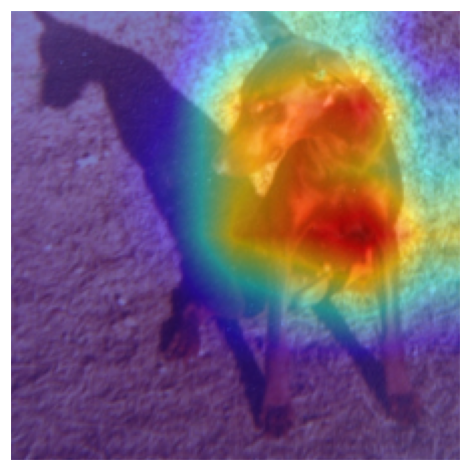}
        \hfill     \includegraphics[width=.24\textwidth]{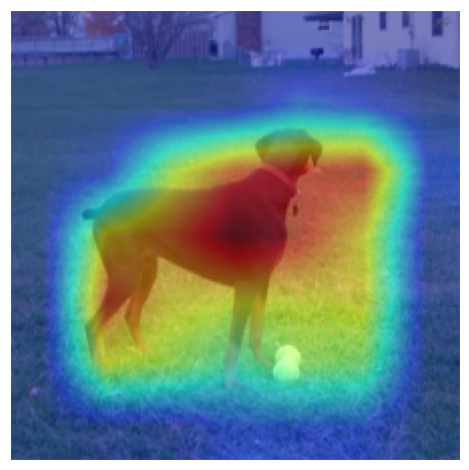}
        \caption{Correctly Classified Doberman Examples}
        \label{Gradsubfig:CDob}
        \end{subfigure}
        
        \vspace{0.3cm} %
        \begin{subfigure}{\textwidth}            \includegraphics[width=.24\textwidth]{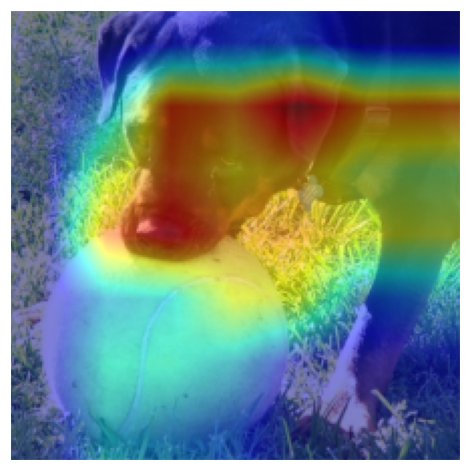}      
        \hfill     \includegraphics[width=.24\textwidth]{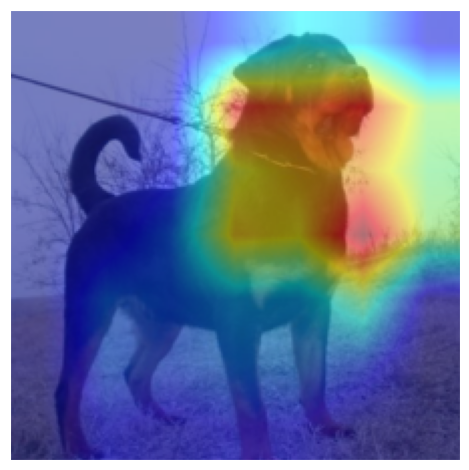}       
        \hfill     
\includegraphics[width=.24\textwidth]{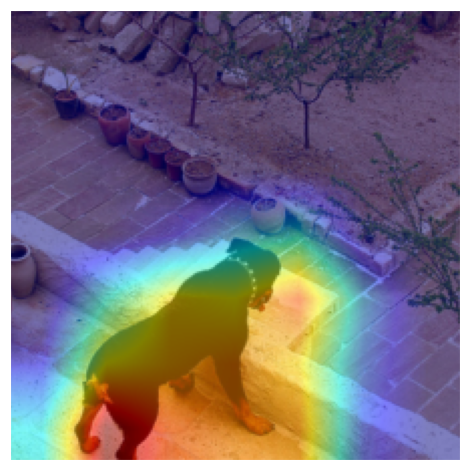}    
        \hfill     \includegraphics[width=.24\textwidth]{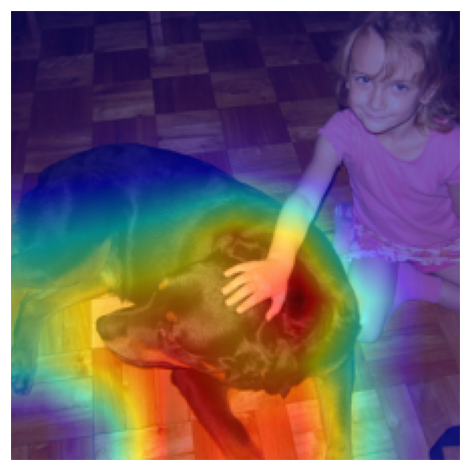}
        \caption{Correctly Classified Rottweiler Examples}
        \label{Gradsubfig:CRot}
        \end{subfigure}        
        
         \vspace{0.3cm}
        \begin{subfigure}[b]{0.24\textwidth}            \includegraphics[width=\textwidth]{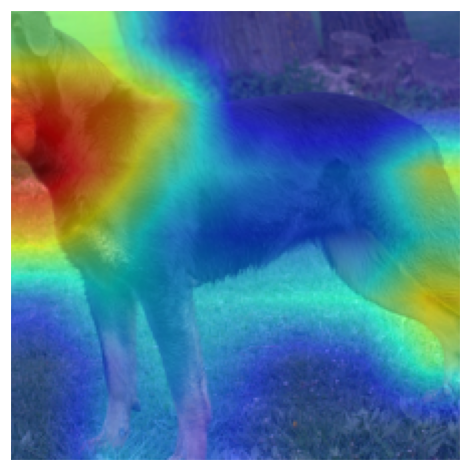}
            \caption{Wrongly classified as Rottweiler}
            \label{Gradsubfig:DRPCutoff}
        \end{subfigure}
        \hfill
        \begin{subfigure}[b]{0.24\textwidth}           \includegraphics[width=\textwidth]{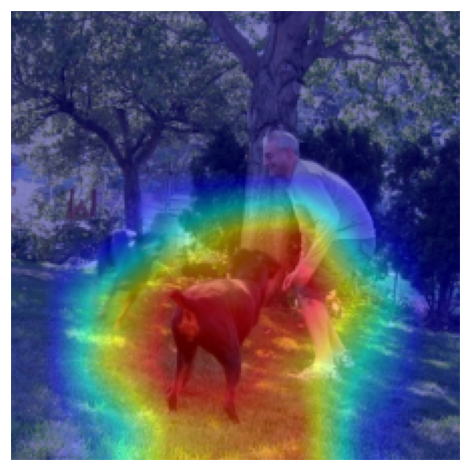}
            \caption{Wrongly classified as Rottweiler}
            \label{Gradsubfig:DRPPose}
        \end{subfigure}
        \hfill
        \begin{subfigure}[b]{0.24\textwidth}
        
            \includegraphics[width=\textwidth]{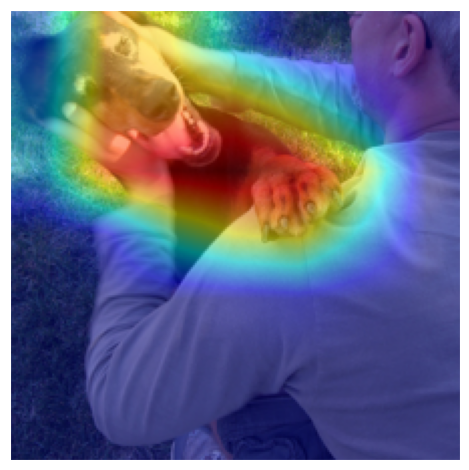}
            \caption{Wrongly classified as Black \& tan coonhound}
            \label{Gradsubfig:DBPCutoff}
        \end{subfigure}
        \hfill
        \begin{subfigure}[b]{0.24\textwidth}
            \includegraphics[width=\textwidth]{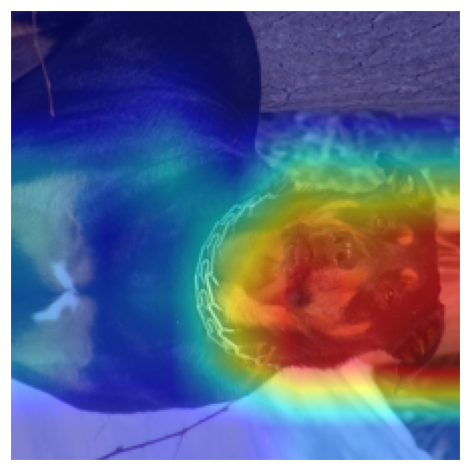}
            \caption{Classified as \\Rottweiler}\label{Gradsubfig:DRPLabel}
        \end{subfigure}
        \vspace{0.3cm} %
        \begin{subfigure}[b]{0.24\textwidth}
            \includegraphics[width=\textwidth]{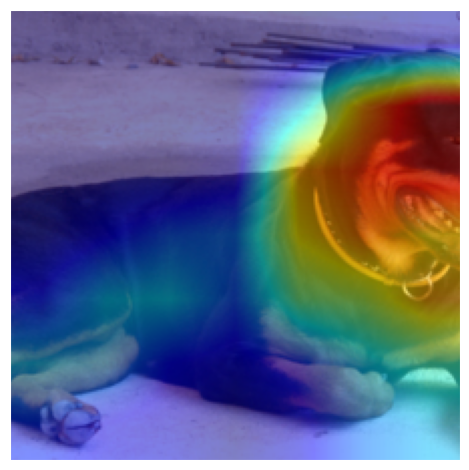}
            \caption{Wrongly classified as Doberman}
            \label{Gradsubfig:RDPCutoff}
        \end{subfigure}
        \hfill
        \begin{subfigure}[b]{0.24\textwidth}
            \includegraphics[width=\textwidth]{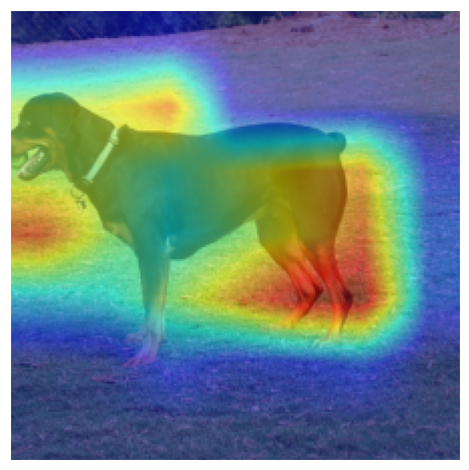}
            \caption{Wrongly classified as Doberman}
            \label{Gradsubfig:RDPPose}
        \end{subfigure}
        \hfill
        \begin{subfigure}[b]{0.24\textwidth}
            \includegraphics[width=\textwidth]{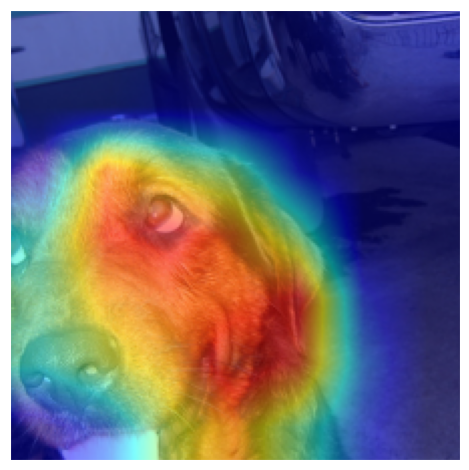}
            \caption{Wrongly classified as Black \& tan coon hound}
            \label{Gradsubfig:RBPCutoff}
        \end{subfigure}
        \hfill
        \begin{subfigure}[b]{0.24\textwidth}
            \includegraphics[width=\textwidth]{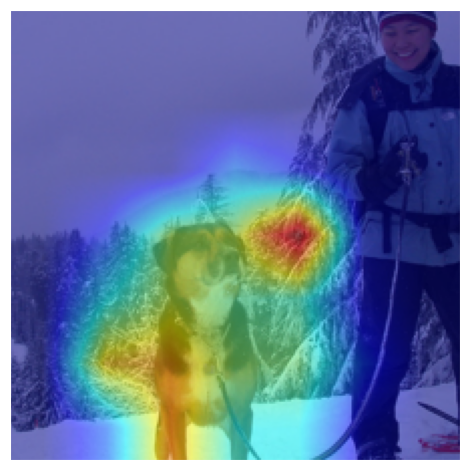}
            \caption{Classified as Greater Swiss Mountain dog}
            \label{Gradsubfig:RGPLabel}
        \end{subfigure}
 \caption{Gradcam Visuliazations for \cref{fig:failuresOver}.}
    \label{Gradfig:failuresOver}
\end{figure}

\begin{figure*}[t]
  \centering
\includegraphics[width=\linewidth]{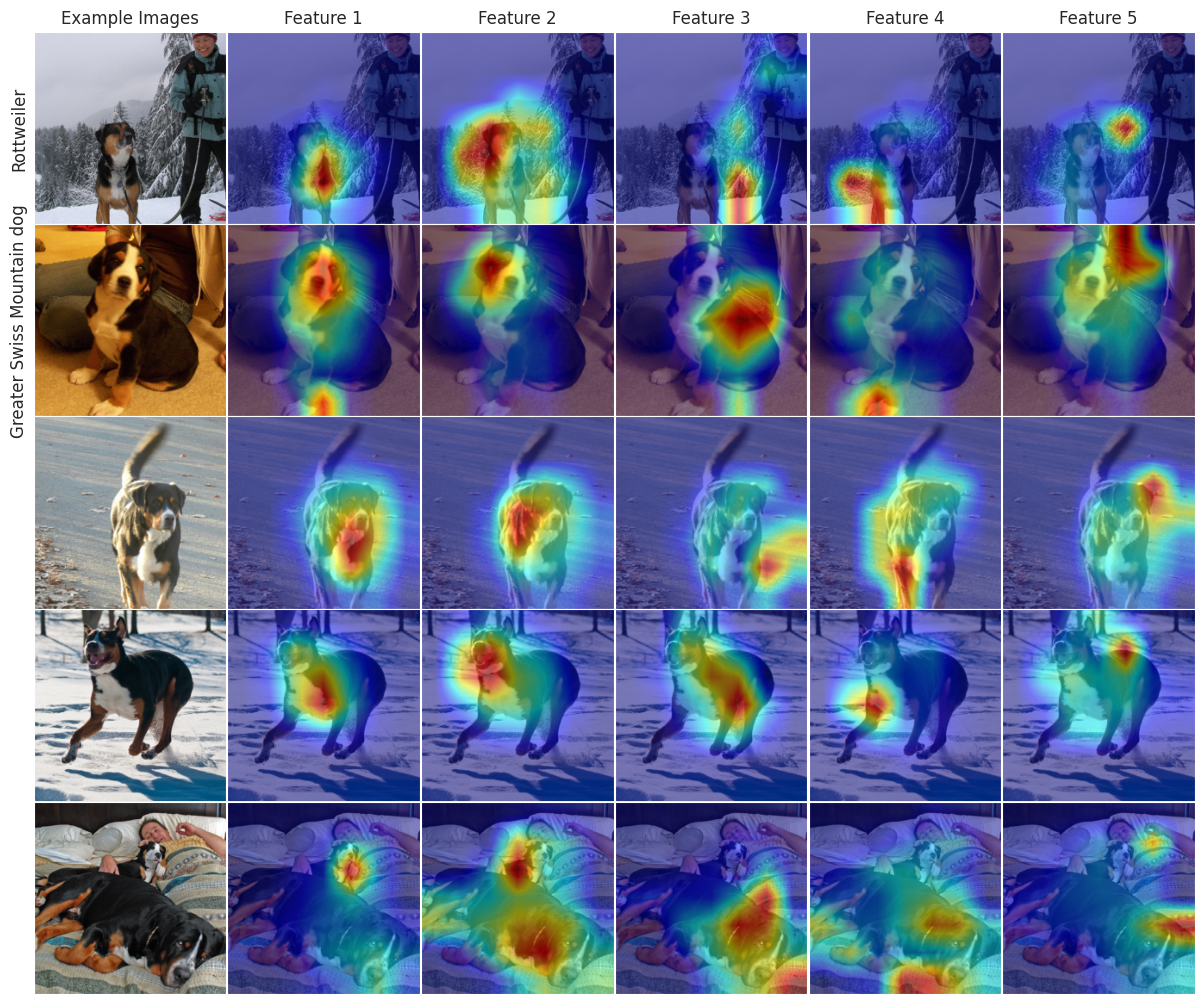}

   \caption{Features of Greater Swiss Mountain Dog and their localization on the sample in \cref{subfig:RGPLabel} (first row), that is presumably falsely labeled Rottweiler. The 4 lower rows contain examples of Greater Swiss Mountain Dog and the features consistently localize around semantically similar regions, also on the Rottweiler labeled one. }
  \label{subfig:Mountain}
\end{figure*}

\begin{figure*}[t]
  \centering
\includegraphics[width=\linewidth]{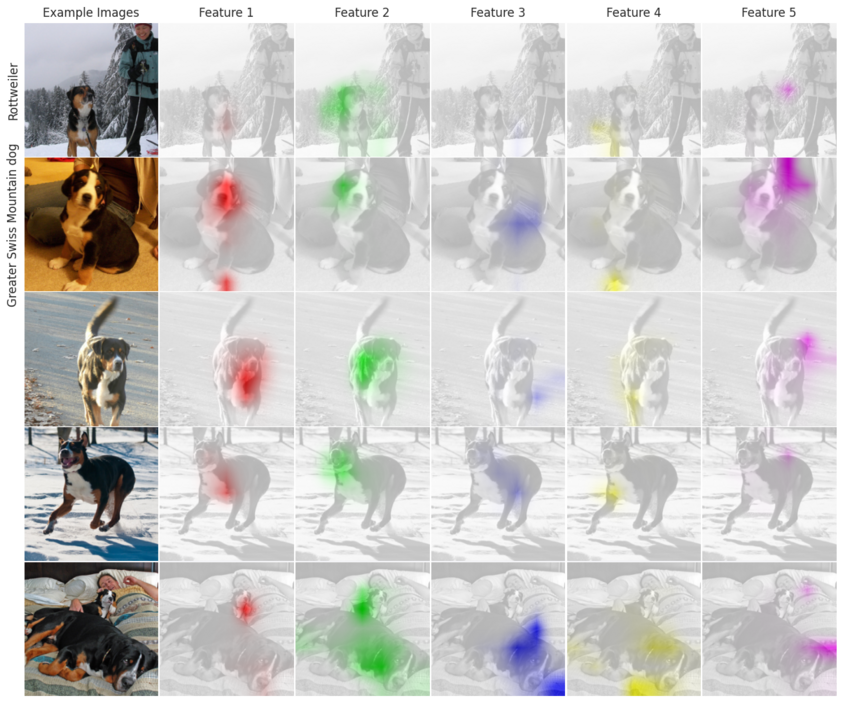}

   \caption{Features of Rottweiler and their localization, scaled by column, on the sample in \cref{subfig:RGPLabel} (first row), that is presumably falsely labeled Rottweiler. The 4 lower rows contain examples of Rottweiler. The color code is consistent with \cref{fig:metrics_full} and features $4$ and $5$ are shared with Greater Swiss Mountain Dog.}
  \label{subfig:MountainRott}
\end{figure*}

\begin{figure*}[t]
  \centering
\includegraphics[width=\linewidth]{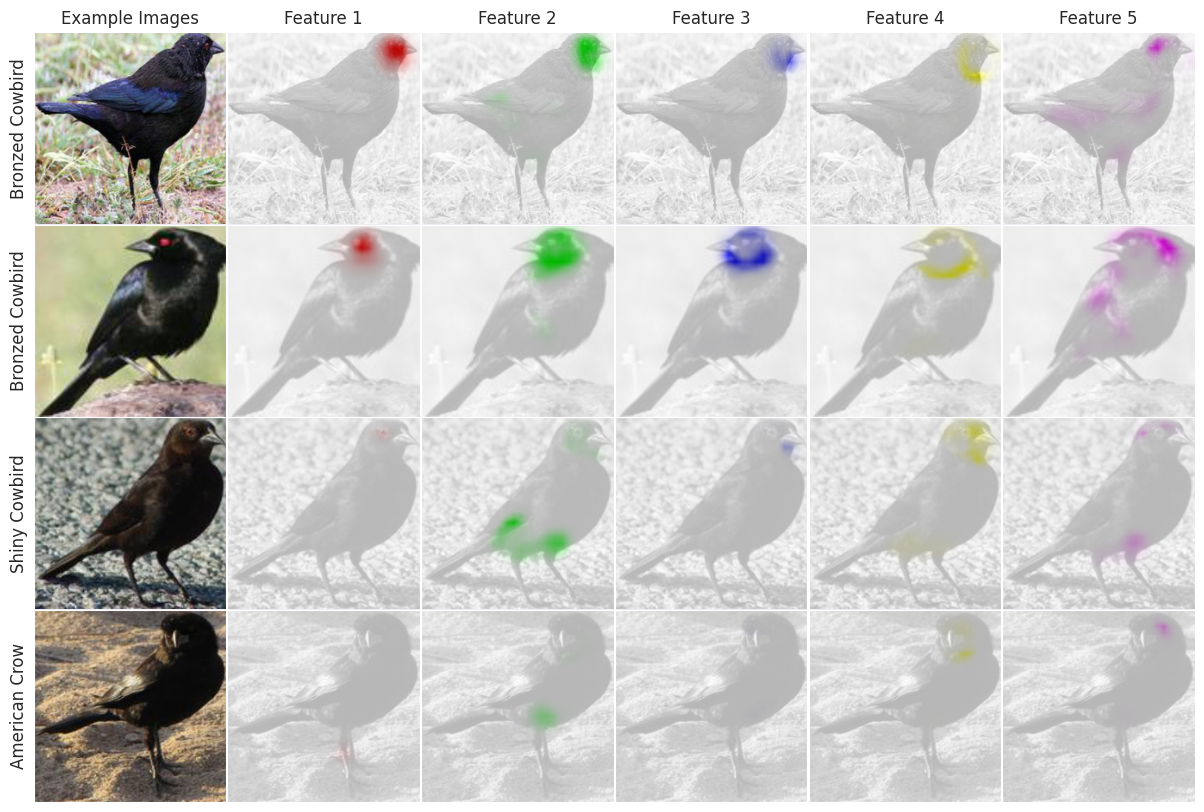}

   \caption{Features of Bronzed Cowbird, explained and compared with Shiny Cowbird in \cref{fig:CubSim}, the predictions of the QPM, and their localizations, normed across column, on Bronzed Cowbird labeled test samples. When all features, including the red eye (feature 1) are visible (rows 1 and 2), the model is correct. However, as expected from the global explanation, without the red eye it can be wrong and confuse e.g. Shiny Cowbird with it. The probed QPM represents American Crow with features 2,4,5 and 2 further not shown features, that localize on wing and beak of crows.  }
  \label{subfig:BronzedEx}
\end{figure*}

This section presents examples where \gls{NewlayerName} predicts wrongly.
For that, \cref{fig:failuresOver} shows exemplary images of Rottweiler and Doberman with classification results of the probed \gls{NewlayerName} trained on \imgnetheader{} and with global explanations in \cref{fig:metrics_full,fig:blacktanS,fig:rottweilerS,fig:dobermanS}. 
Note that the accuracy across the two classes is $87\%$, well above the average, reflected in correct classifications across poses, backgrounds and settings in \cref{subfig:CDob,subfig:CRot}.
Additionally, \cref{Gradfig:failuresOver} shows the GradCAM~\citep{selvaraju2020grad} visualizations and demonstrates that QPM always focuses on the dog in the image.
For the erroneous predictions, the model behaves just like the global explanations would indicate.
Rottweiler and Doberman may be swapped, if the head is occluded as in \cref{subfig:DRPCutoff,subfig:RDPCutoff} or in a difficult pose to gauge the shape, shown in \cref{subfig:DRPPose,subfig:RDPPose}. 
Since the Black and tan coon hound is assigned both head features of Rottweiler and Doberman, they can also be confused when primarily the head is visible, demonstrated in \cref{subfig:DBPCutoff,subfig:RBPCutoff}.
Finally, \cref{subfig:DRPLabel,subfig:RGPLabel} seem to contain one of the many~\citep{northcutt2021labelerrors} wrongly  labeled samples in \imgnetheader{}.  \gls{NewlayerName} also robustly classifies wrongly labeled data, as the global explanation would suggest.
\Cref{subfig:Mountain,subfig:MountainRott} show the feature activations of Greater Swiss Mountain Dog and Rottweiler on \cref{subfig:DRPLabel} and other class examples, further suggesting that it is indeed a typical Greater Swiss Mountain rather a Rottweiler for the probed QPM, as the features of the former localize  on the expected regions, whereas most Rottweiler features barely activate.
Finally, \cref{subfig:BronzedEx} shows further test examples for the model explained in \cref{fig:CubSim} and demonstrates that the model does not predict Bronzed Cowbird if the differentiating red eye is not present in the image.
In summary, \gls{NewlayerName}'s local behavior robustly follows the faithful global explanations, which can lead to predictable faulty classifications in case of occlusion or difficult pose.

\section{Runtime Analysis}
\label{ssec:Runtime}
\begin{figure*}[t]
  \centering
\includegraphics[width=.5\linewidth]{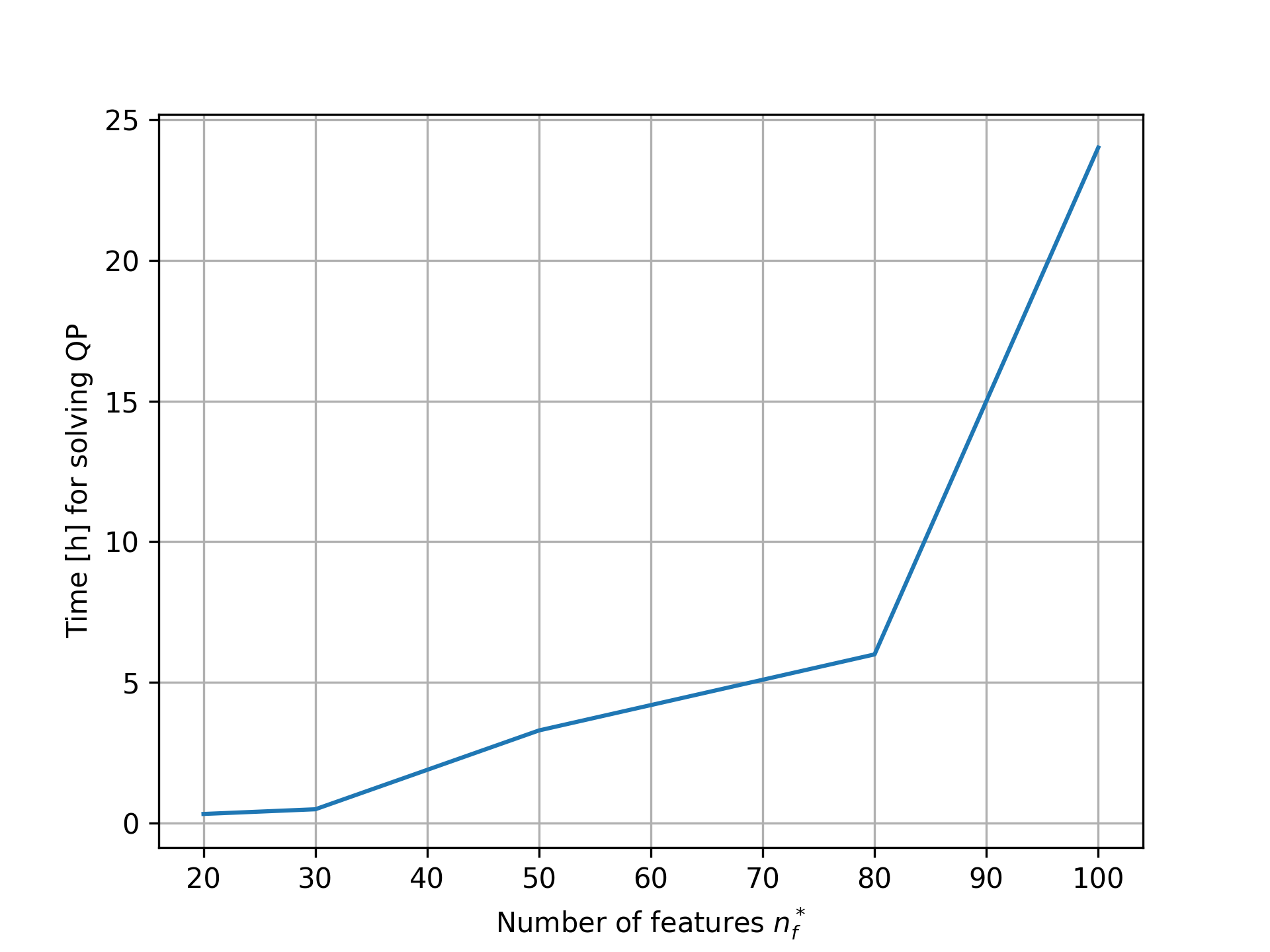}

   \caption{Time it takes to optimize QP for models with varying \gls{nReducedFeatures}in~\cref{fig:interptradeoff}.}
  \label{fig:timeplot}
\end{figure*}

This section discusses the time it takes to obtain a \gls{NewlayerName}, compares it to competing models and discusses the impact of \gls{nReducedFeatures} on it.
\Cref{fig:timeplot} demonstrates that the optimization time strongly increases when increasing \gls{nReducedFeatures}. 
However, for the probed datasets, going beyond $50$ features seems not to be necessary, as the accuracy only improves negligibly, while the interpretability is harmed: Features become less general and there will be fewer class representations with high overlap, which allow for the most intuitive interpretation.
One can further optimize this using suitable priors, which we do not include in this work, as the interpretability and additional accuracy decreases with increasing \gls{nReducedFeatures}.
It is however an avenue for future work, when datasets with sufficient complexity are published.
\Cref{stab:TimeTable} compares the time to obtain the interpretable model between \gls{NewlayerName}, \gls{qsenn} and \gls{layerName}. 
\gls{qsenn} and \gls{layerName} start with a feature selection, that takes $15$ minutes on \cubheader{} and roughly $500$ minutes on \imgnetheader{}. They both use \glm{} for feature selection and computing the sparse matrix and are thus scaling with number of samples \gls{nTrainImages}, which \gls{NewlayerName} is invariant to, as that dimension is summarized in the constants.
\begin{table*}[t]
 \caption{Time in minutes between finishing dense training and obtaining the final model. First value is time for optimization, second time spent fine-tuning. Following \gls{qsenn}, we use the \textit{fast} setting for \imgnetheader{}. \gls{qsenn} trains for $70$ epochs, instead of $40$, in total during fine-tuning and does $4$ iterations of \glm{}. Note that every method runs exclusively on a GPU server, except for the QP optimization, which can be done on just a CPU.}
 \label{stab:TimeTable}
 \centering
 \begin{tabular}{l|cc}
  \toprule
  Method 
  & CUB&INET \\
  \midrule
  \slddtable{} & $(15 + 22) + 78 = \textbf{115}$  & $(500 + 3000) + 3600 = 7100$  \\
   \qsenntable{} &$(15 + 4*22) + 78 *7/4 \approx  240$  & $(500 + 4* 100) + 7/4 * 3600 = 7200$  \\\midrule
   \gls{NewlayerName} (Ours) & $210 + 78 = 298$  & $660 + 3600 = \textbf{4260}$  \\ 
  \bottomrule
 \end{tabular}
 \end{table*}
 \section{Impact of Even Sparsity}
 \label{ssec:balancedAbl}
 \begin{table*}[t]
 \caption{Accuracy with or without exactly \gls{nperClass} features per class (\cref{eq:5perInit}). Instead, on average  \gls{nperClass} features per class are used.}
 \label{stab:evenConstrR}
 \centering
 \begin{tabular}{l|cc}
  \toprule
  Method 
  & CUB&CARS \\
  \midrule
  without \cref{eq:5perInit} & 84.3$\pm0.2$   & 91.6$\pm0.3$   \\
   \gls{NewlayerName} (Ours) & \textbf{85.1}$\pm0.3$   &\textbf{ 91.8}$\pm0.1$   \\ 
  \bottomrule
 \end{tabular}
 \end{table*}
 \begin{figure*}[t]
  \centering
\includegraphics[width=.5\linewidth]{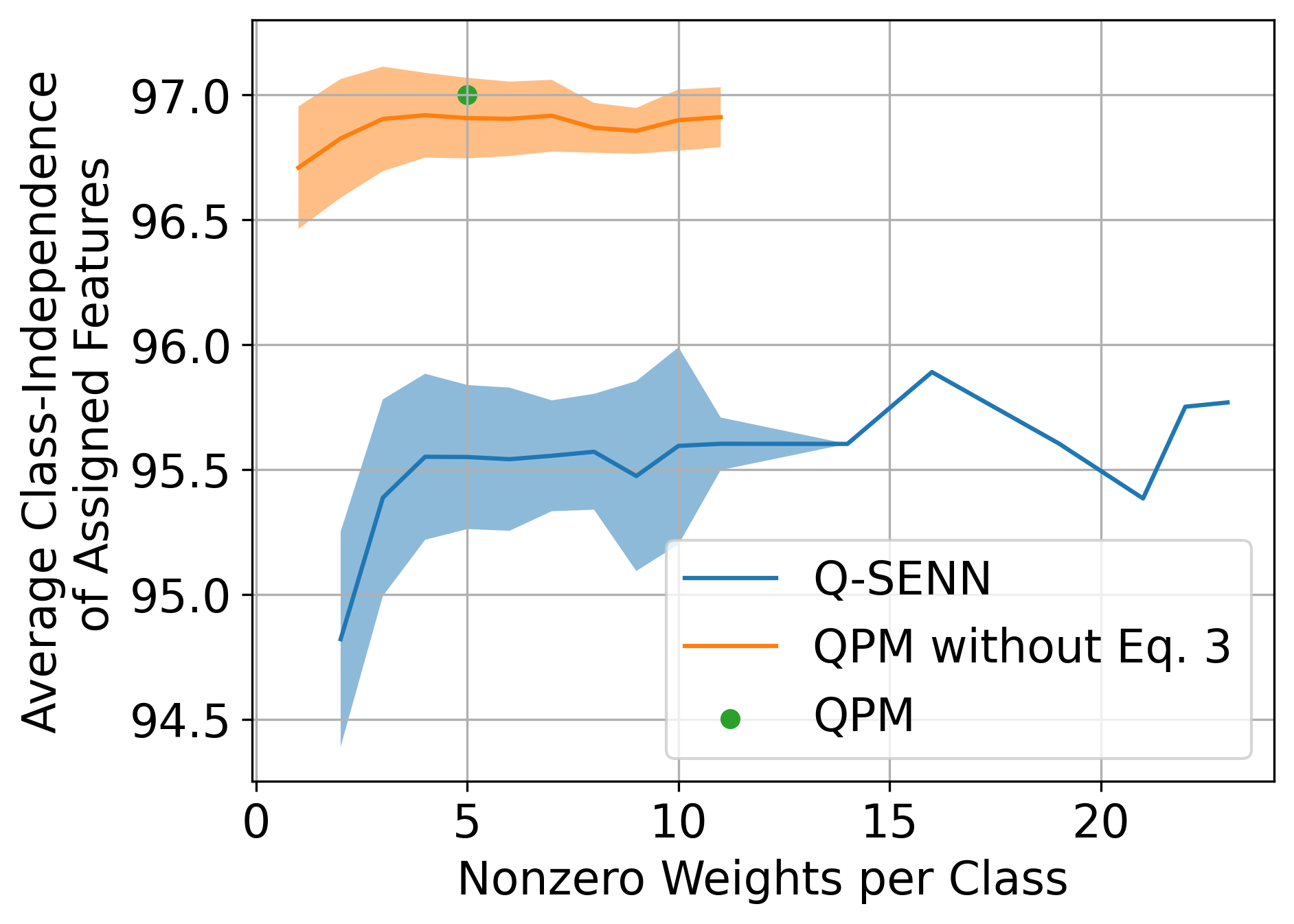}

   \caption{Average Class-Independence of Assigned Features on \cubheader{} as function of the number of features assigned to the class. The distribution of the sparsity is shown in \cref{sfig:densityS}.}
  \label{sfig:CIOverNWC}
\end{figure*}
 \begin{figure*}[t]
  \centering
\includegraphics[width=.5\linewidth]{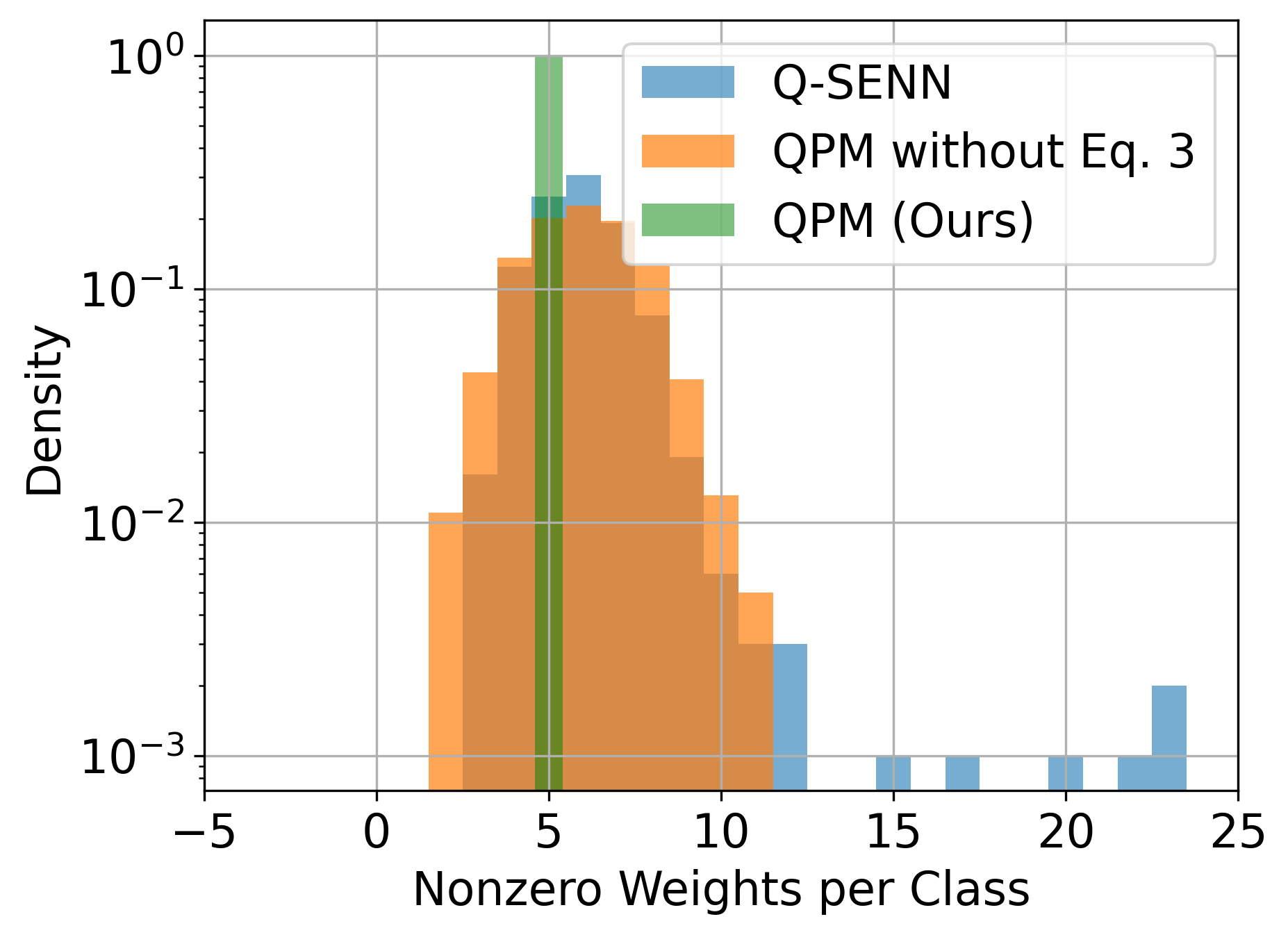}

   \caption{Distribution of Nonzero Weights per Class on \cubheader{}. For each method, all $1000$ classes from $5$ seeds are shown.}
  \label{sfig:densityS}
\end{figure*}
 This section discusses the impact of enforcing exactly \gls{nperClass} features per class, rather than on average. 
 For that, we trained a model without this constraint, but instead with $\boldsymbol{1}^T\wgurobi\fvecgurobi=\gls{nperClass}\gls{nClasses}$ enforcing an average sparsity.
 To counteract the uneven number of features per class, every class got a bias, that is linear to the number of features it is below the average.
 In prior experiments, various forms of counteracting the uneven assignment with a bias have performed similarly.
 \Cref{stab:evenConstrR} shows that the even assignment is beneficial for the accuracy.
 Further, the even assignment boosts interpretability as it leads to more classes that can be contrasted easily and does not introduce an unintuitive bias term. 
 Additionally, \cref{sfig:CIOverNWC} demonstrates that classes, which are assigned to fewer features, cause these features to become less general for QPM and Q-SENN, which hurts interpretability and potentially accuracy.
 \Cref{sfig:densityS} also visualizes that Q-SENN always learns to represent classes with a huge variety in the number of assigned features, necessarily leading to hardly interpretable representations.
 Nevertheless, the impact is disparate on the two datasets and the accuracy  increase is not significant on \stanfordheader{}. 
 Future work might investigate if datasets with classes of varying complexity will benefit from representing classes with a suitable number of features and how this can be combined with contrastive globally interpretable class representations.

\section{Polysemantic Features}
\label{ssec:poly}
 \begin{figure*}[t]
  \centering
\includegraphics[width=.5\linewidth]{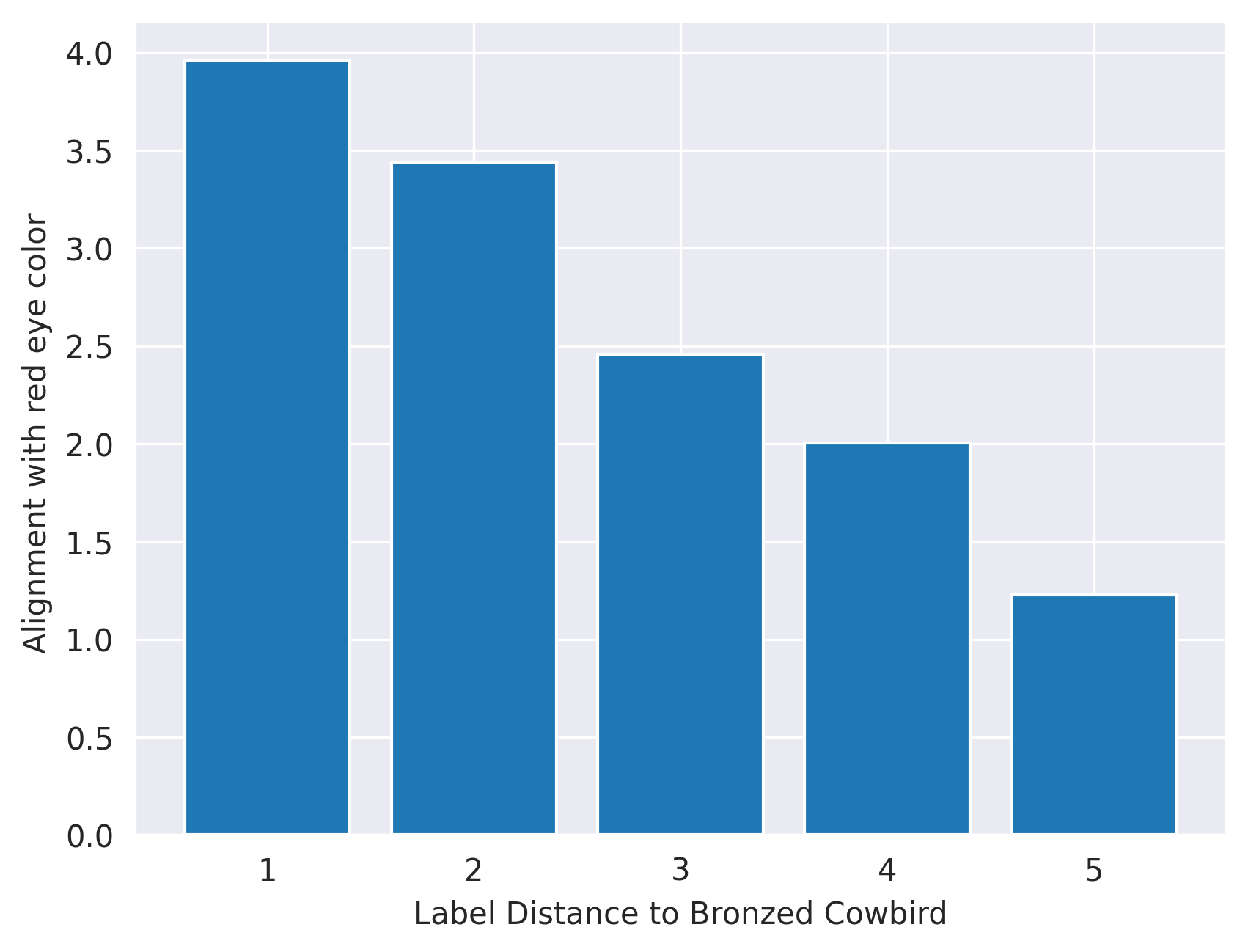}

   \caption{Feature Alignment metric from Q-SENN~\cite{norrenbrock2024q} for the red-eye feature (marked in red) in \cref{fig:CubSim} and the attribute red eye color $r_{red-eye} (x)$. The x-axis describes to which samples the computation is limited, e.g. $x=2$ describes computing the metric only on samples whose label is represented using up to $2$ other features. On the right, $x=5$ refers to the usual global feature alignment. The probed QPM learned a polysemantic, but locally monosemantic feature. When differentiating between bronzed and shiny cowbird only ($x=1$), the feature value increased by almost $4$ times its mean, if the attribute is annotated to be present.}
  \label{sfig:Alignment}
\end{figure*}
This section discusses the phenomenon of polysemantic features and how it relates to QPM.
Like all deep learning models~\citep{scherlis2022polysemanticity} not specifically designed to prevent polysemanticity, QPM learns polysemantic features.
It refers to individual neurons activating on not just one concept $c$ but rather on $n$ seemingly unrelated ones.
While it is an active area of research, their emergence can likely be attributed to being an effective solution to the training objective.
On many training samples, the impact on the loss can be fairly low, if a polysemantic feature activates on any of its $n$ meanings. 
The only exception occurs, when it activates on samples, where its activation contributes significantly to a class that is already showing a lot of activation.
While this is typically very difficult to analyze, the interpretable structure of QPM can offer more insights, as it enables a reliable metric on which to gauge how strongly the activation on another concept would affect the loss: The similarity in QPM's class representation space. Our hypothesis is that QPM learns features that are locally monosemantic, while being globally polysemantic. Around a class, e.g., Bronzed Cowbird, we expect the features to only activate on one of the $n$ concepts that they activate on across the entire dataset. 
As this is generally fairly difficult to measure, we show anecdotal evidence for this in \cref{sfig:Alignment}. 
It shows the Feature Alignment metric from \gls{qsenn}\citep{norrenbrock2024q} relative to the similarity to the Bronzed Cowbird, measured as the number of its features that classes do not share. 
Specifically, given the training features \glspl{trainFeatures},
\begin{align}
    \gtmatrix_{a,j}&=
\frac{1}{\lvert\attributeset{a+}\rvert }\sum_{i\in\attributeset{a+}}\glspl{trainFeatures}_{i,j}- \frac{1}{\lvert\attributeset{a-}\rvert}\sum_{i\in\attributeset{a-}}\glspl{trainFeatures}_{i,j}
\end{align}
describes the average difference in activations when an annotated attribute is present, encoded in \attributeset{a+} or \attributeset{a-} for absent.
\cite{norrenbrock2024q} then scales the difference by the average zero based activation and reports the average maximum per feature:
\begin{align}
    r = \frac{1}{\gls{nReducedFeatures}}\sum_{j=1}^{\gls{nReducedFeatures}} \frac{\gls{nTrainImages}}{\sum_{l=1}^{\gls{nTrainImages}}\glspl{trainFeatures}_{l,j}- \min_{l}\glspl{trainFeatures}_{l,j} }\max_{i} \gtmatrix_{i,j}.
\end{align}
For our analysis, we limit these formulas to just the attribute red eye color $red-eye$ and only consider the one feature $k$ detecting it for Bronzed Cowbird:
\begin{align}
    r_{red-eye} (x) = \frac{\gls{nTrainImages}}{\sum_{l=1}^{\gls{nTrainImages}}\glspl{trainFeatures}_{l,k}- \min_{l}\glspl{trainFeatures}_{l,k} }\gtmatrix_{red-eye,k}(x).
\end{align}
The x-axis additionally describes a filtering applied to the features and attributes based on the similarity of the label, where a sample is considered for computing $\gtmatrix_{red-eye,k}(x)$ if the annotated label shares at least $5-x$ features with Bronzed Cowbird.
\Cref{sfig:Alignment} demonstrates that the feature clearly detecting the red eye of the Bronzed Cowbird is indeed quite sensitive to its presence when the ground truth label is similar to the class, while it globally loses that sensitivity as it also detects other concepts of classes further away from Bronzed Cowbird.

\section{Structural Grounding on ImageNet}
\label{ssec:imgsim}
This section is concerned with evaluating a metric similar to \cubsim{} on \imgnetheader{}.
It is based on comparing the class similarities in reality  $\boldsymbol{\mathrm{\ClassSim}}^{gt}$ with the $\boldsymbol{\mathrm{\ClassSim}}^{Model}$ ones learned by our model.
\cubsim{} relies on the annotations of \cubheader{} to compute $\boldsymbol{\mathrm{\ClassSim}}^{gt}$. However, there are no such annotations on a fine-grained scale for \imgnetheader{}.
Therefore, we use the similarity of the text-names in CLIP~\citep{radford2021learning} as proxy to obtain our ground-truth class similarities. 
Specifically, we compute the cosine similarity $\boldsymbol{\mathrm{\ClassSim}}^{clip}\in[-1,1]^{\gls{nClasses}\times\gls{nClasses}}$  between the text embeddings of the class names, obtained from the powerful pretrained \textit{ViT-L-14}, that is broadly used, e.g. to condition Stable Diffusion XL\cite{podell2023sdxl}. 
We always take the first description given for every class.

When inspecting the most similar classes, several issues are apparent. 
Many of them include shared tokens or words, e.g., \textit{ski} and \textit{ski mask}, \textit{lion} and \textit{sea lion}, \textit{rule} and \textit{stole} or \textit{digital clock} and \textit{wall clock}. While some of these indeed describe a similar class, e.g., \textit{giant Schnauzer} and \textit{Standard Schnauzer}, others do not. 
Including classes without high similarity as ground-truth similar classes harms the quality of the evaluation drastically and demonstrates the value of having human annotations.
Another issue in the clip similarities is that fine-grained knowledge about the classes seems to be less dominant than literal exact matching, as with higher similarity only more commonly used terms are correctly associated with similar terms,  e.g., \textit{Orangutan} and \textit{Gorilla}, but not \textit{Rottweiler} and \textit{Doberman}. Therefore, the number of similar classes to consider is set to $1250$, as the latter pair ranks at position $828$ and we definitely consider it as a similarity worth measuring.
Notably, this pair is ranked behind pairs such as \textit{hog} and \textit{tank},  \textit{lemon} and \textit{yawl} or \textit{hamster} and \textit{snail}, further demonstrating the weakness of the language model to exactly model the similarities.
The final apparent issue lies in the ambiguity of class names which leads to \textit{crane} appearing twice as class name, once referring to birds, once to a machine on a construction site.
Notably, the distribution of class sparsity has significant impact. While QPM is limited to a class similarity of up to $80\%$, due to deviating in at least one feature with all features sharing the same weight and every class being represented by $\gls{nperClass}=5$ features, \gls{layerName}, \gls{qsenn} and \glmtable{} all exhibit multiple (14, 9 and 5) class pairs, that have a class similarity of above $99\%$. 
The \gls{layerName} for instance repeatedly represents classes with one feature with positive weight and one with an extremely low negative weight, resulting even in cosine similarities of $1$ due to floating point precision. While this generally hurts interpretability, it can be beneficial for \cubsim{}.

Despite these issues, \Cref{stab:imgStruct} shows that QPM still performs comparatively to \gls{layerName} and \gls{qsenn} with their extremely high similarities and learns significantly more aligned representations than the dense baseline, even on \imgnetheader{}.
Future work might incorporate a more fine-grained class hierarchy, building upon the very general WordNet, into this metric or profit off of further improved language models.
\begin{table*}[t]
 \caption{Results for \oldloc{5}~\citep{norrenbrocktake} demonstrating its weakness to capture the locality of the by-design very local features of PIP-Net. Note that $20$ is the worst possible value.}
 \label{stab:oldloc5}
 \centering
 \begin{tabular}{l|cc}
  \toprule
  Method 
  & CUB&CARS \\
  \midrule
  Baseline \resnet{}& 61.1$\pm$0.4 &57.4$\pm$0.3\\
  \hline
  \glmtable{} & 55.3$\pm$0.5 &52.6$\pm$0.3  \\
  \pipnettable{} & 20.5$\pm$0.0 & 20.5$\pm$0.0  \\ 
  \protopooltable{} & 25.5$\pm$0.4  &23.4$\pm$0.5 \\
  \slddtable{} & 79.2$\pm$0.3 & 81.9$\pm$0.9 \\
  \qsenntable{}& 87.0$\pm$0.5 & 89.6$\pm$0.3 \\
  \midrule
   \gls{NewlayerName} (Ours) & \textbf{89.9}$\pm$0.2   &\textbf{ 91.4}$\pm$0.3 \\ 
  \bottomrule
 \end{tabular}
 \end{table*}
 \begin{table*}[t]
 \caption{Structural Grounding based on Clip Similarities on ImageNet}
 \label{stab:imgStruct}
 \centering
\begin{tabular}{l|c}
  \toprule
  Method & \cubsim{} \\
  \midrule
  Dense Resnet50 &  17.9$\pm0.0$  \\
  \glmtable{}& 10.3$\pm0.0$   \\
  \gls{layerName} & \textbf{36.9}$\pm$0.4  \\ 
  \qsenntable{} & 33.2$\pm$0.2 \\
  \midrule
\gls{NewlayerName} (Ours) & \underline{34.5}$\pm0.6$\\
  \bottomrule
 \end{tabular}
 \end{table*}
\section{Impact of Class-Feature Similarity Metric}
\label{ssec:auroc}
This section contains an ablation study on the choice of Pearson correlation as metric for the feature-class similarity matrix \simmat. 
While it captures the desired linear relationship, that is also utilized during the following predictions, an intuitive alternative is the Area under the receiver operating characteristic curve (AUROC), which is  highly non-linear and frequently used to capture the predictive power with a varying threshold.
\Cref{stab:auroctab} shows that AUROC is also suitable but inferior to the simple correlation.
 \begin{table*}[t]
 \caption{Accuracy with different criteria used as Feature-Class Similarity matrix \simmat{}}
 \label{stab:auroctab}
 \centering
\begin{tabular}{l|cc}
  \toprule
  \simmat{} Metric 
  & CUB&CARS \\
  \midrule
  AUROC & 84.8$\pm0.2$   & 91.6$\pm0.2$   \\
   Correlation (Ours) & \textbf{85.1}$\pm0.3$   &\textbf{ 91.8}$\pm0.1$   \\ 
  \bottomrule
 \end{tabular}
 \end{table*}
\section{Limitations and Future Work}
\label{ssec:limits}
This section discusses limitations for the proposed QPM and avenues for future work.

In this work, QPM is applied to the generally available 
 and typical datasets for image classification, with \imgnetheader{} indicating broad applicability.
However, QPM's high interpretability is especially beneficial for high-stakes applications such as the medical domain or autonomous driving, where each individual situation can not be accessed by an expert.
Rather, after training the QPM and before deploying it to cars, its class explanations can be obtained to gain insights into whether it is right for the right reasons and if these are robust to all deployment conditions.
Thus, applying QPM to suitable high-stakes applications is a promising avenue for future work.
However, to our knowledge, there is no suitable dataset from these domains published yet. 

A limitation of our QPM in its current form lies in its inability to model negative assignments.
Compared to the \gls{layerName} and \gls{qsenn}, which use negative weights, it is evident that the varied datasets used in this paper, do not require it. 
Further, while we believe that it is generally preferable to represent classes only using positive assignments, as e.g., also done by recent prototypical models~\citep{nauta2023pipnet, rymarczyk2022interpretable}, one can think of other datasets where negative reasoning may be superior.
If, e.g., all classes in a  dataset containing birds had a black beak, except for one with all other colors, it would likely be the most efficient solution to represent that one with a negative assignment on a feature activating on black beaks, rather than have every other class positively assigned to it, which the current QPM might do.
Thus, future work may incorporate negative assignments into the optimization, which might lead to even more compact representations.

As discussed in \cref{ssec:poly}, the learned features of our QPM are generally polysemantic, while potentially being monosemantic locally. 
For aligning them with human concepts, all post-hoc methods, such as TCAV~\citep{kim2018interpretability}, Clip-Dissect~\citep{oikarinen2023clipdissect}, or the alignment methods from \gls{layerName} or \gls{qsenn} can be applied.
Notably, aligning a feature with their human concepts is more beneficial for QPM than it is for e.g., black-box models, as they are used in an intuitively interpretable way. 
Further, the interpretable assignment can even help with alignment, as shown in \cref{ssec:poly}.
Nevertheless, polysemantic features are a challenge for interpretability and future work in this direction can focus on preventing their emergence while still using them in an interpretable way or robustly measuring alignment to multiple concepts.

For many explanations from our QPM, a saliency map for its individual features is used.
While we typically just visualize each individual feature map via upscaling, resulting in a comparable resolution to GradCAM~\citep{selvaraju2020grad}, other saliency methods, like Integrated Gradients~\citep{sundararajan2017axiomatic}, LRP~\citep{binder2016layer} or RISE~\citep{Petsiuk2018rise} can be applied.
Because QPM is backbone independent, even models with built-in more faithful saliency maps such as B-cos Networks~\citep{bohle2023holistically} can be used.
Since the use of these features is easy-to-interpret, evaluating the localizations of our model should focus on the feature explanations rather than class-level ones. 
Future work might incorporate these faithful saliency maps to measure insertion or deletion methods, akin to those used for class-level saliency maps~\citep{Petsiuk2018rise}.
Ideally, one is able to overcome the issue of moving out-of-distribution with removing pixels~\citep{hooker2019benchmark}.
Finally, the contrastive nature of QPM's features might lead to an intuitive threshold that can be used during the removal of pixels, similar to how previous metrics try to change the class prediction.
\section{Detailed Results}
\label{suppsec:Results}
\begin{figure*}[t]
  \centering
\includegraphics[width=\linewidth]{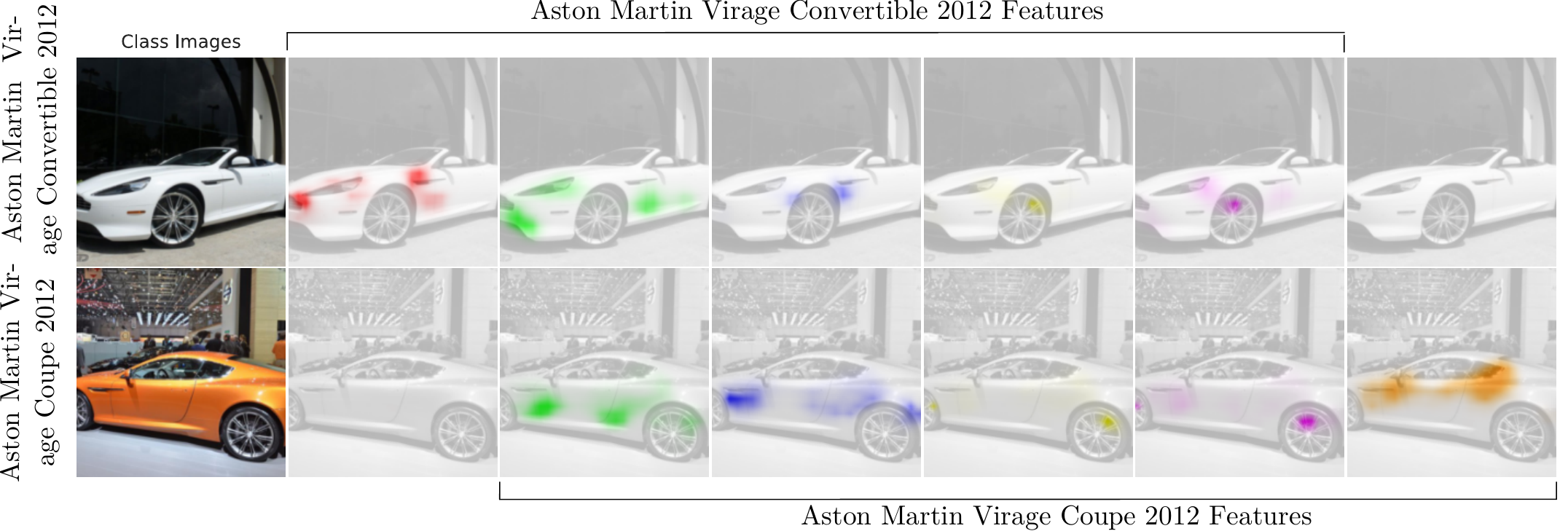}

   \caption{Faithful global interpretability of our \gls{NewlayerName} trained on \stanfordheader{}{}: 
   Without any additional supervision, 
   \gls{NewlayerName} learns to represent 
   the Convertible and Coupe Variant using  $5$ diverse and general features.}
  \label{fig:stanfordRep}
\end{figure*}

\begin{figure*}[t]
  \centering
\includegraphics[width=\linewidth]{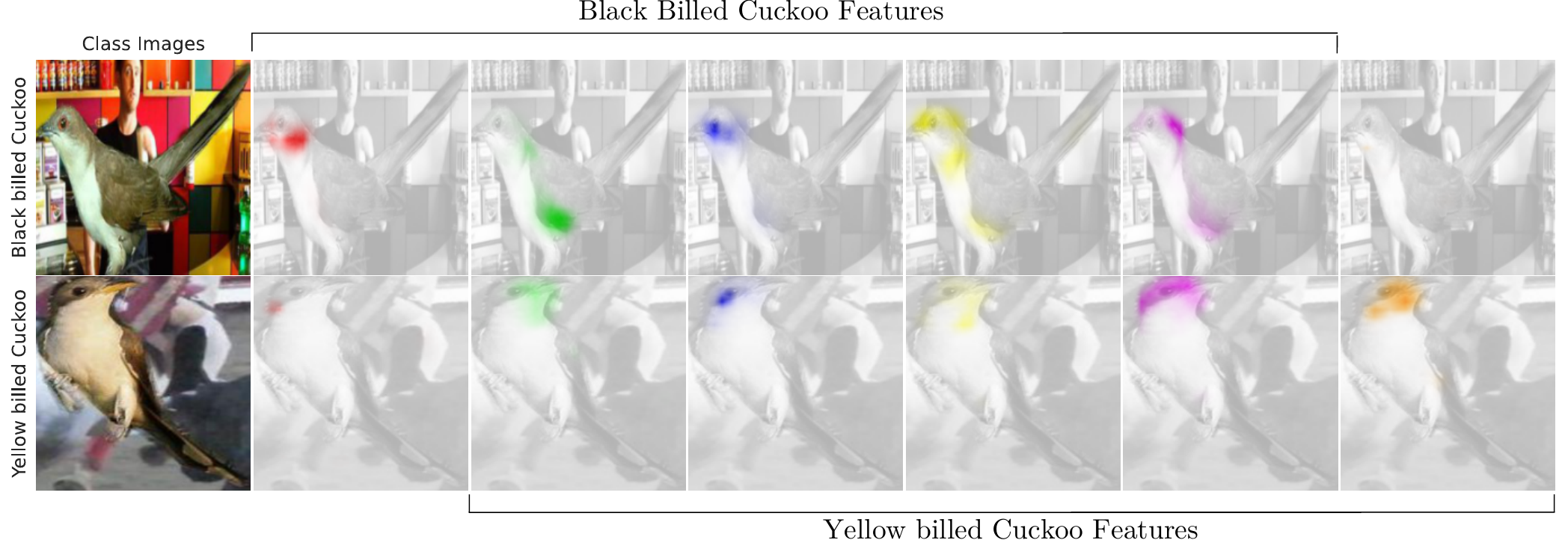}

   \caption{Faithful global interpretability of our \gls{NewlayerName} trained on \travelingheader{}: 
   Without any additional supervision, 
   \gls{NewlayerName} learns to represent 
   the Yellow and Black billed Cuckoo using  $5$ diverse and general features, correctly ignoring the correlated background.}
  \label{fig:TravRep}
\end{figure*}

\begin{figure*}[t]
  \centering
  \includegraphics[width=.9\linewidth]{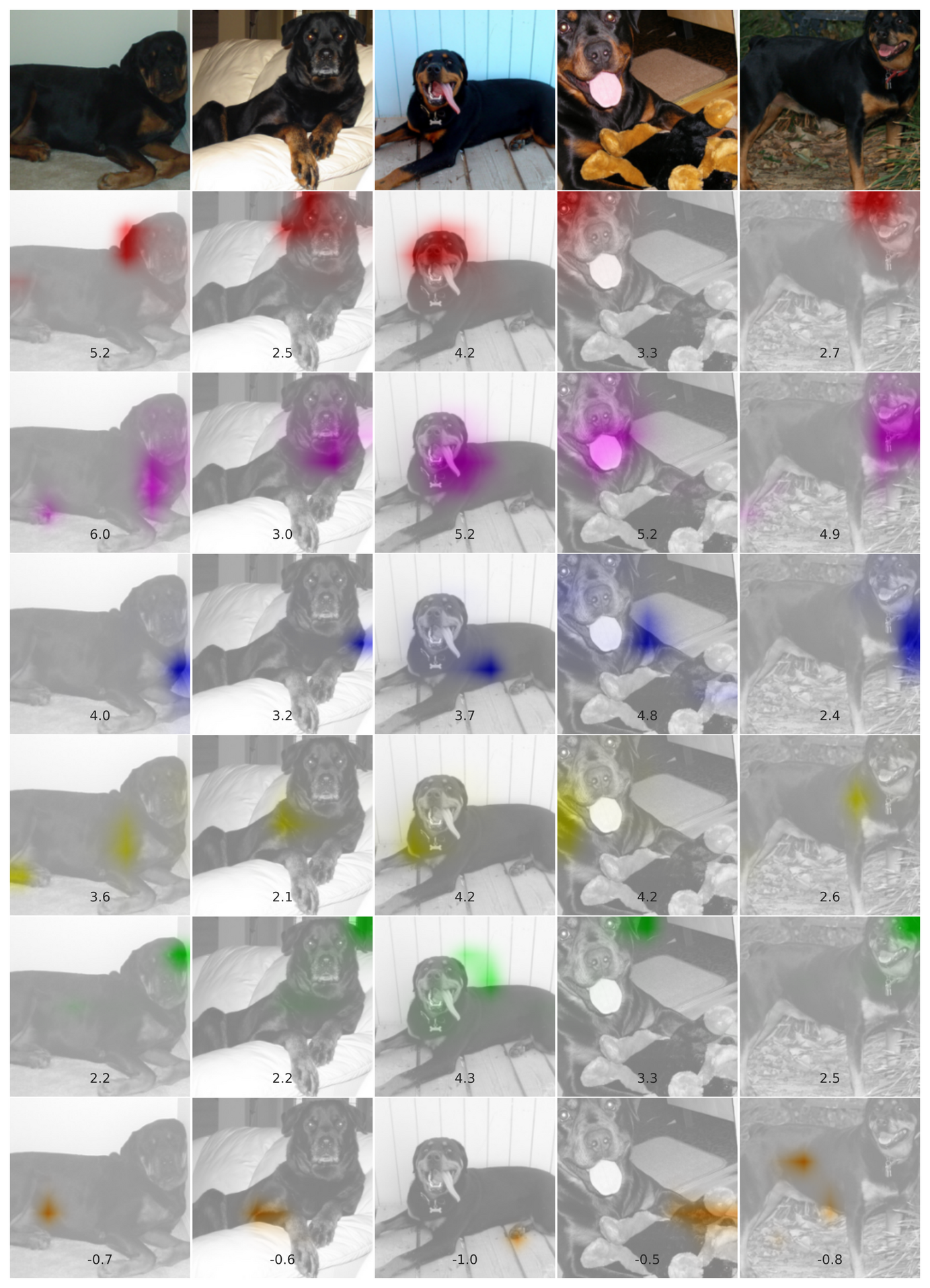}
 
  \caption{Exemplary Activations of Features in ~\cref{fig:metrics_full} on further Rottweiler images. 
The feature values after normalization are written on the images.
Note that all shown activations are scaled from 0 to 1, resulting in an arbitrary localization of the brown feature detecting the Doberman-like head.}
  \label{fig:rottweilerS}
\end{figure*}

\begin{figure*}[t]
  \centering
  \includegraphics[width=.9\linewidth]{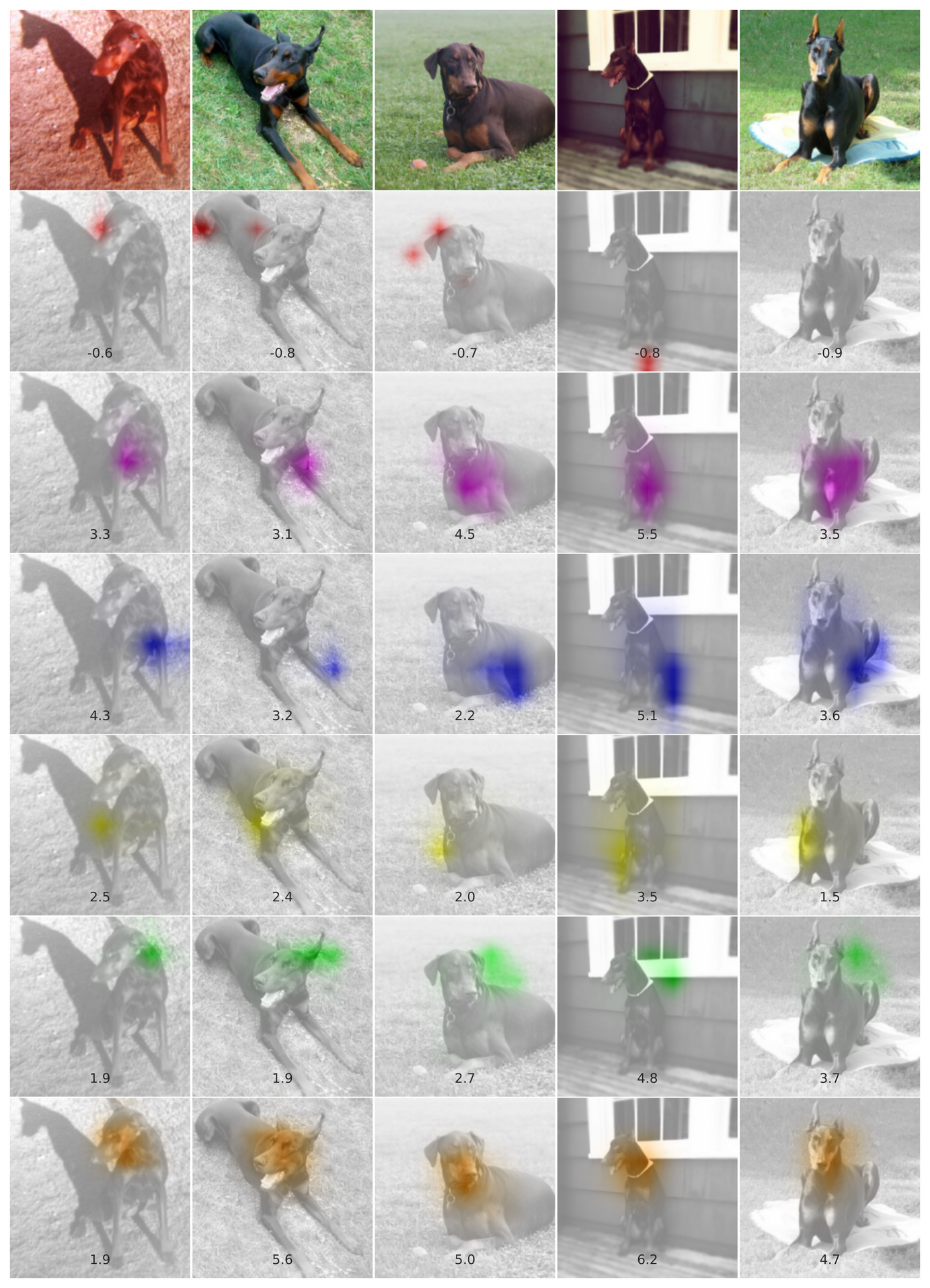}
 
  \caption{Exemplary Activations of Features in ~\cref{fig:metrics_full} on further Doberman images. 
The rounded feature values after normalization are written on the images.
Note that all shown activations are scaled from 0 to 1, resulting in an arbitrary localization of the red feature detecting the Rottweiler-like head.}
  \label{fig:dobermanS}
\end{figure*}

\begin{figure*}[t]
  \centering
  \includegraphics[width=.9\linewidth]{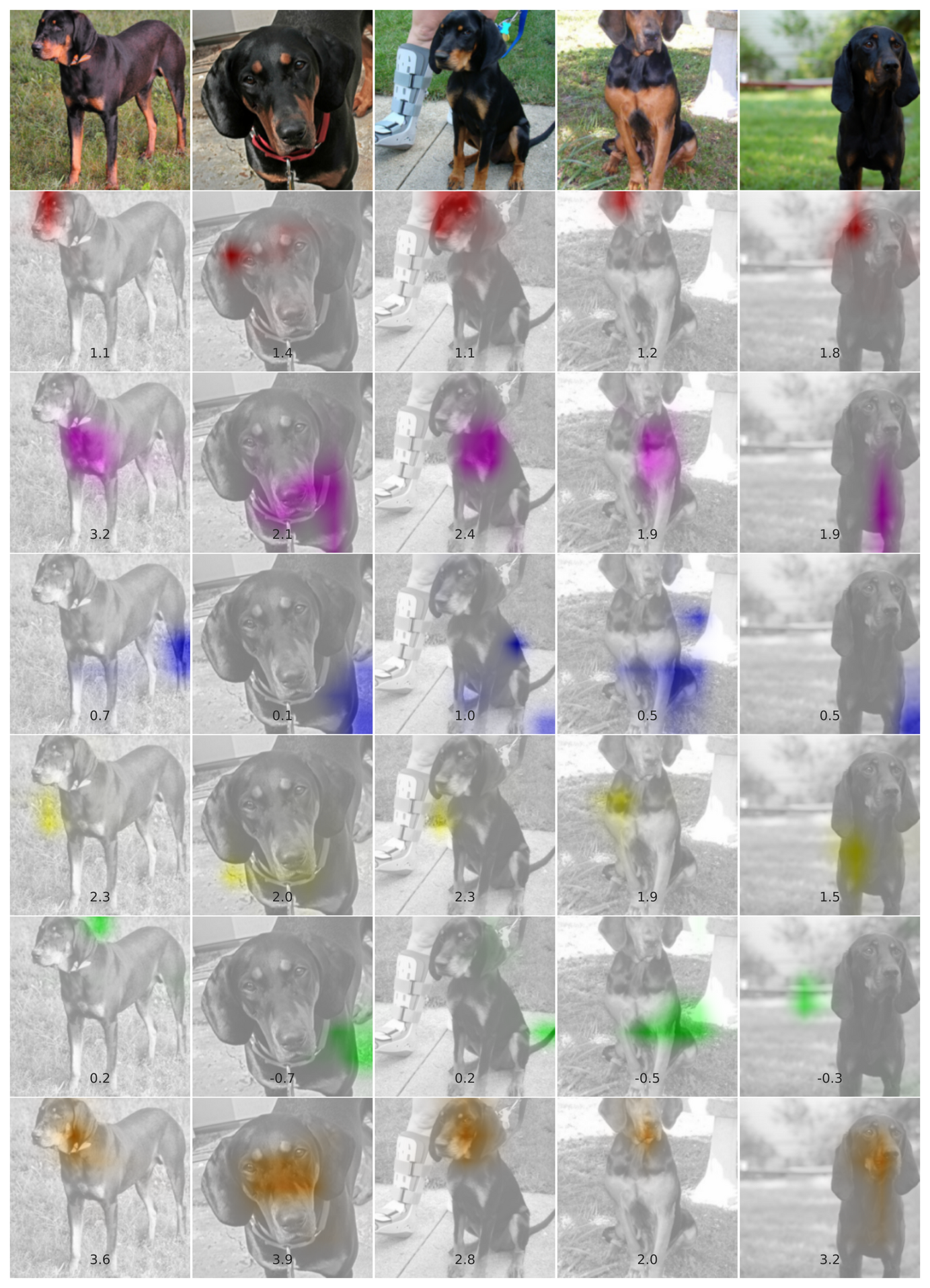}%
 
  \caption{Exemplary Activations of Features in ~\cref{fig:metrics_full} on Black and Tan Coonhound images. 
The rounded feature values after normalization are written on the images.
Note that all shown activations are scaled from 0 to 1, resulting in an arbitrary localization of the two not assigned and barely activated blue and green features. The fifth assigned feature is shared with dog types such as  Newfoundlands, bears and pandas, localizing on the neck region.}
  \label{fig:blacktanS}
\end{figure*}

This section contains detailed results with standard deviations, including experiments with 
Resnet34, \gls{incv}, Swin-Transformer-small and Swin-Transformer-tiny,
in \suppl{} \cref{stab:noCrop} to 
\cref{tstab:incInterpproto-table}.
The good results across architectures demonstrate an independence between backbone and our proposed method.
They further seem robust as the difference in mean is usually large compared to the standard deviation. 
Further, \cref{fig:rottweilerS,fig:dobermanS} show how the features of \cref{fig:metrics_full} continue to localize on the same human attribute across different poses.
Additionally, we included the activations of these features on images of another class in \cref{fig:blacktanS} to showcase the global interpretability enabled through the binary assignment of more interpretable features. 
Instead of the blue and green feature, this probed \gls{NewlayerName} recognizes the Black and Tan Coonhound through both doberman-like and rottweiler-like head features, as well as a neck that is also assigned to pandas or bears.
\Cref{fig:stanfordRep,fig:TravRep} additionally include examples for contrastive class representations learned on \stanfordheader{} and \travelingheader{}.
Finally, \cref{stab:oldloc5} contains results for \oldloc{5}, to quantify its inability to capture the high spatial diversity of PIP-Nets class detectors.
\section{Feature Diversity Loss}
\label{suppsec:ldiv}
This section further describes the Feature Diversity Loss~\gls{customLoss}, proposed in \cite{norrenbrocktake}.
It is defined per sample, for which the model predicted the
class $\hat{c}= \arg\max(\gls{outputVector})\label{eq:MaxClassDiv}$ and ensures a local diversity of the used feature maps \Gls{featureMaps}.
\begin{align}
\hat{s}^\findex_{ij} &=\frac{\exp(\gls{featureMaps}^\findex_{ij})}{\sum_{i'=1}^{\gls{featuresMapheigth}}\sum_{j'=1}^{\gls{featuresMapwidth}}\exp(\gls{featureMaps}^\findex_{i'j'})} \frac{\glsentrylong{featureVector}_\findex}{\max\glsentrylong{featureVector}}  \frac{|w_{\hat{c},\findex}|}{\Vert\boldsymbol{w}_{\hat{c}}\Vert_2} \label{eq:ScaleDiv}\\
     \gls{customLoss} &= -\sum_{i=1}^{\gls{featuresMapheigth}}\sum_{j=1}^{\gls{featuresMapwidth}}\max(\hat{s}^1_{ij},\hat{s}^2_{ij}, \dots, \hat{s}^{\gls{nFeatures}}_{ij})\label{eq:CCMPDiv}
\end{align}
Equation~\ref{eq:ScaleDiv} employs the softmax function to normalize the entries $\gls{featureMaps}^l_{ij}$ of the feature maps \Glspl{featureMaps} across spatial dimensions. It then scales the maps to emphasize visible and significant features, maintaining the relative mean of \Glspl{featureMaps} while weighting them according to the predicted class.
Equation~\ref{eq:CCMPDiv} then applies cross-channel-max-pooling of the normalized and scaled feature maps $\hat{\textbf{S}}$.
The result is negatively weighted and thus encourages the model to learn features that localize on different image regions.
The resulting total training loss is
\begin{equation}
    \mathcal{L}_{\mathrm{final}} =  \mathcal{L}_{CE} + \gls{cLW}\gls{customLoss},\label{seq:customLossW}
\end{equation} with $\gls{cLW}\in\mathbb{R}_+$ as weighting factor.
\section{Optimality of Solution}
\label{suppsec:optimal}
In order to test the optimality of our solution, we try to solve the problem without our relaxation in \cref{eq:NotGLob} with more compute and time.
We used $3$ days and $250$ GB on an AMD EPYC 72F3 to solve the problem globally across $5$ seeds on \cubheader{} with a target gap to optimality of $1\%$ to ensure sufficient deduplication.
The time limit was left to $3$ hours for one iteration, as otherwise multiple iterations would not finish.
Across the $5$ seeds used for \gls{NewlayerName}, the average obtained objective value for the global problem was $0.5\%$ above the one computed with our simplifications.  
Similar to our ablations in \cref{tab:ablations}, the resulting accuracy for the extensively optimized model was not improved, but even $0.1$ percent points lower.
As mentioned in \cref{sec:gurobitricks}, the objective does not perfectly correlate with downstream metrics, as the constants \simmat{}, \FSimeaturemat{} and \BSimeaturemat{} only approximate the desired behaviour.
However, the average gap to the best bound was still $3.2\%$, with only negligible progress during the final iteration, suggesting that a longer time limit would not significantly improve it. 
Note, that the best bound might be violating constraints, already added or not.
In summary, the gap between our easy-to-compute solution and an obtainable solution of the global problem is $0.5\%$, which leads to no improved model, with an upper bound on the gap of $3.7\%$ ($372$ to $386$).

\section{Standard Form for Quadratic Problem}
\label{suppsec:QP}
The quadratic problem, described in \cref{sec:QPMethod}, can be expressed in the standard form for quadratic programming problems.
The aim is to optimize quadratic problems of the form $\frac{1}{2} \mathbf{x}^T \mathbf{Q} \mathbf{x} + \mathbf{c}^T \mathbf{x}$ with respect to specified constraints.
To describe our quadratic problem in standard form, we therefore define the \textbf{variables} $\mathbf{x}$,  $\mathbf{Q}$, $\mathbf{c}$ as well as the \textbf{constraints}.
For notation, \zerovec{x} and \onevec{x} describe a vector with $x$ zeros or ones respectively and \zeromat{m}{n} describes a $m\times n$ matrix of zeros.

\paragraph{Variables}
Let $\mathbf{x}$ be the binary decision variable vector, combining \fvecgurobi{} and the vectorized form of $\mathbf{W}$:
\[
\mathbf{x} = \begin{bmatrix}
\mathbf{\fvecgurobi} \\
\text{vec}(\wgurobi)\\
\end{bmatrix}\in \{0,1\}^{n_f + n_c \cdot n_f}
\]

\paragraph{Objective Function}
The standard objective function includes all objectives:
\[
\text{Maximize:} \quad \frac{1}{2} \mathbf{x}^T \mathbf{Q} \mathbf{x} + \mathbf{c}^T \mathbf{x}
\]
Here
\begin{equation}
    \mathbf{Q} = \begin{bmatrix}
 -\FSimeaturemat & \zeromat{\gls{nFeatures}}{\gls{nFeatures}\cdot \gls{nClasses}}\\ %
\simmat_\mathrm{stack}& \zeromat{\gls{nFeatures}\cdot\gls{nClasses}}{\gls{nFeatures}\cdot \gls{nClasses}}\\
\end{bmatrix}
\end{equation}
combines all quadratic objectives and
\begin{equation}
\mathbf{c} =  \begin{bmatrix}
\BSimeaturemat \\
\zerovec{\gls{nFeatures}\cdot \gls{nClasses}}\\
\end{bmatrix}    
\end{equation}
the linear term.
Here
\begin{equation}
\simmat_\mathrm{stack} =  \begin{bmatrix}
diag(\boldsymbol{a}_1) \\
diag(\boldsymbol{a}_2)  \\
\vdots\\
diag(\boldsymbol{a}_{\gls{nClasses}}))
\end{bmatrix}    
\end{equation}
connects the vectorized entries of \wgurobi{} with \simmat{}.

\paragraph{Constraints}
\begin{enumerate}

\item  Constraint for the number of selected features (\cref{eq:FeatureSel}):
\begin{equation}
\begin{bmatrix}
\onevec{\gls{nFeatures}}\\ \zerovec{\gls{nFeatures}\cdot \gls{nClasses}}\\
\end{bmatrix}^T
\mathbf{x} = \gls{nReducedFeatures}%
\end{equation}
\item No assignments on unselected features:
\begin{align}
  \begin{bmatrix} 
  \boldsymbol{\mathrm{featureSum}} &
\zerovec{\gls{nFeatures}\cdot\gls{nClasses}}
\end{bmatrix}
(\onevec{\gls{nFeatures}\cdot(\gls{nClasses}+1)}-\mathbf{x}) =0 \\
\boldsymbol{\mathrm{featureSum}} = 
  \begin{bmatrix} 
 \zeromat{\gls{nFeatures} }{\gls{nFeatures}} &
\mathrm{\boldsymbol{FeatureSel}}^{\gls{nFeatures} }
\end{bmatrix}
\mathbf{x}
\end{align}
where $\boldsymbol{\mathrm{FeatureSel}}^{\gls{nFeatures} }\in\{0,1\}^{\gls{nFeatures}\times\gls{nFeatures}\cdot\gls{nClasses}}$ is a matrix of zeros with $\mathrm{FeatureSel}_{i,j}=1$ where $(j-i)\mod\gls{nFeatures}=0$.
The vector
$\boldsymbol{\mathrm{featureSum}}$ captures the total number of assignments per feature.
\item  Constraint for the number of assignments per class (\cref{eq:5perInit}):

\begin{equation}
\begin{bmatrix}
\zeromat{\gls{nClasses}}{\gls{nFeatures}}
& \mathrm{\boldsymbol{UBD}}^{\gls{nClasses},\gls{nFeatures} } %
\end{bmatrix}
\mathbf{x} =\gls{nperClass} \cdot \onevec{ \gls{nClasses}}  %
\label{eq:5perclass}
\end{equation}

Where the upper block diagonal matrix
\begin{equation}
  \mathrm{\boldsymbol{UBD}}^{\gls{nClasses},\gls{nFeatures} } = \begin{bmatrix}
\onevec{ \gls{nFeatures}} & \zerovec{ \gls{nFeatures}} & \cdots & \zerovec{ \gls{nFeatures}}  \\
\zerovec{ \gls{nFeatures}} & \onevec{ \gls{nFeatures}} & \cdots  & \zerovec{ \gls{nFeatures}} \\
\vdots & \vdots & \vdots &\vdots &\\
\zerovec{ \gls{nFeatures}} & \zerovec{ \gls{nFeatures}} & \cdots & \onevec{ \gls{nFeatures}}  \\
\end{bmatrix}^T
\end{equation}
is
a block-diagonal matrix with \gls{nFeatures} ones per row, one $1$ in each of the $\gls{nFeatures}\cdot\gls{nClasses}$ columns and \gls{nClasses} total rows.
\item No duplicated classes (\cref{eq:Unique}):%
\begin{align}
  \begin{bmatrix} 
  \boldsymbol{Eq}^{\cindex,\cindex'} &
\zerovec{\gls{nFeatures}\cdot\gls{nClasses}}
\end{bmatrix}
\onevec{\gls{nFeatures}\cdot(\gls{nClasses}+1)} > 0 \quad \forall \cindex \neq \cindex ' &\in \{1, \dots,\gls{nClasses}\}\\
\boldsymbol{Eq}_\findex^{\cindex,\cindex'} = |\boldsymbol{x}_{\cindex\cdot\gls{nFeatures} + \findex} -
 \boldsymbol{x}_{\cindex'\cdot\gls{nFeatures} + \findex}| \quad \forall \findex &\in \{1, \dots,\gls{nFeatures}\}
\end{align}

\end{enumerate}

\begin{table*}[t]
 \caption{Accuracy without background removal based on the Ground-Truth with \resnet{}. \gls{NewlayerName} is less susceptible to the spuriously correlated backgrounds with and without Center Bias $\BSimeaturemat{}^{\mathrm{Center}}$. NA indicates no convergence.}
 \label{stab:noCrop}
 \centering
 \begin{tabular}{l|cc}
  \toprule
  Method 
  & CUB& TRAVEL 
  \\
  \midrule
  Baseline \resnet{}& 84.2$\pm$0.3 & 33.8$\pm$0.6\\
  \midrule
  \glmtable{}& 75.0$\pm$0.9 & 35.6$\pm$1.4 \\
  \pipnettable{} & 74.9$\pm$0.0 & 59.4$\pm$1.0 \\ %
  \protopooltable{}& 75.0$\pm$0.3 & NA \\
  \slddtable{} & 82.2$\pm$0.1 & 62.6$\pm$1.6 \\
   \qsenntable{} & \underline{82.8}$\pm$0.3  & \underline{67.0}$\pm$0.5  \\\midrule
   \gls{NewlayerName} (Ours) & \textbf{82.9}$\pm$0.1 & {64.7}$\pm$0.7 \\
  w/ Center Bias   $\BSimeaturemat{}^{\mathrm{Center}}$ & {82.4}$\pm$0.3 & \textbf{68.9}$\pm$0.5 \\
  \bottomrule
 \end{tabular}

 \end{table*}
 
 \begin{table*}[htbp]
 \caption{Ablation Studies investigating the impact of incorporating feature-feature similarity through \FSimeaturemat{} and locality bias \BSimeaturemat{} on \cubheader{} with \resnet{}.%
 }
 \label{stab:ablations}
 \centering
 \begin{tabular}{l|c|ccc}
  \toprule
 \BSimeaturemat & \FSimeaturemat 
   & Accuracy \arrowUp& \loc{5} \arrowUp & Correlation \arrowDown\\
  
  \midrule
 
 \xmark & \xmark & 84.6$\pm$0.4 & 89.5$\pm$0.2 & 33.9$\pm$0.8 \\ 
 
 \cmark & \xmark & 84.4$\pm$0.2 & \textbf{90.4}$\pm$0.3 & 33.5$\pm$9.4 \\ 
  \xmark & \cmark & \underline{85.0}$\pm$0.3 & 89.4$\pm$0.3 & \textbf{22.7}$\pm$1.1 \\
   \midrule
   \cmark & \cmark &
   \textbf{85.1}$\pm$0.3 & \underline{90.1}$\pm$0.3 & \underline{24.6}$\pm$1.1\\

   \bottomrule
 \end{tabular}

\end{table*}

\begin{sidewaysfigure}
     \centering\includegraphics[width=\linewidth]{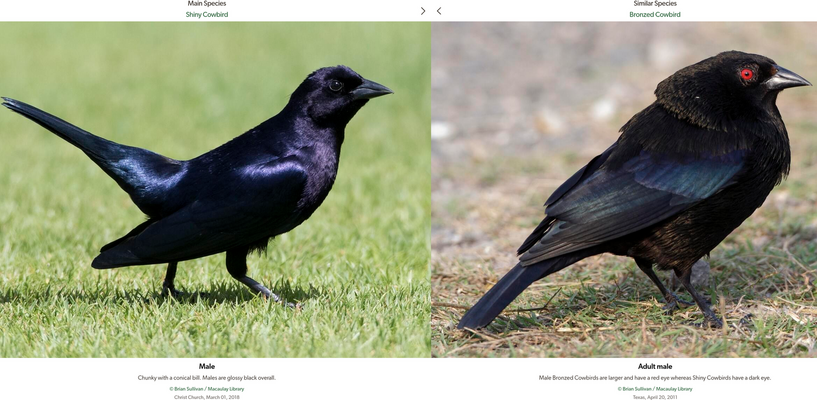}
     \caption{
     Screenshot of the Cornell-Lab website~\citep{Sullivan}, describing how the similar species shiny and bronzed cowbird differ:
     The only differences explained are in the size, which is not usable for our \gls{NewlayerName} due to cropping, and the eye color, which our \gls{NewlayerName} learned without any supervision and explained  in \cref{fig:CubSim} as the differentiating factor.}
     \label{fig:ScreenshotDiffShiny}
\end{sidewaysfigure}
\begin{sidewaystable}[htbp]
  \centering

    \caption{
  \accmetricstablestart{\resnet{}}
  The compactness-accuracy trade-off is 
 shown 
 in \cref{fig:interptradeoff}.
  \boldnessstatement
  }
  \label{stab:aCCproto-table}
  \begin{tabular}{l|cccc|cccc|cccc}
    \toprule
    Method &  \multicolumn{4}{c|}{Accuracy \arrowUp} &
    \multicolumn{4}{c|}{Total Features\arrowDown} & \multicolumn{4}{c}{Features / Class\arrowDown} \\
  & CUB& CARS & TRAVEL & IMGNET & CUB& CARS & TRAVEL & IMGNET & CUB& CARS &TRAVEL & IMGNET
    \\
    \midrule
    Baseline \resnet{} & {86.6}$\pm$0.2 & {92.1}$\pm$0.1 & {59.5}$\pm$0.9 &76.1  & 2048 & 2048 &2048& 2048& 2048& 2048 &2048&2048\\
    \midrule
    \glmtable{} & {78.0}$\pm$0.4 & {86.8}$\pm$0.6 & {58.0}$\pm$1.2 & 58.0$\pm$0.0 & 809$\pm$8 & 807$\pm$10 & 781$\pm$6 & 1627$\pm$1 & \textbf{5}& \textbf{5} &\textbf{5}&\textbf{5} \\
    \pipnettable{} & 82.0$\pm$0.3 & 86.5$\pm$0.3 & 74.0$\pm$0.4  &-& 731$\pm$19 &669$\pm$13& 744$\pm$15  &-& 12 &11 & 6& - \\ %
    \protopooltable{} & 79.4$\pm$0.4  &87.5$\pm$0.2 & 38.1$\pm$1.9 &-& 202 & 195 &202& -  &202 & 195 & 202 & - \\ %
    \slddtable{}  & {84.5}$\pm$0.2 & {91.1}$\pm$0.1 & \underline{75.6}$\pm$0.2 & 72.7$\pm$0.0 & \textbf{50} & \textbf{50} &\textbf{50}& \textbf{50}& \textbf{5}& \textbf{5} &\textbf{5}&\textbf{5} \\
     \midrule
     \gls{NewlayerName} (Ours)  & \underline{85.1}$\pm$0.3 & \underline{91.8}$\pm$0.3 & \textbf{75.7}$\pm$0.8& \underline{74.2}$\pm$0.0 & \textbf{50} & \textbf{50} &\textbf{50}& \textbf{50}& \textbf{5}& \textbf{5} &\textbf{5}&\textbf{5} \\ %
    $\gls{nperClass}=10$ (Ours)  & \textbf{85.7}$\pm$0.2 & \textbf{92.1}$\pm$0.1 & {75.2}$\pm$0.6  & \textbf{74.5}$\pm$0.1 & \textbf{50} &\textbf{50}& \textbf{50}&\textbf{50}& 10 &10 &10 &10 \\
    \bottomrule
  \end{tabular}

\end{sidewaystable}
\begin{sidewaystable}[htbp]
  \centering
  \caption{
  \interpmetricstablestart{\resnet{}}}
  \resizebox{\textwidth}{!}{
  \begin{tabular}{l|cccc|cccc|cccc|cc}
    \toprule
    Method &  \multicolumn{4}{c|}{\loc{5} \arrowUp} &
    \multicolumn{4}{c|}{\generality{}\arrowUp} & \multicolumn{4}{c}{\contrastiveness{}\arrowUp} & \multicolumn{2}{c}{\cubsim{}\arrowUp} \\
   & CUB& CARS & TRAVEL & IMGNET & CUB& CARS & TRAVEL & IMGNET & CUB& CARS &TRAVEL & IMGNET & CUB&TRAVEL
    \\
    \midrule
   Baseline \resnet{} & {57.7}$\pm$0.4 & {54.4}$\pm$0.3 & {59.9}$\pm$0.4 &37.1  & {98.0}$\pm$0.0 & {97.8}$\pm$0.0 & {98.0}$\pm$0.0&  99.4& {74.4}$\pm$0.1 & {75.1}$\pm$0.1 & {74.4}$\pm$0.1&71.6 &  {34.0}$\pm$0.3 &  {32.1}$\pm$0.2 \\
  \midrule
  \glmtable{}  & {55.4}$\pm$0.5 & {51.8}$\pm$0.3 & {56.0}$\pm$0.8&35.8$\pm$0.0   & \underline{97.8}$\pm$0.0 & \underline{97.6}$\pm$0.0 & \underline{97.8}$\pm$0.0 & \underline{99.4}$\pm$0.0  & {74.0}$\pm$0.1 & {74.5}$\pm$0.1 & {73.8}$\pm$0.1&71.7$\pm$0.0 & {2.5}$\pm$1.0 & {4.4}$\pm$2.8 \\
    \pipnettable{} & \textbf{99.2}$\pm$0.1 & \textbf{99.0}$\pm$0.1 & \textbf{98.7}$\pm$0.1&-  & 75.6$\pm$0.4 &62.9$\pm$0.1 & 73.4$\pm$0.1&-  & \textbf{99.6}$\pm$0.0 &\textbf{99.7}$\pm$0.0 & \textbf{99.7}$\pm$0.0&- & 6.7$\pm$0.9 & 6.9$\pm$1.3 \\ %
    \protopooltable{} & 24.5$\pm$0.8 &30.7$\pm$3.4& 31.5$\pm$1.6&-& 96.9$\pm$0.1 & 96.0$\pm$0.5 &95.5$\pm$0.1&- &76.7$\pm$1.0& 78.9$\pm$2.0 & 85.2$\pm$0.5&- & 13.9$\pm$0.9 & 7.6$\pm$2.5 \\
    \slddtable{}
     & {88.2}$\pm$0.2 & {88.6}$\pm$0.6 & {87.5}$\pm$0.4&\underline{64.7}$\pm$0.7 & {96.2}$\pm$0.1 & {95.5}$\pm$0.1 & {96.5}$\pm$0.1&{98.6}$\pm$0.0  & {87.3}$\pm$0.2 & {89.7}$\pm$0.3& {86.3}$\pm$0.2&\textbf{93.4}$\pm$0.1 &   {29.2}$\pm$4.0 &  {30.7}$\pm$3.1  \\
      \midrule
     \gls{NewlayerName} (Ours)  & {90.1}$\pm$0.3 & {89.6}$\pm$0.4 & {89.7}$\pm$0.2&{64.1}$\pm$0.7  & {97.0}$\pm$0.0 & {96.5}$\pm$0.0 & {97.2}$\pm$0.0 &{99.1}$\pm$0.0&   \underline{96.0}$\pm$0.4 & {97.7}$\pm$0.4& {94.0}$\pm$0.3&\underline{89.3}$\pm$0.1  &   \underline{47.9}$\pm$2.7 & \underline{54.3}$\pm$4.0\\
     $\gls{nperClass} = 10$ (Ours)  & \underline{95.8}$\pm$0.7 & \underline{96.6}$\pm$0.5 & \underline{95.1}$\pm$0.3 & \textbf{80.1}$\pm$0.9 & \textbf{98.1}$\pm$0.0 & \textbf{98.0}$\pm$0.1 & \textbf{98.1}$\pm$0.0 & \textbf{99.5}$\pm$0.0 & {95.9}$\pm$0.5 & \underline{98.6}$\pm$0.2&\underline{94.2}$\pm$0.3& 87.4$\pm$0.1 & \textbf{52.3}$\pm$1.6 &  \textbf{62.9}$\pm$2.8\\
    \bottomrule
  \end{tabular}
  }
\end{sidewaystable}
\begin{sidewaystable}[htbp]
  \caption{
   \accmetricstablestart{Resnet34}
   \boldnessstatement
  }
  \label{stab:r34aCCproto-table}
  \centering
  \begin{tabular}{l|ccc|ccc|ccc}
    \toprule
    Method &  \multicolumn{3}{c|}{Accuracy \arrowUp} &
    \multicolumn{3}{c|}{Total Features\arrowDown} & \multicolumn{3}{c}{Features / Class\arrowDown} \\
   & CUB& CARS & TRAVEL   & CUB& CARS & TRAVEL  & CUB& CARS &TRAVEL 
    \\
    \midrule
    Baseline Resnet34 & 85.7$\pm$0.3 & 91.5$\pm$0.2 & 61.3$\pm$0.4   & 2048 & 2048 &2048&  2048& 2048 &2048\\
    \midrule
    \glmtable{} & {72.0}$\pm$1.0 & {82.0}$\pm$0.6 & {53.5}$\pm$0.8 & 442$\pm$5 & 453$\pm$6 & 443$\pm$6  & \textbf{5}& \textbf{5} &\textbf{5} \\
    \slddtable{} & \underline{83.2}$\pm$0.3 & {90.7}$\pm$0.3 & {74.0}$\pm$0.2  & \textbf{50} & \textbf{50} &\textbf{50}&  \textbf{5}& \textbf{5} &\textbf{5} \\
     \midrule
     \gls{NewlayerName} (Ours)  & {83.0}$\pm$0.2 & \underline{91.3}$\pm$0.0 & \underline{75.1}$\pm$0.3& \textbf{50} & \textbf{50} &\textbf{50}&  \textbf{5}& \textbf{5} &\textbf{5}\\ %
    $\gls{nperClass}=10$ (Ours) & \textbf{83.9}$\pm$0.1 & \textbf{91.7}$\pm$0.1 & \textbf{75.7}$\pm$0.6 & \textbf{50} &\textbf{50}& \textbf{50}& 10 &10 &10  \\
    \bottomrule
  \end{tabular}

\end{sidewaystable}
\begin{sidewaystable}[htbp]
  \caption{
  \interpmetricstablestart{Resnet34}}
  \label{stab:r34Interpproto-table}
  \centering
  \begin{tabular}{l|ccc|ccc|ccc|cc}
    \toprule
    Method &  \multicolumn{3}{c|}{\loc{5} \arrowUp} &
    \multicolumn{3}{c|}{\generality{}\arrowUp} & \multicolumn{3}{c}{\contrastiveness{}\arrowUp} & \multicolumn{2}{c}{\cubsim{}\arrowUp} \\
  & CUB& CARS & TRAVEL   & CUB& CARS & TRAVEL   & CUB& CARS &TRAVEL   & CUB&TRAVEL
    \\
    \midrule
    Baseline Resnet34  & {62.1}$\pm$0.3 & {56.6}$\pm$0.4 & {64.1}$\pm$0.6  & {97.9}$\pm$0.0 & {97.7}$\pm$0.0 & {98.0}$\pm$0.0   & {76.4}$\pm$0.1 & {77.9}$\pm$0.2 & {76.0}$\pm$0.1   & {39.6}$\pm$0.2  & {36.0}$\pm$0.4 \\
    \midrule
    \glmtable{}  & {59.9}$\pm$0.4 & {55.3}$\pm$0.3 & {60.6}$\pm$0.4
  & \textbf{97.9}$\pm$0.0 & \underline{97.7}$\pm$0.0 & \underline{97.9}$\pm$0.0& {76.5}$\pm$0.0 & {77.8}$\pm$0.2 & {76.0}$\pm$0.1  &   {7.6}$\pm$2.2  & {9.4}$\pm$3.8 \\
    \slddtable{}
     & {90.1}$\pm$0.8 & {86.7}$\pm$2.5 & {87.3}$\pm$0.3    & {97.5}$\pm$0.0 & {97.6}$\pm$0.2 & {97.8}$\pm$0.0 & {86.0}$\pm$1.0 & {83.3}$\pm$4.6 & {82.0}$\pm$1.4  & {24.5}$\pm$2.7  & {29.4}$\pm$5.2 \\
      \midrule
     \gls{NewlayerName} (Ours)  & \underline{90.5}$\pm$0.5 & \underline{89.1}$\pm$1.1 & \underline{89.7}$\pm$0.7 & {97.5}$\pm$0.0 & {96.9}$\pm$0.1 & {97.6}$\pm$0.0  &   \underline{95.5}$\pm$0.2 & \underline{94.7}$\pm$1.1 & \underline{94.3}$\pm$0.5 & \underline{39.0}$\pm$2.9 & \underline{49.7}$\pm$4.7\\
     $\gls{nperClass} = 10$ (Ours)  & \textbf{97.0}$\pm$0.3 & \textbf{98.1}$\pm$0.5  & \textbf{98.3}$\pm$0.0 & \textbf{97.9}$\pm$0.1 & \textbf{98.4}$\pm$0.0 & {96.1}$\pm$0.1 & \textbf{96.9}$\pm$0.3 & \textbf{98.7}$\pm$0.5 & \textbf{95.4}$\pm$0.2  & \textbf{54.7}$\pm$3.8 &  \textbf{60.3}$\pm$2.0\\
    \bottomrule
  \end{tabular}

\end{sidewaystable}
\begin{sidewaystable}[htbp]
  \caption{
  \accmetricstablestart{\gls{incv}}
  \boldnessstatement
  }
  \label{stab:incaCCproto-table}
  \centering
  \begin{tabular}{l|ccc|ccc|ccc}
    \toprule
    Method &  \multicolumn{3}{c|}{Accuracy \arrowUp} &
    \multicolumn{3}{c|}{Total Features\arrowDown} & \multicolumn{3}{c}{Features / Class\arrowDown} \\
   & CUB& CARS & TRAVEL   & CUB& CARS & TRAVEL   & CUB& CARS &TRAVEL  
    \\
    \midrule
    Baseline  \gls{incv} & {86.1}$\pm$0.1 & {92.6}$\pm$0.2 & {68.7}$\pm$0.4 & 2048 & 2048 & 2048& 2048& 2048 &2048\\
    \midrule
    \glmtable{} & {79.2}$\pm$0.5 & {89.3}$\pm$0.3 & {63.4}$\pm$0.5  & 814$\pm$9 & 795$\pm$8 & 813$\pm$9 & \textbf{5}& \textbf{5} &\textbf{5} \\
    \slddtable{} & {83.1}$\pm$0.4 & {91.1}$\pm$0.2 & {69.9}$\pm$0.2  &  \textbf{50} & \textbf{50} &\textbf{50}& \textbf{5}& \textbf{5} &\textbf{5} \\
     \midrule
     \gls{NewlayerName} (Ours)  & \underline{84.2}$\pm$0.4 & \textbf{{91.7}}$\pm$0.1 & \textbf{71.5}$\pm$0.4 & \textbf{50} & \textbf{50} &\textbf{50}&  \textbf{5}& \textbf{5} &\textbf{5} \\ %
    $\gls{nperClass}=10$ (Ours)  & \textbf{84.4}$\pm$0.4 & \textbf{91.7}$\pm$0.2 & \underline{70.8}$\pm$0.3 &\textbf{50} &\textbf{50}& \textbf{50}& 10 &10 &10  \\
    \bottomrule
  \end{tabular}

\end{sidewaystable}
\begin{sidewaystable}[htbp]
  \caption{
  \interpmetricstablestart{\gls{incv}}}
  \label{stab:incInterpproto-table}
  \centering
  \begin{tabular}{l|ccc|ccc|ccc|cc}
    \toprule
    Method &  \multicolumn{3}{c|}{\loc{5} \arrowUp} &
    \multicolumn{3}{c|}{\generality{}\arrowUp} & \multicolumn{3}{c}{\contrastiveness{}\arrowUp} & \multicolumn{2}{c}{\cubsim{}\arrowUp} \\
  & CUB& CARS & TRAVEL   & CUB& CARS & TRAVEL   & CUB& CARS &TRAVEL   & CUB&TRAVEL
    \\
    \midrule
    Baseline  \gls{incv} & {38.9}$\pm$0.3 & {33.1}$\pm$0.2 & {40.7}$\pm$0.4  & {96.1}$\pm$0.0 & {95.7}$\pm$0.0 & {95.9}$\pm$0.0  & {89.6}$\pm$0.2 & {91.7}$\pm$0.2 & {89.8}$\pm$0.1& {7.1}$\pm$9.6 &  {24.1}$\pm$0.3 \\
    \midrule
    \glmtable{}  & {39.3}$\pm$0.2 & {34.0}$\pm$0.4 & {41.0}$\pm$0.3  & \underline{95.4}$\pm$0.0 & \underline{95.0}$\pm$0.0 & \underline{95.3}$\pm$0.0 & {91.3}$\pm$0.3 & {93.4}$\pm$0.2 & {91.2}$\pm$0.1 & {0.3}$\pm$0.4 &  {2.5}$\pm$2.8 \\
    \slddtable{}
     & \textbf{58.1}$\pm$1.2 & \textbf{52.1}$\pm$1.5 & \textbf{60.5}$\pm$1.3  & {92.6}$\pm$0.1 & {92.1}$\pm$0.1 & {92.6}$\pm$0.2 & \underline{93.0}$\pm$0.3 & \textbf{94.4}$\pm$0.2 & {92.3}$\pm$0.3 & {24.4}$\pm$2.3 &  {27.2}$\pm$5.2 \\
      \midrule
     \gls{NewlayerName} (Ours) & {48.6}$\pm$0.9 & {42.8}$\pm$0.8 & {50.2}$\pm$0.4 & {95.1}$\pm$0.1 & {94.7}$\pm$0.0 & {95.1}$\pm$0.1 & \textbf{93.4}$\pm$0.1 & \underline{94.3}$\pm$0.1 & \textbf{93.4}$\pm$0.1&  \underline{34.8}$\pm$3.4 &  \underline{44.3}$\pm$3.8 \\
     $\gls{nperClass}=10$ (Ours)  & \underline{54.6}$\pm$0.6 & \underline{47.1}$\pm$0.5 & \underline{55.0}$\pm$1.4  &   \textbf{96.9}$\pm$0.1 & \textbf{96.8}$\pm$0.0 & \textbf{96.9}$\pm$0.1  & {92.6}$\pm$0.2 & {93.6}$\pm$0.2 & \underline{92.6}$\pm$0.2  & \textbf{41.9}$\pm$3.0 &  \textbf{50.3}$\pm$1.6 \\
    \bottomrule
  \end{tabular}

\end{sidewaystable}

\begin{sidewaystable}[htbp]
  \caption{
  \accmetricstablestart{Swin Transformer small}
  \boldnessstatement
  }
  \label{sstab:incaCCproto-table}
  \centering
  \begin{tabular}{l|ccc|ccc|ccc}
    \toprule
    Method &  \multicolumn{3}{c|}{Accuracy \arrowUp} &
    \multicolumn{3}{c|}{Total Features\arrowDown} & \multicolumn{3}{c}{Features / Class\arrowDown} \\
   & CUB& CARS & TRAVEL   & CUB& CARS & TRAVEL   & CUB& CARS &TRAVEL  
    \\
    \midrule
    Baseline  Swin Transformer small & {87.0}$\pm$0.1 & {90.6}$\pm$0.6 & {59.0}$\pm$0.4  & 768 & 768 & 768& 768& 768 &768\\
    \midrule
    \glmtable{}  & {76.5}$\pm$0.4 & {75.5}$\pm$1.2 & {46.4}$\pm$1.0  & 572$\pm$4 & 559$\pm$8 & 561$\pm$11 & \textbf{5}& \textbf{5} &\textbf{5} \\
    \slddtable{} &   \underline{85.3}$\pm$0.4  & \textbf{89.1}$\pm$0.7 & {60.3}$\pm$1.0 &  \textbf{50} & \textbf{50} &\textbf{50}& \textbf{5}& \textbf{5} &\textbf{5} \\
     \midrule
     \gls{NewlayerName} (Ours)   & {85.0}$\pm$0.4 & {88.7}$\pm$0.5 & \textbf{61.0}$\pm$1.0& \textbf{50} & \textbf{50} &\textbf{50}&  \textbf{5}& \textbf{5} &\textbf{5} \\ %
    $\gls{nperClass}=10$ (Ours)  &  \textbf{85.4}$\pm$0.3&  \underline{89.0}$\pm$0.8 & \underline{60.9}$\pm$1.1 &\textbf{50} &\textbf{50}& \textbf{50}& 10 &10 &10  \\
    \bottomrule
  \end{tabular}

\end{sidewaystable}
\begin{sidewaystable}[htbp]

  \caption{
  \interpmetricstablestart{Swin Transformer small}}
  \label{sstab:incInterpproto-table}
  \resizebox{\textwidth}{!}{
  \centering
  \begin{tabular}{l|ccc|ccc|ccc|cc}
    \toprule
    Method &  \multicolumn{3}{c|}{\loc{5} \arrowUp} &
    \multicolumn{3}{c|}{\generality{}\arrowUp} & \multicolumn{3}{c}{\contrastiveness{}\arrowUp} & \multicolumn{2}{c}{\cubsim{}\arrowUp} \\
  & CUB& CARS & TRAVEL   & CUB& CARS & TRAVEL   & CUB& CARS &TRAVEL   & CUB&TRAVEL
    \\
    \midrule
    Baseline  Swin Transformer small  & {26.4}$\pm$0.1 & {26.0}$\pm$0.1 & {29.6}$\pm$0.2   & {96.8}$\pm$0.0 & f{96.6}$\pm$0.0 & {96.9}$\pm$0.1  & {98.3}$\pm$0.1 & {98.8}$\pm$0.1 & {97.4}$\pm$0.1 & {24.5}$\pm$0.4 & {23.6}$\pm$0.9 \\
    \midrule
    \glmtable{} & {26.4}$\pm$0.2 & {26.1}$\pm$0.1 & {30.0}$\pm$0.4 &  \underline{96.6}$\pm$0.0 & \underline{96.4}$\pm$0.0 & \underline{96.7}$\pm$0.0  & \textbf{99.1}$\pm$0.1 & \textbf{99.6}$\pm$0.0 & {98.3}$\pm$0.1  & {8.8}$\pm$2.8  & {6.9}$\pm$1.5 \\
    \slddtable{}
     & \underline{38.0}$\pm$0.5 & \underline{35.6}$\pm$1.1 & \underline{43.2}$\pm$1.4 & {93.4}$\pm$0.1 & {93.3}$\pm$0.2 & {93.6}$\pm$0.2  & \underline{99.0}$\pm$0.2 & \underline{99.4}$\pm$0.2 & \textbf{99.3}$\pm$0.4  & {37.2}$\pm$3.4 &  {40.6}$\pm$2.4\\
      \midrule
     \gls{NewlayerName} (Ours) & {33.6}$\pm$0.4 & {32.0}$\pm$0.3 & {37.9}$\pm$0.7  & {95.2}$\pm$0.0 & {94.7}$\pm$0.0 & {95.2}$\pm$0.1 &  {98.5}$\pm$0.3 & {99.1}$\pm$0.2 & \underline{99.1}$\pm$0.3   & \underline{45.1}$\pm$3.2 & \underline{43.1}$\pm$3.4 \\
     $\gls{nperClass}=10$ (Ours)  & \textbf{40.4}$\pm$0.7 & \textbf{37.3}$\pm$1.7 & \textbf{44.6}$\pm$1.2  & \textbf{96.9}$\pm$0.0 & \textbf{96.6}$\pm$0.0 & \textbf{96.9}$\pm$0.1   & {95.5}$\pm$0.3 & {97.7}$\pm$0.6 & {94.7}$\pm$1.0  & \textbf{52.1}$\pm$2.1 & \textbf{52.5}$\pm$2.0 \\
    \bottomrule
  \end{tabular}
 }

\end{sidewaystable}

\begin{sidewaystable}[htbp]

  \caption{
  \accmetricstablestart{Swin Transformer tiny}
  \boldnessstatement
  }
  \label{tstab:incaCCproto-table}
  \centering
  \begin{tabular}{l|ccc|ccc|ccc}
    \toprule
    Method &  \multicolumn{3}{c|}{Accuracy \arrowUp} &
    \multicolumn{3}{c|}{Total Features\arrowDown} & \multicolumn{3}{c}{Features / Class\arrowDown} \\
   & CUB& CARS & TRAVEL   & CUB& CARS & TRAVEL   & CUB& CARS &TRAVEL  
    \\
    \midrule
    Baseline  Swin Transformer tiny & {86.6}$\pm$0.2 & {90.3}$\pm$0.3 & {60.1}$\pm$0.6 & 768 & 768 & 768& 768& 768 &768\\
    \midrule
    \glmtable{} & {71.8}$\pm$2.0 & {70.8}$\pm$1.4 & {46.6}$\pm$1.5  & 559$\pm$15 & 559$\pm$9 & 569$\pm$10 & \textbf{5}& \textbf{5} &\textbf{5} \\
    \slddtable{} & {84.4}$\pm$0.3 & \textbf{88.4}$\pm$1.3 & {58.3}$\pm$0.7 &  \textbf{50} & \textbf{50} &\textbf{50}& \textbf{5}& \textbf{5} &\textbf{5} \\
     \midrule
     \gls{NewlayerName} (Ours)   & \underline{84.5}$\pm$0.4 & {88.0}$\pm$1.3 & \underline{59.3}$\pm$0.9  & \textbf{50} & \textbf{50} &\textbf{50}&  \textbf{5}& \textbf{5} &\textbf{5} \\ %
    $\gls{nperClass}=10$ (Ours)   & \textbf{84.6}$\pm$0.4 & \underline{88.3}$\pm$1.0 & \textbf{59.8}$\pm$0.9&\textbf{50} &\textbf{50}& \textbf{50}& 10 &10 &10  \\
    \bottomrule
  \end{tabular}

\end{sidewaystable}
\begin{sidewaystable}[htbp]
  \caption{
  \interpmetricstablestart{Swin Transformer tiny}}
  \label{tstab:incInterpproto-table}
  \resizebox{\textwidth}{!}{
  \centering
  \begin{tabular}{l|ccc|ccc|ccc|cc}
    \toprule
    Method &  \multicolumn{3}{c|}{\loc{5} \arrowUp} &
    \multicolumn{3}{c|}{\generality{}\arrowUp} & \multicolumn{3}{c}{\contrastiveness{}\arrowUp} & \multicolumn{2}{c}{\cubsim{}\arrowUp} \\
  & CUB& CARS & TRAVEL   & CUB& CARS & TRAVEL   & CUB& CARS &TRAVEL   & CUB&TRAVEL
    \\
    \midrule
    Baseline  Swin Transformer tiny  & {27.3}$\pm$0.2 & {26.3}$\pm$0.1 & {30.9}$\pm$0.2  & {96.8}$\pm$0.0 & {96.6}$\pm$0.0 & {96.9}$\pm$0.0 & {98.8}$\pm$0.0 & {99.0}$\pm$0.1 & {98.4}$\pm$0.1 & {26.1}$\pm$0.2 & {23.1}$\pm$0.4\\
    \midrule
    \glmtable{}   & {26.4}$\pm$0.2 & {25.9}$\pm$0.1 & {29.9}$\pm$0.3  & \underline{96.6}$\pm$0.0 & \textbf{96.5}$\pm$0.0 & \underline{96.7}$\pm$0.0& \textbf{99.3}$\pm$0.0 & \textbf{99.5}$\pm$0.1 & \textbf{98.9}$\pm$0.1  & {12.1}$\pm$3.9 & {6.1}$\pm$2.2 \\
    \slddtable{}
      & \underline{38.9}$\pm$0.6 & \underline{35.4}$\pm$1.6 & \underline{46.6}$\pm$0.7  & {93.3}$\pm$0.2 & {92.9}$\pm$0.1 & {93.6}$\pm$0.2 & \underline{99.2}$\pm$0.3 & \underline{99.3}$\pm$0.4 & \textbf{98.9}$\pm$0.2  & {43.8}$\pm$5.7 & {41.2}$\pm$1.4\\
      \midrule
     \gls{NewlayerName} (Ours)  & {36.9}$\pm$0.4 & {31.5}$\pm$1.4 & {41.8}$\pm$0.3 & {95.1}$\pm$0.1 & {94.6}$\pm$0.1 & {95.3}$\pm$0.1 & {98.2}$\pm$0.4 & {98.7}$\pm$0.4 & {98.4}$\pm$0.2& \underline{50.9}$\pm$4.9 & \underline{51.4}$\pm$2.1 \\
     $\gls{nperClass}=10$ (Ours)   & \textbf{44.6}$\pm$0.5 & \textbf{37.5}$\pm$0.4 & \textbf{48.5}$\pm$0.9  & \textbf{96.9}$\pm$0.0 & \textbf{96.5}$\pm$0.0 & \textbf{97.0}$\pm$0.0  & {93.8}$\pm$0.2 & {97.4}$\pm$0.8 & {92.8}$\pm$0.7   & \textbf{54.7}$\pm$3.8 & \textbf{54.5}$\pm$2.6\\
    \bottomrule
  \end{tabular}
  }

\end{sidewaystable}

\end{document}